\documentclass[twoside]{article}

\usepackage{pifont}
\usepackage{graphicx}
\usepackage[utf8]{inputenc} % allow utf-8 input
\usepackage[T1]{fontenc}    % use 8-bit T1 fonts
\usepackage{hyperref}       % hyperlinks
\usepackage{url}            % simple URL typesetting
\usepackage{booktabs}       % professional-quality tables
\usepackage{amsfonts}       % blackboard math symbols
\usepackage{nicefrac}       % compact symbols for 1/2, etc.
\usepackage{microtype}      % microtypography
\usepackage{subfig}
\usepackage{mathrsfs}
\usepackage{amsmath}
\usepackage{adjustbox}
\usepackage{algorithm} 
\usepackage{algpseudocode} 
\usepackage{color}

\renewcommand{\vec}[1]{\boldsymbol{#1}}
\newcommand{\given}{\, | \,}

\newcommand{\fromto}{\longrightarrow}

\newcommand*{\defeq}{\mathrel{\vcenter{\baselineskip0.5ex \lineskiplimit0pt
			\hbox{\footnotesize.}\hbox{\footnotesize.}}}%
	=}

\newcommand{\cY}{\mathcal{Y}}

\newcommand{\sebcom}[1]{}

\usepackage[accepted]{aistats2025}
\usepackage[round]{natbib}

\begin{document}



\twocolumn[

\aistatstitle{Random Forest Calibration}

\aistatsauthor{ Mohammad Hossein Shaker \And Eyke H\"ullermeier}

\aistatsaddress{ 

Institute of Informatics, LMU Munich Munich\\ 
Center for Machine Learning (MCML)\\
Germany 

\And  

Institute of Informatics, LMU Munich\\
Munich Center for Machine Learning (MCML)\\
Germany} ]

\begin{abstract}
    % The Random Forest (RF) classifier is often considered well-calibrated compared to other machine learning methods. However, literature suggests that traditional calibration methods like isotonic regression offer little improvement to RF probability estimates unless large calibration datasets are available, which can be challenging in data-scarce scenarios. Despite these claims, there is no comprehensive study specifically comparing state-of-the-art calibration methods for RF. To address this, we investigate a wide range of calibration techniques applicable to RF, from scaling methods to advanced algorithms. Our experiments on synthetic and real-world data examine RF probability estimates, the influence of hyper-parameters, systematically compare calibration methods, and show that a well-optimized RF performs as well as or better than state-of-the-art calibration methods. 

    The Random Forest (RF) classifier is often claimed to be relatively well calibrated when compared with other machine learning methods. Moreover, the existing literature suggests that traditional calibration methods, such as isotonic regression, do not substantially enhance the calibration of RF probability estimates unless supplied with extensive calibration data sets, which can represent a significant obstacle in cases of limited data availability. Nevertheless, there seems to be no comprehensive study validating such claims and systematically comparing state-of-the-art calibration methods specifically for RF.  To close this gap, we investigate a broad spectrum of calibration methods tailored to or at least applicable to RF, ranging from scaling techniques to more advanced algorithms. Our results based on synthetic as well as real-world data unravel the intricacies of RF probability estimates, scrutinize the impacts of hyper-parameters,  compare calibration methods in a systematic way. We show that a well-optimized RF performs as well as or better than leading calibration approaches. 

\end{abstract}

\section{Introduction}

The Random Forest (RF) classifier \citep{breiman2001random} is a versatile machine learning (ML) algorithm, which is easy to use and proved to be fast, robust, and extremely competitive across a broad range of applications.

Going beyond mere class assignments, it is also able to predict probabilities on the basis of relative class frequencies in a rather natural way. Probabilistic predictions are particularly desirable in safety-critical applications, such as medical diagnoses \citep{khalilia2011predicting}, financial predictions \citep{khaidem2016predicting}, and risk assessments \citep{wang2015flood, yu2022new}, where information about the confidence in a prediction is of utmost importance. 

\begin{figure}
\begin{center}

  \includegraphics[width=0.8\columnwidth]{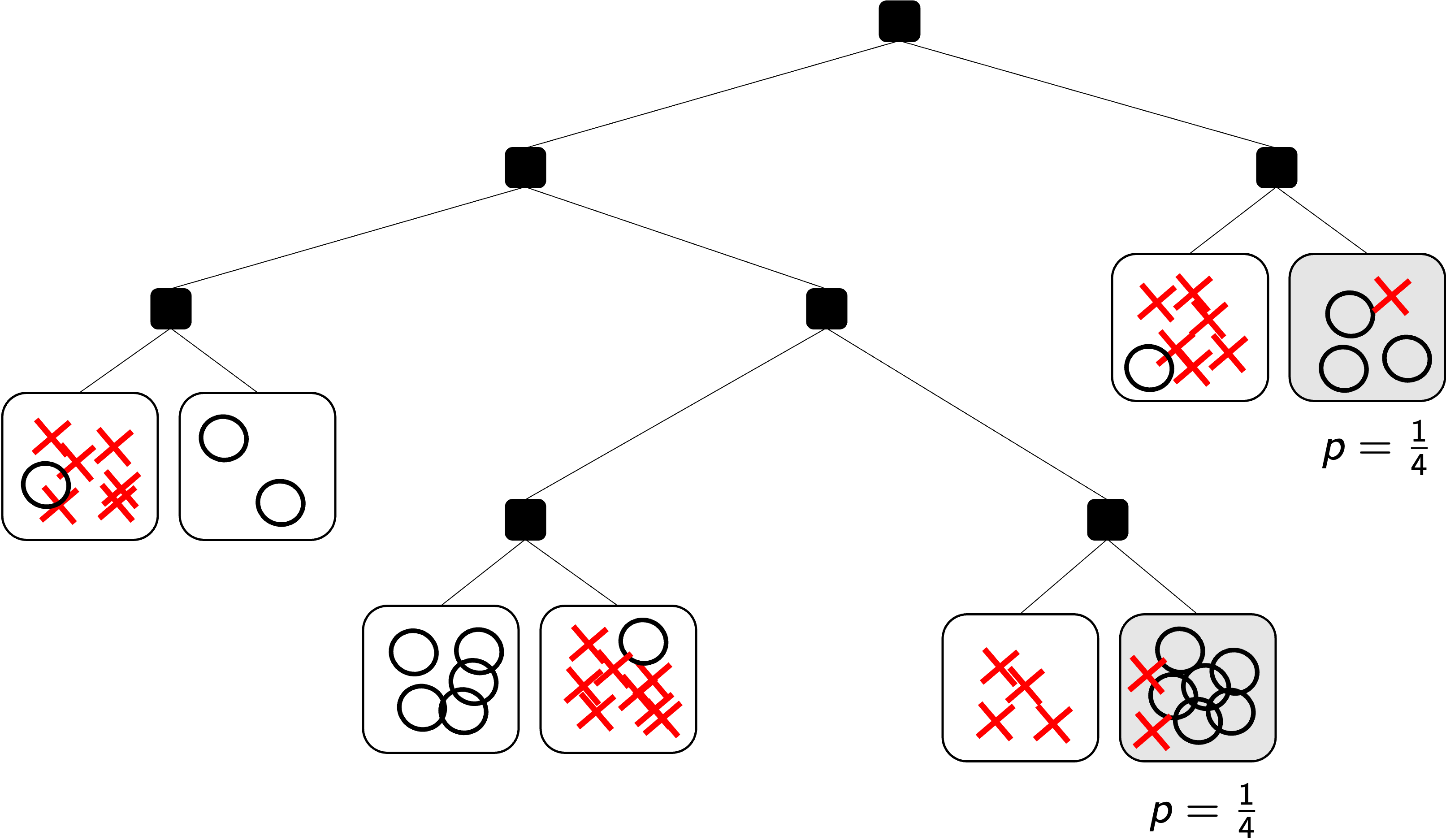}

\caption{This decision tree estimates the probability of the positive class as $1/4$ for all instances falling in either of the two leaf nodes shaded in grey. Class-wise calibration then requires that $\mathbb{P}(Y = +1 \given \vec{x} \in A \cup B) = 1/4$, where $A, B \subset \mathcal{X}$ are the regions in the instance space associated with the two nodes, respectively. Note that this neither implies instance-wise calibration nor ``leaf-wise'' calibration (i.e., $\mathbb{P}(Y = +1 \given \vec{x} \in A) = 1/4$ and $\mathbb{P}(Y = +1 \given \vec{x} \in  B) = 1/4$).}
\label{fig:treeexample}
\end{center}
\end{figure}

The reliability of probability estimates, whether coming from an RF or any other ML algorithm, is becoming a subject of growing concern, however, and the question arises how trustworthy such estimates actually are. In machine learning, the notion of \emph{calibration} is used to characterize this trustworthiness: a predictor is called well-calibrated if its predicted probabilities align well with the ground-truth probabilities. The topic of calibration has attracted increasing attention in machine learning in the recent past, and various calibration methods have been proposed \citep{silv_cc23}. Typically, these methods seek to improve probability estimates in a post-processing step by learning a function that maps estimates to ``better'' estimates.

RF is often claimed to be relatively well calibrated, especially compared to other machine learning methods. Moreover, the existing literature suggests that traditional calibration methods, such as isotonic regression, do not substantially enhance the calibration of RF probability estimates unless supplied with extensive calibration data sets, which can represent a significant obstacle in cases of limited data availability. Nevertheless, there seems to be no comprehensive study validating such claims and systematically comparing state-of-the-art calibration methods specifically for RF. In this paper, we aim to address this gap. The key contributions of our work are as follows:

% \paragraph{Contribution.}
\begin{enumerate}
    \item [\textbf{1.}] We conduct extensive experimental studies using synthetic data with known true probabilities. These studies aim to provide insights into the behavior of RF probability estimates with regard to \emph{instance-wise} and \emph{probability-wise} calibration metrics.
    \item [\textbf{2.}] We examine the influence of RF hyper-parameters on calibration performance, with the aim of improving theoretical understanding and providing a comprehensive guide for researchers, data scientists, and machine learning practitioners working with RF calibration.
    \item [\textbf{3.}] We investigate a broad spectrum of calibration methods, including traditional ones but also more recent proposals, model-agnostic methods as well as methods specifically tailored to RF for a systematic comparison of state-of-the-art calibration methods.
\end{enumerate}

% Sum of the paper
% The paper is organized as follows:
% Section \ref{sec:cal} defines calibration formally and explores the concept in detail.  
% Section \ref{sec:eval} presents the evaluation metrics used to assess calibration performance.  
% Section \ref{sec:exp} focuses on our experimental studies, comparing calibration methods on real and synthetic data.  
% Finally, Section \ref{sec:conc} concludes the paper.

\section{Calibration}
\label{sec:cal}
% supervised setting

In the context of supervised learning, we consider a standard scenario where a learner is provided with a set of training data denoted $\mathcal{D} \defeq \{ (\vec{x}_i , y_i )\}_{i=1}^N \subset \mathcal{X} \times \mathcal{Y}$, where $\mathcal{X}$ represents the instance space and $\mathcal{Y}$ denotes the possible outcomes associated with each instance. Specifically, we focus on classification and let $\cY=\{1, \ldots, K\}$ denote a finite set of $K$ class labels, with binary classification ($K=2$) being a notable special case (for which we let $\cY = \{ 0,1 \}$ instead of $\cY = \{ 1, 2 \}$). The data $\mathcal{D}$ is commonly assumed to be generated according to an underlying (yet unknown) probability distribution $Q$ on $\mathcal{X} \times \mathcal{Y}$, i.e., the data points $(\vec{x}_i , y_i )$ are realization of random variables $(\vec{X} , Y) \sim Q$ sampled (independently) from $Q$.

The joint distribution $Q$ induces marginal distributions $Q_\mathcal{X}$ and $Q_\mathcal{Y}$ on $\mathcal{X}$ and $\mathcal{Y}$, respectively, as well as different conditional distributions. Here, we are specifically interested in conditional distributions $Q( \cdot \given \vec{x})$ on $\mathcal{Y}$, where $Q( j \given \vec{x})$ is the probability of class $j$ given $\vec{x}$. Since $\mathcal{Y}$ is a finite set in our case, $Q( \cdot \given \vec{x})$ is a categorical distribution that can be represented in the form of a probability vector 
\begin{equation}\label{eq:gtp}
\vec{q}(\vec{x}) = (q_1(\vec{x}), \ldots , q_K(\vec{x})) \in \Delta_K \, , 
\end{equation}
where $q_j(\vec{x}) = Q( j \given \vec{x})$ and $\Delta_K \subset [0,1]^K$ denotes the $(K-1)$-simplex. In other words, with each instance $\vec{x} \in \mathcal{X}$, we can associate a ground-truth probability distribution (\ref{eq:gtp}) on $\mathcal{Y}$. In the following, if clear from the context, we will often omit the instance $\vec{x}$ and simply write $\vec{q} = (q_1, \ldots , q_K)$ for a ground-truth distribution of interest.

Distributions $\vec{q}(\vec{x})$ constitute the key targets in probabilistic machine learning. A predictor in the form of a class probability estimator is a function $f: \, \mathcal{X} \fromto \Delta_K$ that maps instances to probability distributions over outcomes. A prediction $\vec{p} = (p_1, \ldots , p_K) = f(\vec{x})$ is considered as an estimate of the true (conditional) distribution $\vec{q} = (q_1, \ldots , q_K) = Q(\cdot \given \vec{x})$. Ideally, the predictions $f(\vec{x})$ match well with the ground-truth probabilities $\vec{q}(\vec{x})$, i.e., $f(\vec{x}) \approx \vec{q}(\vec{x})$ for all $\vec{x} \in \mathcal{X}$. In that case, we say that the predictor $f$ is \emph{instance-wise} calibrated or that calibration is \emph{per-instance}.

For statistical reasons, however, instance-wise calibration is very difficult to achieve in practice \citep{foyg_tl21}. Therefore, the common notion of calibration, which originated in forecasting \citep{hall_fp20,murp_ro77,dawi_tw82} and is now also adopted in machine learning \cite{silv_cc23}, is less demanding. It refers to specific conditional distributions, specifying the probability of class observations given certain events (predictions produced by $f$). For these probabilities, which (for a fixed predictor $f$) all derive from the underlying data-generating process $Q$, we will subsequently use the generic notation $\mathbb{P}$. 

Let $\mathcal{P}(f) \subseteq \Delta_K$ denote the set of all probability estimates $\vec{p} = (p_1, \ldots , p_K)$ that can be produced by the predictor $f$. The standard notion of calibration, also known as \emph{class-wise calibration}, can then be defined as follows:  
A predictor $f$ is calibrated if the following equality holds for all $j \in \{ 1,\ldots,K\}$ and all predictions $\vec{p} = (p_1, \ldots , p_K) \in \mathcal{P}(f)$: 
\begin{equation}
    \mathbb{P}[ Y = j \given P_j = p_j  ] = p_j \, .
    \label{eq:ccalib}
\end{equation}
Two random variables are involved in this definition: $Y$ is the observed class label and $P_j$ the predicted probability for class $j$ (both are indeed random, assuming that instances $\vec{x}$ are chosen at random). The condition (\ref{eq:ccalib}) then means the following: Given that the predicted probability for class $j$ is $p_j$, the probability that $j$ occurs is indeed $p_j$. 

Please note that, unlike instance-wise calibration, (\ref{eq:ccalib}) does no longer condition on individual instances $\vec{x}$. Instead, class occurrences are conditioned on predictions made by $f$. Broadly speaking, all instances $\vec{x}$ for which $f$ predicts the same probability $p_j$ are ``grouped'' together, even if the true probabilities for class $j$ may differ. Then, (\ref{eq:ccalib}) merely requires that the average probability of class $j$ within such a group is indeed $p_j$. For example, if probabilities are estimated through relative class frequencies in the leaf nodes of a decision tree, then all instances falling in the same leaf node are grouped together, and possibly even instances from different nodes with the same probability (cf.\ Fig.\ \ref{fig:treeexample}). Since conditioning is now done on (predicted) probabilities instead of instances $\vec{x}$, we subsequently refer to this notion of calibration as \emph{probability-wise calibration}. 

% Fig \label{fig:treeexample} original place

Obviously, class-wise calibration is a relatively weak property. For example, a predictor that completely ignores context information provided by $\vec{x}$ and constantly predicts the marginal distribution on $\mathcal{Y}$ (and hence puts all instances in a single group), i.e., $f(\vec{x}) \equiv Q_\mathcal{Y}$, is calibrated according to (\ref{eq:ccalib}). Clearly, a predictor that always predicts, say, a 60\% chance for a home win, regardless of the teams playing against each other, is not very useful, even if the home team is indeed winning in 60\% of the cases on average.

A slightly more stringent version of probability-wise calibration is \emph{multiclass calibration}, which considers all class probabilities and their predictions \emph{simultaneously}: A predictor $f$ is calibrated if, for all $j \in \{ 1,\ldots,K\}$ and all predictions $\vec{p} \in \mathcal{P}(f)$, 
\begin{equation}
    \mathbb{P}[ Y = j \given \vec{P} = \vec{p}  ] = p_j \, .
    \label{eq:mcalib}
\end{equation}
Technically, the difference is that (\ref{eq:mcalib}) conditions on the entire vector of predicted probabilities (the random variable $\vec{P}$), whereas (\ref{eq:ccalib}) only conditions on its $j^{th}$ component. Obviously, multiclass calibration implies class-wise calibration, but not the other way around. An exception is the case of binary classification, where both definitions coincide.

Although this is less important for the purpose of our study, let us mention that further definitions of calibration can be found in the literature. For example, so-called \emph{confidence-calibration} \citep{guo2017calibration} merely requires calibration (in the above sense) for the top-label, i.e., the class with the highest predicted probability. This has been generalized by \cite{gupta2021top} to \emph{top-label calibration}, which, broadly speaking, requires confidence-calibration for each class label separately.  

With regards to the calibration methods used in this study, we employed a range of post-calibration techniques suitable for RF models. These methods include Platt Scaling, Beta Calibration, Isotonic Regression (ISO), Venn-Abers (VA), Parameterized Probability Adjustment (PPA), Curtailment (CT), and the RF rank calibrator (Rank). A comprehensive description of each calibration approach is provided in Appendix~\ref{sec:cm}.

\section{Evaluation Metrics of Calibration}
\label{sec:eval}

% the difference between probability distributions from a machine learning model and the aforementioned probability vectors.
%Acknowledging that obtaining complete information about both $Q$ and $C$ is often unfeasible in real-world scenarios, this section presents various metrics used to gauge calibration performance. It's important to highlight that two distinct measurements are being explored here. 

In accordance with the distinction between per-instance and probability-wise calibration as discussed in the previous section, two types of measures for evaluating calibration performance can be found in the literature. In this section, we give an overview of commonly used measures of that kind, starting with the per-instance case.

As observational data only provides class labels as outcomes, but no (true) probabilities, predicted probabilities $\vec{p} \in \Delta_K$ are often compared to the observed classes directly. The latter are then typically treated as degenerate distributions, assigning probability 1 to the observed class and 0 to all others. In the following, we denote by $\vec{q}^y =(q^y_1, \ldots , q^y_K ) \in \{0,1\}^K$ the (degenerate) probability vector associated with an observed class $y$, where the $j^{th}$ entry $q^y_j$ is $1$ if $y = j$ and $0$ otherwise.

%The first aims to approximate the discrepancy between the predicted probability distributions and the actual posterior probability vector $Q$ on an instance-by-instance basis. 

%The second seeks to estimate the difference between the predicted probability distributions and the group-wise calibrated probability vector $C$. Thus, in the subsequent sections, we will classify the evaluation metrics into Instance-wise and Group-wise metrics.

\subsection{Instance-wise Metrics}
\label{sec:instance-wise_m}

The \textbf{true calibration error} (TCE)  measures the calibration of a predicted probability $\vec{p}=(p_1, \ldots , p_K)$ in terms of its deviation from the true distribution $\vec{q}$. To this end, any metric or measure of divergence between probability distributions can in principle be used, such as the mean squared error (MSE):
\begin{equation}
\text{TCE}(\vec{p},\vec{q}) = \sum_{j=1}^{K}  (p_j - q_j)^2 \, .
\end{equation}
Note that TCE normally cannot be computed in practice, as it requires knowledge of the true distribution $\vec{q}$. This information might be available for synthetic data but not for empirical data. 

Replacing the true distribution $\vec{q} \in \Delta_K$ by an observed class label $y$ (or the associated distribution $\vec{q}^y$) yields the \textbf{Brier score} \citep{brier1950verification}:
\begin{equation}
   \text{BS}(\vec{p}, y) =  \text{BS}(\vec{p}, \vec{q}^y) = \sum_{j=1}^K (p_j - q^y_j)^2 \, .
\end{equation}
Another common measure for comparing a predicted distribution with an observed class label is the \textbf{logistic loss} (log-loss) or cross-entropy loss \citep{shannon1948mathematical}:
\begin{equation}
   \text{LL}(\vec{p}, y) =  \text{LL}(\vec{p}, \vec{q}^y) = - \log ( p_y ) \, .
\end{equation}

Both the log-loss and the Brier score are special cases of (strictly) \emph{proper scoring rules}. Consider a loss function $\phi: \Delta_K \times \cY \fromto \mathbb{R}$ and denote the  expected loss of a prediction $\vec{p}$ with respect to a (ground-truth) distribution $\vec{q}$ by
\begin{equation*}
    s( \vec{p}, \vec{q}):=\mathbb{E}_{Y \sim \vec{q}} \phi( \vec{p}, Y)=\sum_{j=1}^K \phi\left(\vec{p},  j \right) q_j \, .
\end{equation*}
The scoring rule $\phi$ is called proper if $\vec{p} \mapsto s( \vec{p}, \vec{q})$ is minimized for $\vec{p} = \vec{q}$ and strictly proper if the minimizer is unique for all $\vec{q} \in \Delta_K$. In other words, a (strictly) proper scoring rule urges a risk-minimizing learner to predicting the true probabilities.

The irreducible part (IL) of the (expected) loss, $e(\vec{q}) = s(\vec{q}, \vec{q})$ is also called the \emph{entropy} of $\vec{q}$, whereas $d(\vec{p}, \vec{q}) = s(\vec{p}, \vec{q}) - s(\vec{q}, \vec{q})$ is called the \emph{divergence} of $\vec{p}$ from $\vec{q}$. For the log-loss, $d$ is KL-divergence and $e$ is Shannon entropy, whereas for Brier score, $d$ is mean squared difference and $e$ is the Gini index. Interestingly, for any strictly proper scoring rule, the divergence can be further decomposed into a calibration loss (CL) and a grouping loss (GL) \citet{kull2015novel}:\footnote{Note that $d( \vec{p}, y) =  d( \vec{p}, \vec{q}^y) = s( \vec{p}, \vec{q}^y)$, because $s( \vec{q}^y, \vec{q}^y) = 0$.}
\begin{equation}
\begin{split}
&\mathbb{E}_{Q}[ \, d( \vec{p}(\vec{x}), y) \, ] = 
\underbrace{\mathbb{E}_{Q_\mathcal{X}} [ \, d(\vec{p}(\vec{x}), \vec{c}(\vec{x})) \, ]}_{\text{CL}} \\
&+ \underbrace{\mathbb{E}_{Q_\mathcal{X}} [ \, d(\vec{c}(\vec{x}), \vec{q}(\vec{x})) ]}_{\text{GL}} 
+ \underbrace{\mathbb{E}_{Q} [\, d(\vec{q}(\vec{x}), y) \, ]}_{e(\vec{q})} \, ,
\end{split}
\label{eq:decomp_CGI}
\end{equation}

where $\vec{c}$ is the vector of calibrated probabilities (\ref{eq:mcalib}), i.e., the entries of which are given by the class probabilities  conditioned on the prediction $\vec{p} = \vec{p}(\vec{x})$. 

\begin{figure}
\begin{center}

  \includegraphics[width=\columnwidth]{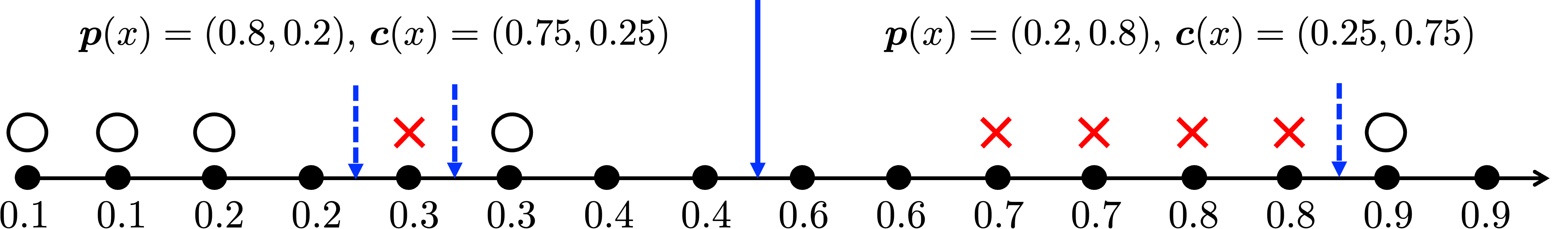}

\caption{Illustration of loss decomposition for binary classification and a finite one-dimensional instance space $\mathcal{X}$ (the black points). The numbers at the bottom indicate the true probabilities of the positive class. The blue arrow indicates a split of the data into two groups, leading to probability estimates $\vec{p}(x) = (0.8, 0.2)$ for the left and $\vec{p}(x) = (0.2, 0.8)$ for the right group, based on the given training data (positive and negative class indicated by red crosses and black circles, respectively). }
\label{fig:split}
\end{center}
\end{figure}

See Fig.\ \ref{fig:split} for an illustration with a finite one-dimensional instance space $\mathcal{X}$ (the black points) and a small data set consisting of a few positive (red crosses) and a few negative (black circles) examples. The true probabilities of the positive class are shown at the bottom (e.g., $\vec{q}(x) = (0.9, 0.1)$ for the left-most point). Now, imagine that the data is split into two parts (blue arrow in the moddle), say, by a decision tree learner, and probabilities for the two groups thus created are estimated by relative frequencies. Thus, $\vec{p}(x) = (0.8, 0.2)$ for all points on the left and $\vec{p}(x) = (0.2, 0.8)$ for all points on the right. Assuming a uniform distribution on $\mathcal{X}$, the calibrated distributions are given by $\vec{c}(x) = (0.25, 0.75)$ and $\vec{c}(x) = (0.75, 0.25)$ for the group on the left and on the right, respectively. For the Brier score as a loss, (\ref{eq:decomp_CGI}) yields the decomposition $0.38 = 0.005 + 0.025 +  0.35$ of the expected loss into CL, GL, and entropy (in this case Gini).

Note that the above decomposition is particularly interesting from a decision tree (and hence RF) learning point of view. According to (\ref{eq:decomp_CGI}), minimizing the expected loss comes down to minimizing the sum of calibration and grouping loss (as entropy only depends on the ground-truth $\vec{q}$ and thus cannot be influenced by the prediction $\vec{p}$). The grouping loss compares $\vec{c}$ with $\vec{q}$. Obviously, the more fine-granular the partition of $\mathcal{X}$, the smaller this loss tends to be, because the better the ``average'' probability $\vec{c}$ within a group (leaf node of a tree) approximates the probabilities $\vec{q}(\vec{x})$ for the instances $\vec{x}$ in the group. Thus, minimizing GL requires sufficiently large tress with enough leaf nodes. The calibration loss, on the other side, compares $\vec{p}$ and $\vec{c}$, and tends to be lower for coarse partitions: $\vec{p}$ estimates $\vec{c}$ through relative frequencies, and the more observations are available, the better this estimate will be. Overall, this means that loss minimization comes down to finding an optimal balance between CL and GL, which, for the case of decision trees, translates into finding the right complexity of the tree. In our example, the simplest tree, which consists of a single node and puts all instances in the same group, has a CL of 0 (the calibrated probability $\vec{c} = (0.5, 0.5)$ is exactly reflected in the training data), but a high GL of 0.15, yielding an overall expected loss of 0.5. A complex tree that is split to purity may induce the partition indicated by the additional dashed arrows in Fig.\ \ref{fig:split}. In this case, GL is 0.12 and CL increases to 0.29, giving an even higher overall expected loss of 0.75.

As an interesting side remark, note that increasing the size of a tree does not necessarily lead to a refinement of the grouping induced by the corresponding predictor $f$, i.e., an increase of $|\mathcal{P}(f)|$. As already said, a group does not necessarily correspond to a leaf node of a tree (or, more precisely, the region in the instance space associated with that leaf node). Instead, it consists of the union of all leaf nodes with the same predicted probability (cf.\ Fig.\ \ref{fig:treeexample}), and the smaller the nodes become (in terms of the number of training examples covered), the higher the chance that these probabilities coincide.\footnote{This chance is lowered if probabilities are estimated with Laplace correction, which effectively means that leaf nodes are not only grouped based on class frequencies but the number of examples covered.} In the last example above, only extreme probabilities of 0 and 1 are predicted, whence we again end up with only two groups (which, in particular due to the negative ``outlier'' on the right, are even much worse than those obtained with a single split).

\subsection{Probability-wise Metrics}

%In contrast to the fine-grained analysis provided by instance-wise metrics, group-wise metrics present an alternative approach to evaluating calibration in machine learning models. Due to the absence of true probability vectors necessary for calculating instance-wise metrics like TCE, group-wise metrics are introduced to tackle this limitation. This subsection explores calibration assessment through aggregated statistics, where metrics are calculated based on groups of instances rather than individual data points. Group-wise metrics provides insights into the collective behavior of predictions within specific subsets of the dataset.

As already said, probability-wise metrics directly lend themselves to the definition of calibration as introduced in Section \ref{sec:cal}. A natural way to check the condition (\ref{eq:mcalib}) empirically on given set of (validation) data is to group all data points for which the predictor estimates the same probability $\vec{p}$, and to compare this $\vec{p}$ with the relative frequency distribution in this group. If the predictor is calibrated, the divergence between these two distributions should be low. However, an obvious problem of this approach is the possibly small size of the groups if this condition is checked for all $\vec{p} \in \mathcal{P}(f)$ separately. The relative frequency distribution may then constitute a poor estimate of the true conditional probability (\ref{eq:mcalib}).

A possible way out is to follow the same idea that underlies the approximation of a density function in terms of a histogram, namely, to replace points by intervals or \emph{bins}. This leads to the well-known \emph{expected calibration error} (ECE), which proceeds from a binning $\mathcal{B}$ of the unit interval \citep{naeini2015obtaining}. The latter is a finite collection of $M = |\mathcal{B}|$ bins $B$, each associated with an interval $I_B \subset [0,1]$, so that $\{ I_B \given B \in \mathcal{B} \}$ forms a partition of $[0,1]$. Given a set $\mathcal{D}$ of $N$ data points and predictor $f$, let $p_j(\vec{x}_i)$ be the probability predicted by $f$ for a fixed class $j$. Then, this points falls into bin $B \in \mathcal{B}$, i.e., index $i$ is added to bin $B$, if $p_j(\vec{x}_i) \in I_B$. The ECE for class $j$ is then defined as follows:
\begin{equation}
\text{ECE}_j(\mathcal{D}, f)  = \sum_{B \in \mathcal{B}}  \frac{|B|}{N} |  \bar{p}_{j,B} - p_{j,B} | \, ,
\end{equation}
where $\bar{p}_{j,B}$ denotes the relative frequency of class $j$ in bin $B$, and $p_{j,B}$ the average predicted probability:
\begin{align*}
   \bar{p}_{j,B} & = \frac{1}{|B|} \cdot | \{ i \in B \given y_i = j  \} | \\
       \bar{p}_{j,B} & = \frac{1}{|B|} \sum_{i \in B} p_{j}(\vec{x}_i) \, .
\end{align*}
The overall ECE is then defined by averaging over classes:
\begin{equation}
\text{ECE}(\mathcal{D}, f)  = \frac{1}{K} \sum_{j=1}^K \text{ECE}_j(\mathcal{D}, f) \, .
\end{equation}
Note that the computation of ECE involves another grouping of data points. According to the definition of calibration, instances $\vec{x}$ and $\vec{x}'$ are grouped if being mapped to the same probability by the predictor $f$. Now, they are not only grouped in the case of exact equality of predictions, but as soon as the predicted probabilities are sufficiently similar in the sense of falling into the same bin.  

% where $p$ is the predicted probability from the machine learning model. 
The ECE is known to be quite sensitive with regard to the number of bins $M$ \citep{roelofs2022mitigating}. For example, if $M$ is too high, the number of instances per bin might be very low, resulting in poor estimates $\bar{p}_{j,B}$. Additionally, an experiment by \citet{gruber2022better} shows the impact of data size on the value of ECE for a synthetic version of three models that are representations for a perfect, mediocre, and poorly calibrated model. Surprisingly, under low data size, the perfectly calibrated model can have higher calibration error than the mediocre model.

Let us conclude this section with a note on the classification rate as a standard measure of accuracy. The classification rate of a machine learning model may not directly indicate a well-calibrated model. Indeed, a model can perform well in terms of classification rate while producing over-confident probability distributions. The other way around, a predictor can be well calibrated while performing poorly as a classifier (e.g., a predictor that always forecasts the true class prior is perfectly calibrated but degenerates to the majority classifier). Therefore, to have a comprehensive view, we opt for considering the classification rate of a machine learning model in addition to its calibration performance, especially in the case of probability-wise evaluation metrics such as ECE. Ideally, a high classification rate and small calibration loss can be achieved at the same time.

\section{Related Work}
\label{sec:rw}

% from the Experiment section - 
\begin{figure*}[ht]
\begin{center}

  \subfloat{\includegraphics[scale=0.27]{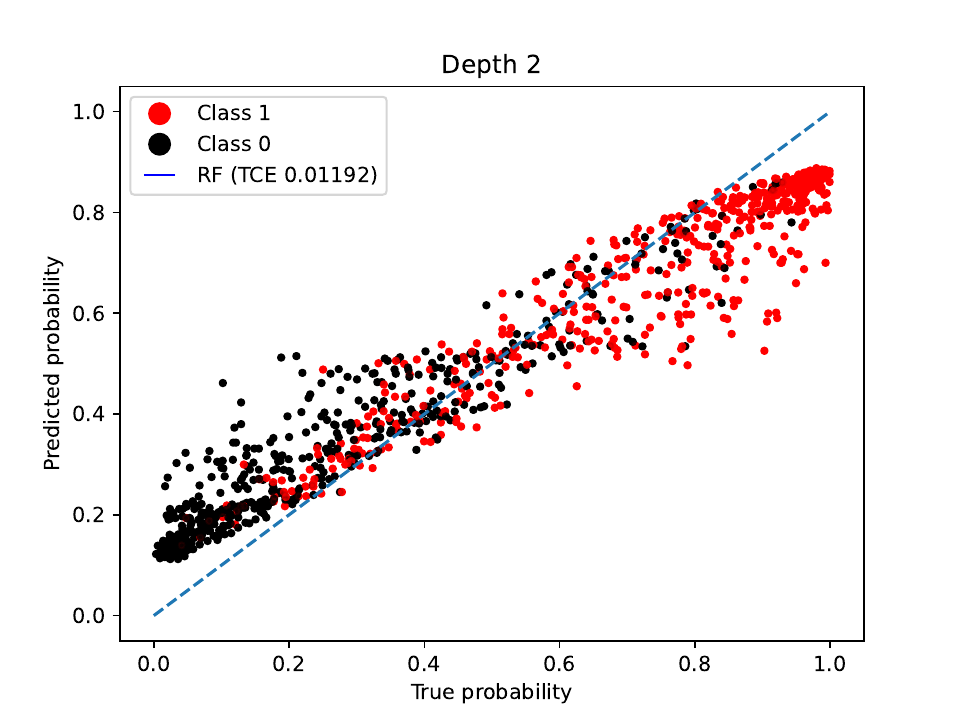}}
  \subfloat{\includegraphics[scale=0.27]{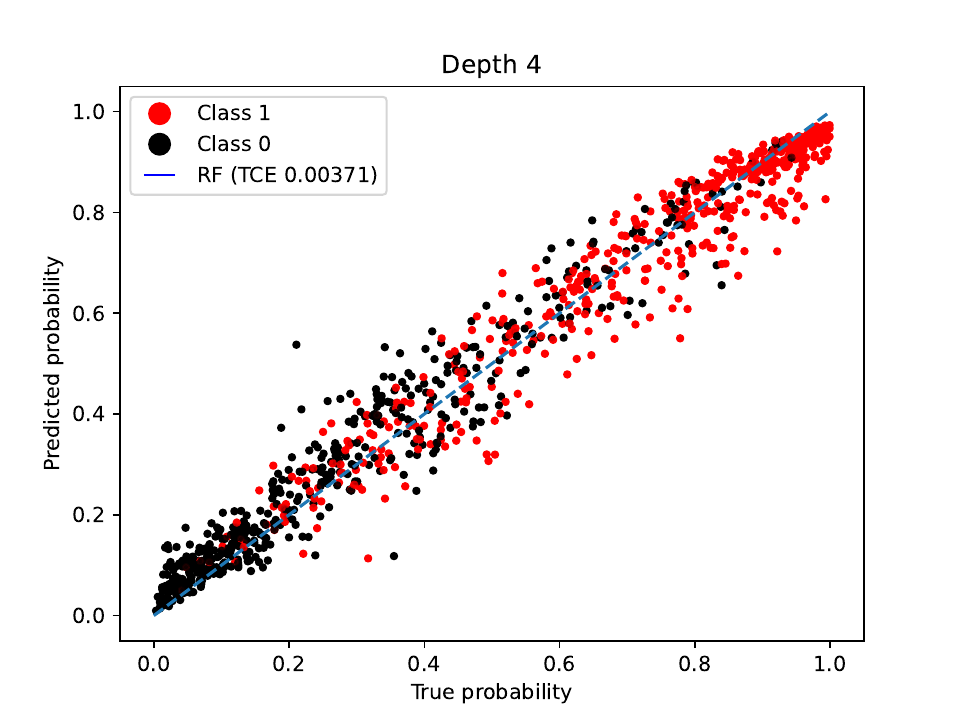}}
  \subfloat{\includegraphics[scale=0.27]{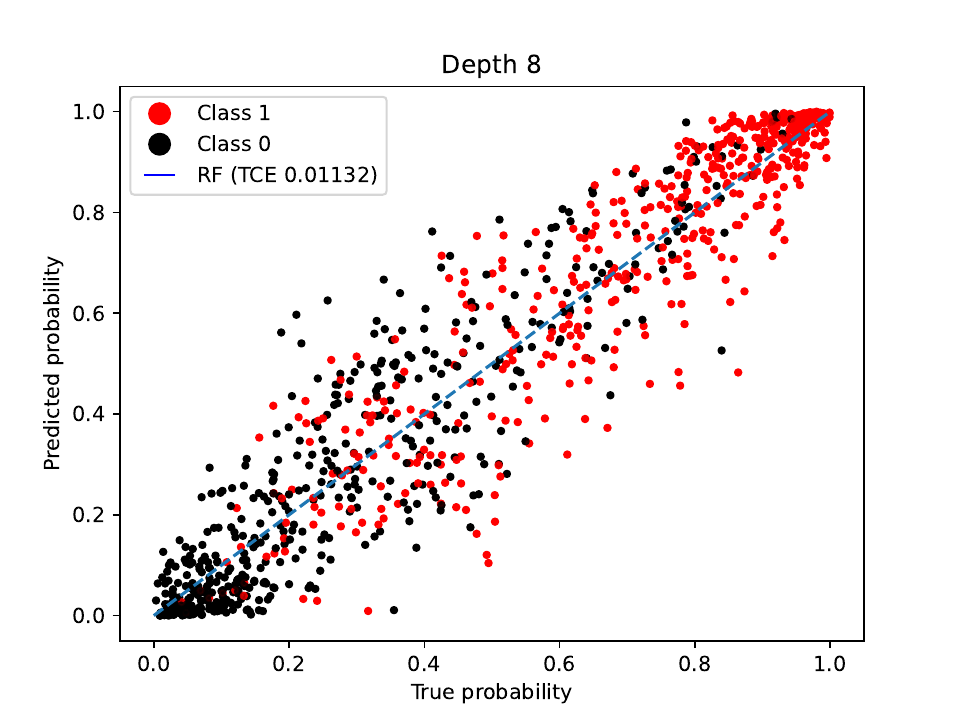}}

\caption{The effect of setting max-depth of trees in an RF on calibration performance on synthetic data. The reliability diagram represents RF with low (left), optimal (middle), and too high (right) value for parameter max-depth.}
\label{fig:depth}
\end{center}
\end{figure*}

\citet{zadrozny2001obtaining} identify two key issues with probability estimates in decision trees: (1) a bias towards extreme probabilities (0 or 1), caused by the search for pure leaf nodes, and (2) high variance from fragmentation, where low sample counts in leaf nodes lead to unstable estimates.

Pruning \citep{esposito1997comparative} addresses the high variance but struggles with imbalanced data. \citet{zadrozny2001obtaining} propose "curtailment," which prunes leaf nodes with insufficient training samples to improve calibration.

Another method is to smooth probability estimates using Laplace correction or m-estimation \citep{provost2000well}. Smoothing reduces extreme probabilities, addressing the first issue. Empirical evaluations show that combining curtailment and smoothing is most effective, with bagged curtailment (similar to RF) yielding the best results. However, \citet{bostrom2007estimating} notes that both Laplace and m-estimation can negatively impact RF accuracy, AUC, and Brier score.

We introduce the RF model in detail in appendix \ref{sec:rf}.
Regarding RF calibration, an important question is which type of decision tree is best as a base learner: classifier trees (CT) or probability estimation trees (PET). \citet{bostrom2008calibrating} found that CT provides better calibration (measured by the Brier score), while PET offers superior accuracy and AUC.

Another question is whether calibration should occur at the individual tree level or for the entire ensemble. \citet{wu2021should} and \citet{rahaman2021uncertainty} examined this and found that calibrating individual trees before aggregation leads to worse final calibration, sometimes even worse than an uncalibrated RF.

% A third question concerns the best data for calibrating an RF model. Options include:

% \begin{itemize}

% \item \textbf{Training data}: The simplest choice, but it can introduce bias since the same data is used for both training and calibration.
% \item \textbf{Separate calibration set}: Reserving part of the training data for calibration avoids bias but reduces data for both training and calibration.
% \item \textbf{Cross-validation}: More resource-intensive, cross-validation allows each example to be calibrated using an independent classifier.
% \item \textbf{Out-of-bag (OOB) data}: Bootstrap resampling generates OOB data, which can be used for calibration without bias.
% \end{itemize}

A third question concerns the best data for calibrating an RF model. There are several approaches to calibration in machine learning models. One option is to use the training data itself, which is the simplest choice, but this can introduce bias since the same data is used for both training and calibration. Another approach is to reserve a separate calibration set, where part of the training data is set aside for calibration. This method avoids bias but reduces the amount of data available for both training and calibration. Cross-validation is a more resource-intensive option that allows each example to be calibrated using an independent classifier. Lastly, bootstrap resampling can be used to generate out-of-bag (OOB) data, which provides a way to calibrate models without introducing bias.

\citet{johansson2019efficient} compared using a separate calibration set versus OOB data and found that OOB data generally outperformed the independent calibration set across experiments, despite being limited to only four calibration methods.

\section{Experiments}
\label{sec:exp}

\begin{figure*}[ht]
    \begin{center}
    
      % \subfloat{\includegraphics[scale=0.22]{images/o2D_acc.pdf}}
      % \subfloat{\includegraphics[scale=0.22]{images/o5D_acc.pdf}}
      % \subfloat{\includegraphics[scale=0.22]{images/o10D_acc.pdf}}
      % \subfloat{\includegraphics[scale=0.22]{images/o20D_acc.pdf}}

      % \subfloat{\includegraphics[scale=0.22]{images/o2D_brier.pdf}}
      % \subfloat{\includegraphics[scale=0.22]{images/o5D_brier.pdf}}
      % \subfloat{\includegraphics[scale=0.22]{images/o10D_brier.pdf}}
      % \subfloat{\includegraphics[scale=0.22]{images/o20D_brier.pdf}}
    
      \subfloat{\includegraphics[scale=0.27]{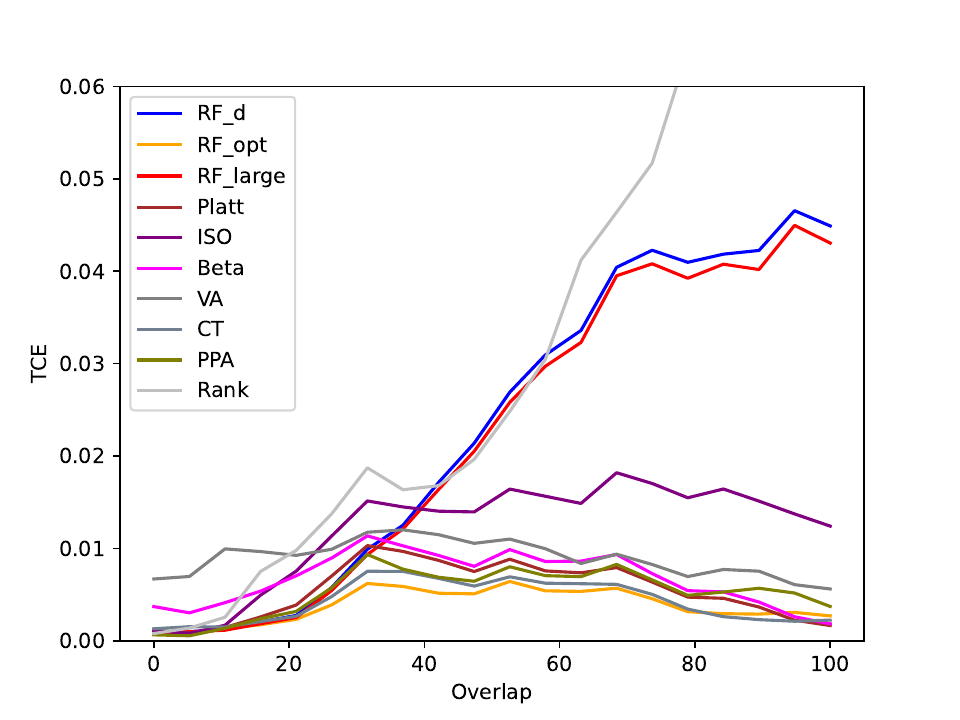}}
      \subfloat{\includegraphics[scale=0.27]{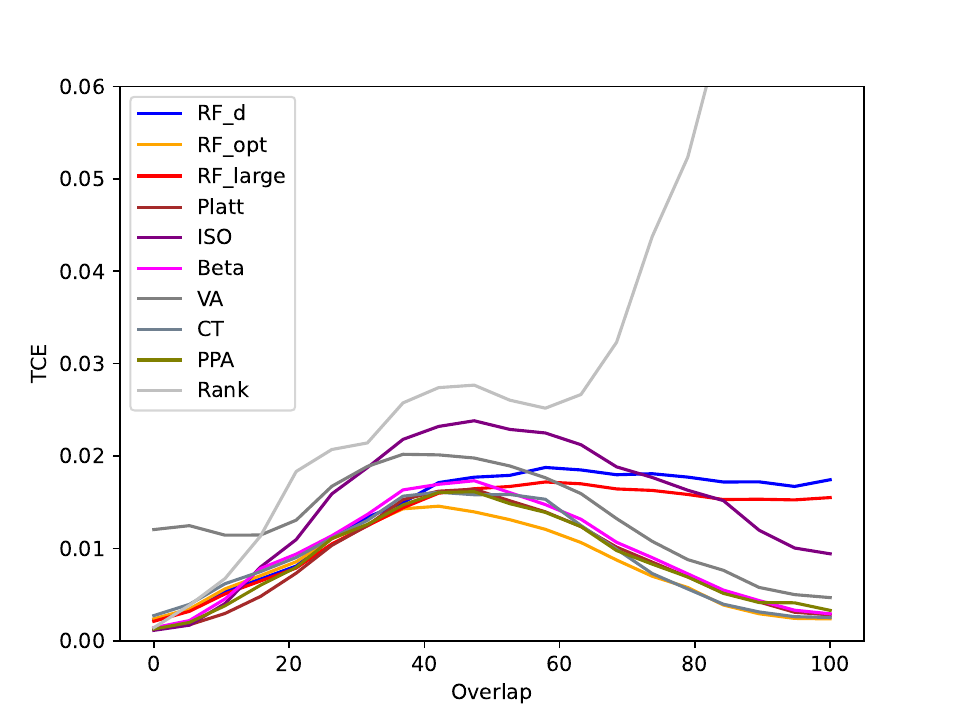}}
      \subfloat{\includegraphics[scale=0.27]{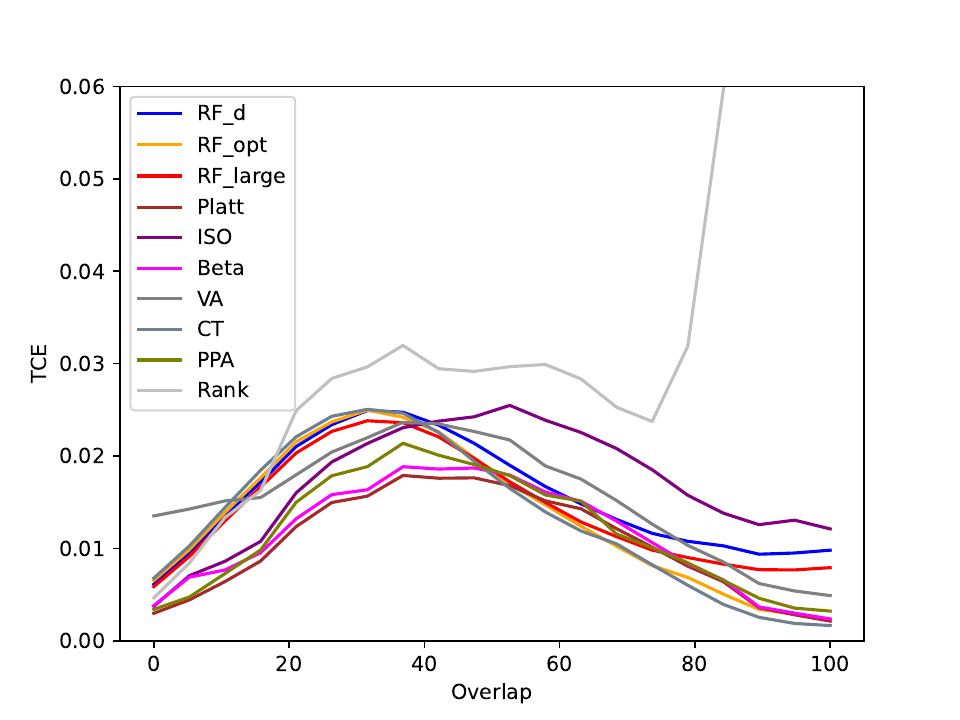}}
      \subfloat{\includegraphics[scale=0.27]{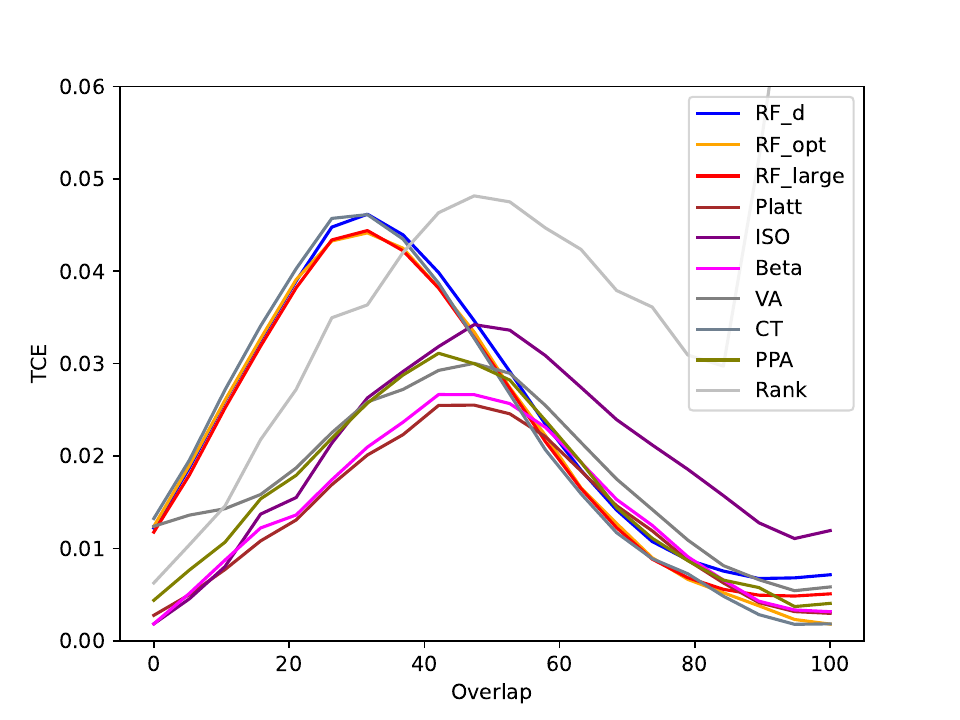}}
    
      % \subfloat{\includegraphics[scale=0.22]{images/o2D_CLGL.pdf}}
      % \subfloat{\includegraphics[scale=0.22]{images/o5D_CLGL.pdf}}
      % \subfloat{\includegraphics[scale=0.22]{images/o10D_CLGL.pdf}}
      % \subfloat{\includegraphics[scale=0.22]{images/o20D_CLGL.pdf}}
    
      \subfloat{\includegraphics[scale=0.27]{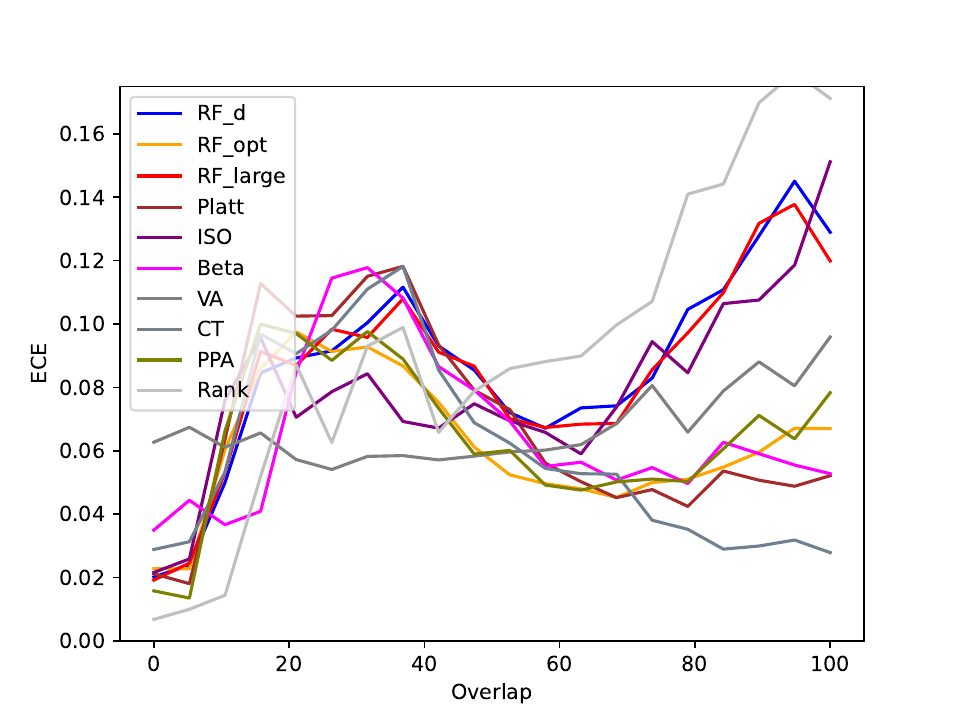}}
      \subfloat{\includegraphics[scale=0.27]{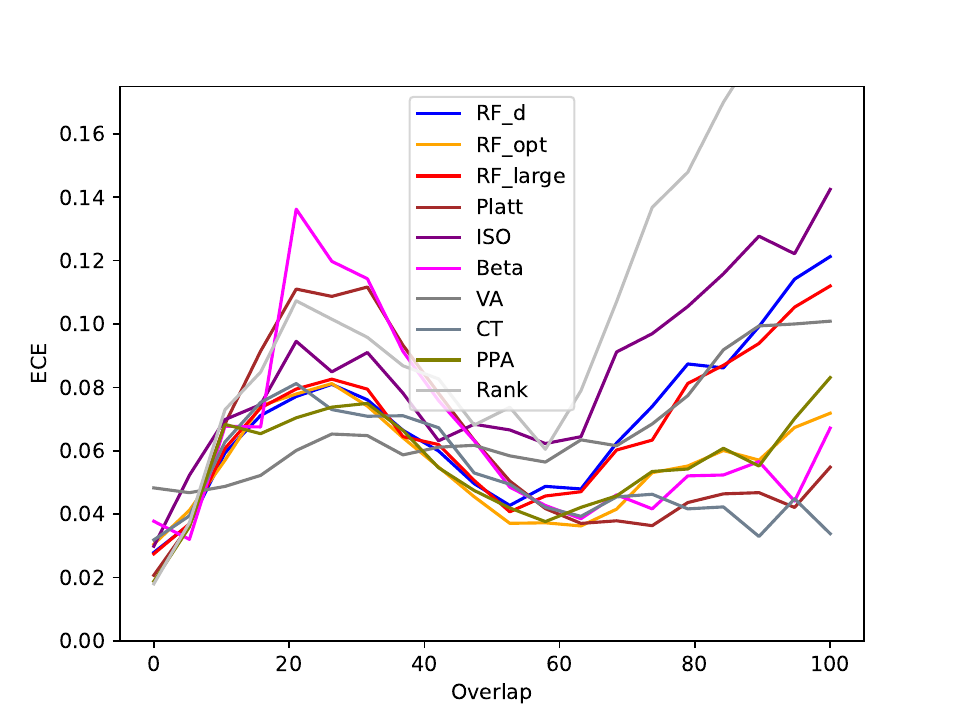}}
      \subfloat{\includegraphics[scale=0.27]{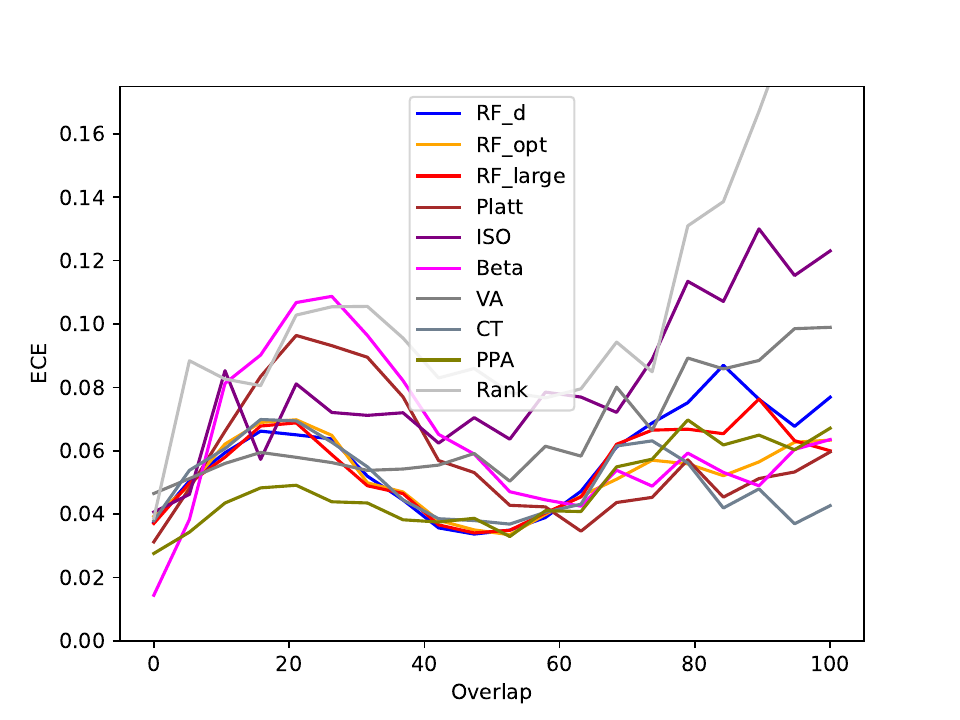}}
      \subfloat{\includegraphics[scale=0.27]{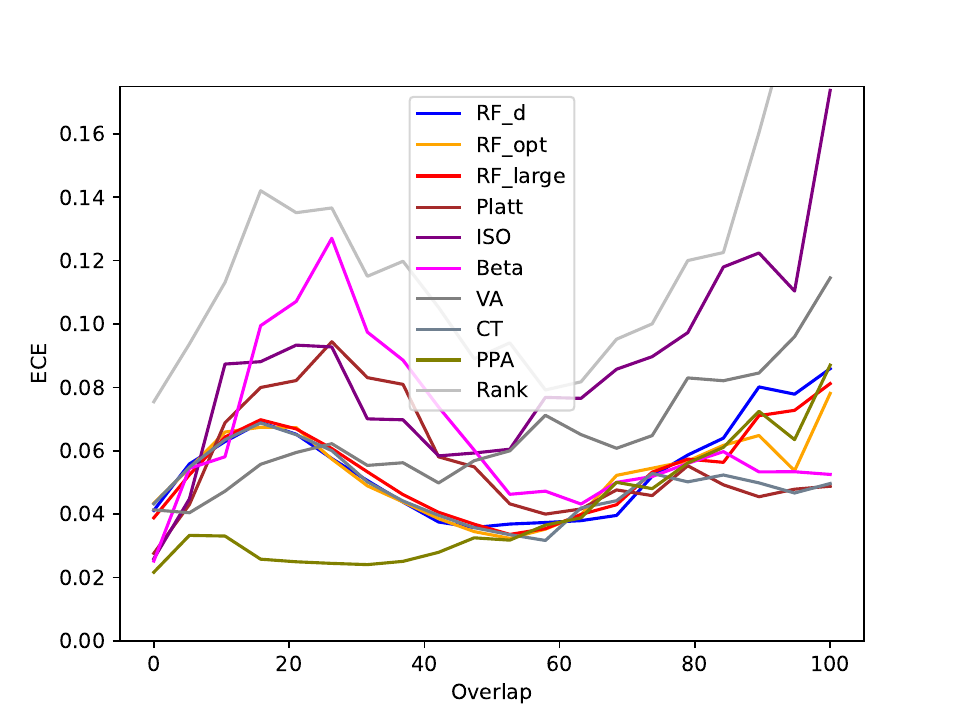}}
    
      % \subfloat{\includegraphics[scale=0.22]{images/o2D_IL.pdf}}
      % \subfloat{\includegraphics[scale=0.22]{images/o5D_IL.pdf}}
      % \subfloat{\includegraphics[scale=0.22]{images/o10D_IL.pdf}}
      % \subfloat{\includegraphics[scale=0.22]{images/o20D_IL.pdf}}
    
    \caption{The impact of varying overlap between two Gaussian distributions on the performance of calibration methods analyzed using synthetic data across increasing dimensions. The results are organized into columns corresponding to each dimensionality—2, 5, 10, and 20—from left to right. The performance metrics—Brier score, TCE, and ECE—are displayed in rows from top to bottom.}
    \label{fig:overlap}
    \end{center}
\end{figure*}

\begin{figure*}[ht]
    \begin{center}
    
      \subfloat{\includegraphics[width=\columnwidth]{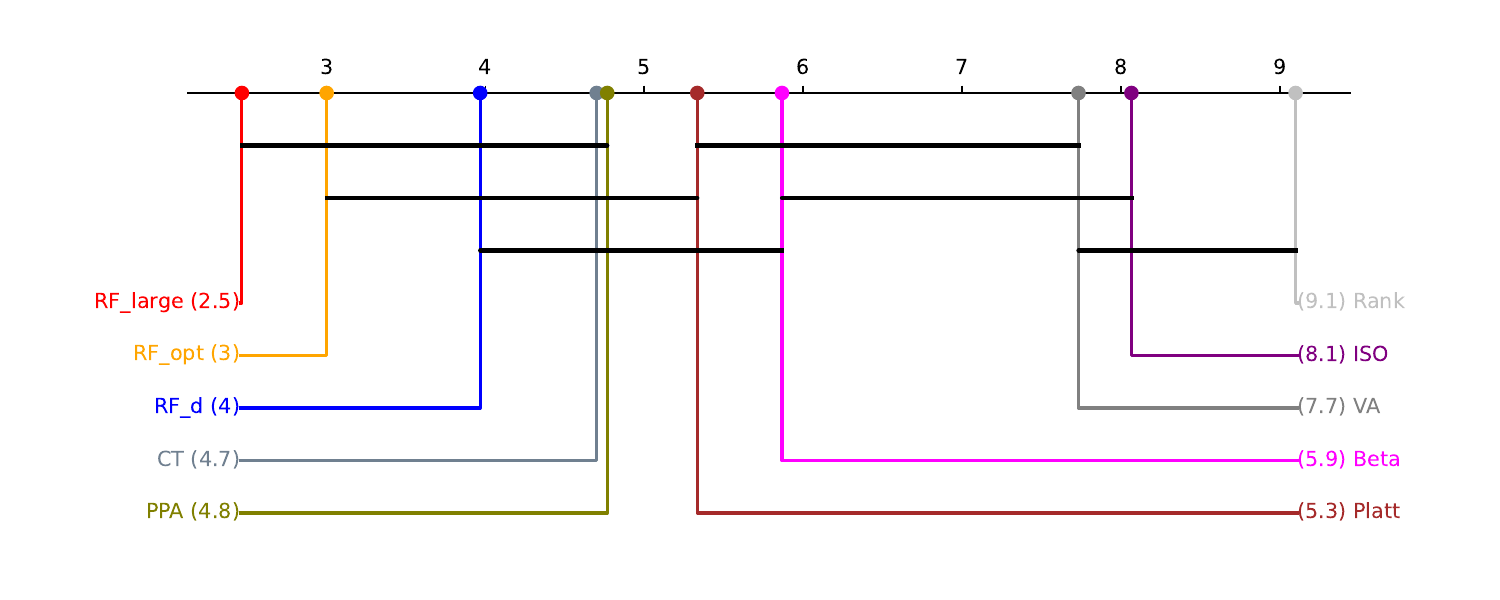}}
      \subfloat{\includegraphics[width=\columnwidth]{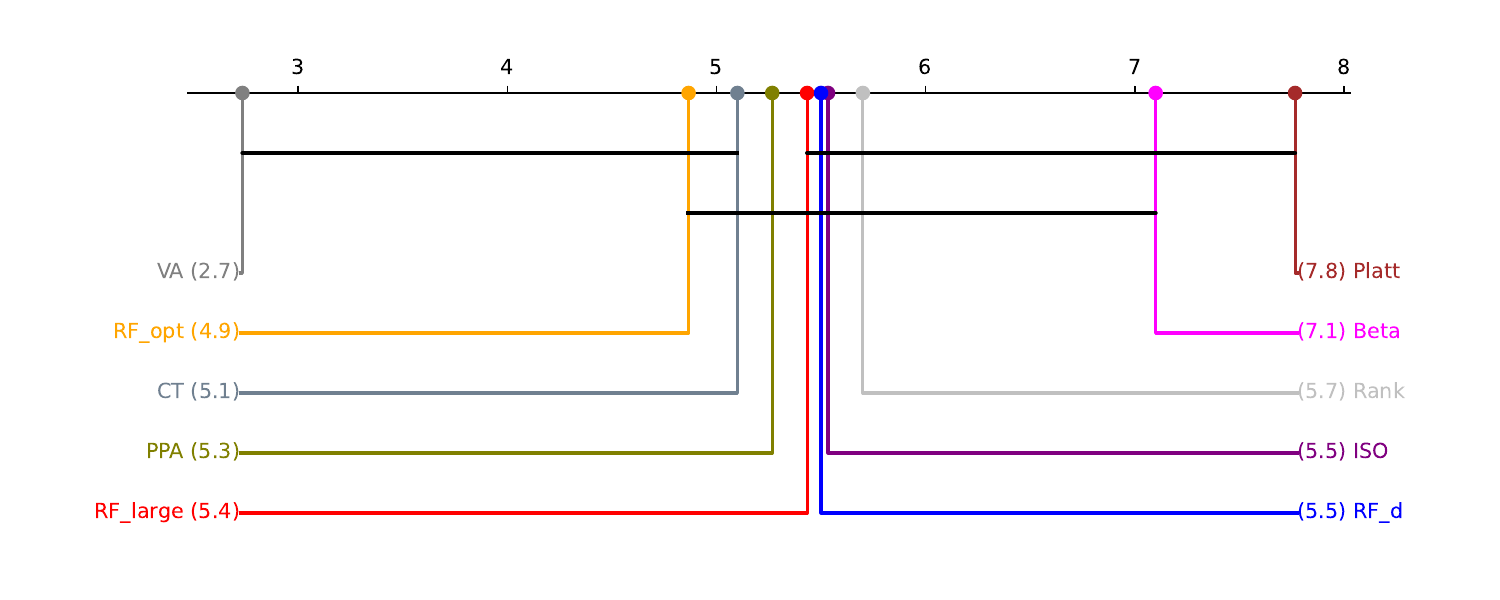}}
    
    \caption{Critical difference diagram of 30 real datasets on Brier score (left) and ECE (right).}
    \label{fig:cd_brier_ece}
    \end{center}
\end{figure*}

The objectives of this section can be summarized as follows:
(i) Gaining insights into RF probability estimation using synthetic datasets.
(ii) Examining the impact of RF hyper-parameters on calibration performance.
(iii) Comparing post-calibration methods applicable to RF and investigating their influence on calibration performance on real datasets.
(iv) The effect of Laplace correction and Out of Bag data on calibration performance (\ref{apx:exp_lap} and \ref{apx:exp_oob}).
(v) Comparing the calibration performance of RF with that of other ML algorithms  (\ref{apx:exp_ml}).

All the codes, experimental setups, and datasets used in this work are available in our Github\footnote{\url{https://anonymous.4open.science/r/RFC-BDC6/README.md}} project link.

\subsection{Effect of RF Maximum Depth on Calibration}

In this section we demonstrate that variations in the hyper-parameters of RF significantly influence it's calibration performance. Thus, conducting hyper-parameter optimization seems to be a promising approach for enhancing RF calibration.

The most important factor of each tree within a forest is its maximal depth. This parameter (max-depth) specifies the maximum number of splits a tree can perform before reaching a leaf node. Increasing the maximum depth of a tree does not necessarily lead to improved performance. With higher values of this parameter, the risk of over-fitting the training data increases. Additionally, performing more splits will influence the calibration performance. 

To analyze the effect of Maximum Depth on calibration performance we generate a synthetic dataset consisting of samples from two overlapping Gaussian distributions. This allows us to compute the true probability distribution for each instance (as detailed and visualized in appendix \ref{apx:exp_sd}). Generating data in this way enables us to produce a basic binary classification dataset comprising two features.

Figure \ref{fig:depth} shows the reliability diagram of three RFs trained on the same synthetic dataset with max-depth set to 2, 4, and 8, respectively. As we can see, there is an optimal depth that will result in the best calibration performance, which is 4 for this particular example; any value higher or lower than the optimal depth results in an ill-trained model or more noise in the reliability diagram, hence, a higher calibration error. 

% original fig:depth place

This behavior can be understood by recalling the decomposition of proper scoring rules (Section \ref{sec:eval}). IL is driven by irreducible aleatoric uncertainty in the data. CL typically increases with node splits, as estimating probabilities for smaller sample sizes in successor nodes becomes harder. In contrast, GL decreases since the groups assigned the same probability estimates shrink. The net effect can be positive or negative. Initially, further splits improve performance, but past a certain threshold, CL dominates, increasing the overall loss. This aligns with the observation that optimal calibration occurs at intermediate tree depths.

\subsection{Overlapping Distributions}
\label{sec:expGO}

% exp run names 100 tree
% 2D  1726176638_Overlap_fix100tree_2D
% 5D  1726193690_Overlap_fix100tree_5D
% 10D 1726214273_Overlap_fix100tree_10D
% 20D 1726244802_Overlap_fix100tree_20D

% \begin{figure*}[ht]
% \begin{center}

%   \subfloat{\includegraphics[scale=0.22]{images/data_o_19.pdf}}
%   \subfloat{\includegraphics[scale=0.22]{images/data_o_11.pdf}}
%   \subfloat{\includegraphics[scale=0.22]{images/data_o_9.pdf}}
%   \subfloat{\includegraphics[scale=0.22]{images/data_o_0.pdf}}

% \caption{A 2D t-SNE visualization of the 10-dimensional synthetic data, depicting the variation in overlap between the two classes.}
% \label{fig:o_data}
% \end{center}
% \end{figure*}

We also explored how the overlap between class-wise distributions in synthetic data impacts calibration performance. Initially, two Gaussian distributions with identical mean vectors and covariance matrices were generated, then the mean vectors were progressively adjusted to increase separation. This experiment was conducted using synthetic datasets of increasing dimensionality (2, 5, 10, and 20 dimensions, each with 1,000 samples) to ensure consistency across dimensions. For more details on data generation please see the appendix \ref{apx:exp_Over}. 
% Figure \ref{fig:o_data} shows a 2D t-SNE visualization of the 10-dimensional data, illustrating four scenarios from no overlap (left) to complete overlap (right).

% original fig:o_data place 

% In addition to the post-calibration methods in appendix \ref{sec:cm}, we also included three RF models as benchmarks:

% \begin{itemize}
%     \item \textbf{RF\_d}: A random forest with default scikit-learn parameters.
%     \item \textbf{RF\_opt}: An optimized random forest using randomized grid search with 5-fold cross-validation (detailed in appendix \ref{apx:exp_SDM}). The number of trees is not optimized, and for fair comparison, RF\_opt uses the same number of trees as RF\_d.
%     \item \textbf{RF\_large}: A random forest with five times more trees than RF\_d.
% \end{itemize}

In addition to the post-calibration methods in Appendix \ref{sec:cm}, we included RF models as benchmarks: RF\_d, a random forest with default scikit-learn parameters; RF\_opt, an optimized version using randomized grid search with 5-fold cross-validation (Appendix \ref{apx:exp_SDM}), with the same number of trees as RF\_d for fair comparison; and RF\_large, which has five times more trees than RF\_d.

The Gaussian overlap experiment was conducted five times, with average performance shown in Figure \ref{fig:overlap}. The key takeaway is that calibration—or proper tuning of RF hyper-parameters—proves beneficial in the "high overlap" regime, where probabilities are less extreme and tend toward a uniform distribution, particularly for low-dimensional data. In this regime, all calibration methods outperform RF\_d and RF\_large, which are prone to overfitting and generating overly extreme probabilities. Calibration shifts these probabilities toward the middle, a result that can also be achieved through hyper-parameter tuning (e.g., with full overlap, the optimal tree has no splits and a depth of 1). However, as data dimensionality increases, more diverse trees are learned, reducing the issue of extreme probabilities in the high-overlap regime. This is notable since the high-overlap scenario is rare in practical applications.

% Note that Brier score is monotonically increasing, due to the increase in irreducible uncertainty. For example, in the case of full overlap, even the best prediction (the uniform distribution) has an expected Brier score of $1/4$. 

Clearly, calibration is easier if ground-truth probabilities are extreme and more difficult for close-to-uniform distributions, which is why we see a small deterioration of TCE in the middle. 
When analyzing datasets with different dimensionalities, we find that instance-wise calibration error increases in the mid-overlap region as the number of dimensions rises. However, ECE plots show a different trend compared to TCE. Interestingly, ECE tends to be lowest in the mid-overlap region but shows higher error in low-overlap settings with fewer dimensions. Since we do not have access to the true probability-wise calibration error, we cannot directly confirm ECE's performance.

% original fig:overlap place 

\subsection{Assessing Calibration Performance on Real datasets}
\label{sec:exp_real}
% exp run name 1725451866_paper10CV100tree

In this section, we evaluate various calibration methods on real datasets. We utilize 30 openly available datasets from the UCI and PROMISE repositories, detailed in the appendix \ref{apx:exp_real} of this paper. The datasets encompass a wide range of sample sizes and feature counts, from small to relatively large.

We employ a 10-fold stratified cross-validation schema, setting aside one fold as a test set, while one fold serves as the calibration set, depending on the experiment type. The remaining folds are used to train the RF classifier. Each cross-validation experiment is repeated five times with different random seeds, and average results are reported. Since true probability distributions are unavailable for real datasets, the evaluation metrics are limited to accuracy, Brier score, log-loss, and ECE with 20 equal-width bins.

We evaluated the performance differences between calibration methods using the Nemenyi-Friedman test for statistical significance \citep{demvsar2006statistical}, with a significance level of 0.05. This analysis was conducted on rankings derived from 30 real-world datasets, reflecting the frequency with which each method outperformed the others (full details of the result tables at appendix \ref{apx:exp_real}). The null hypothesis assumes no statistically significant differences between the methods. A critical difference diagram is used to visualize the results presented in Figure \ref{fig:cd_brier_ece}, highlighting groups of methods for which the null hypothesis cannot be rejected.

The top-performing group varies by metric. For example, the top group for the Brier score includes RF\_large, RF\_opt, RF\_d, CT, and PPA. In contrast, the top group for ECE comprises VA, RF\_opt, and CT. The ECE plot indicates that calibration methods are closely clustered, suggesting minimal significant statistical differences among most methods regarding ECE. This highlights that ECE may not be the most accurate approximation of the probability-wise calibration error.

% original fig:cd_brier_ece place 

\section{Conclusion}
\label{sec:conc}

This paper evaluates and compares various calibration methods for the RF classifier, including model-agnostic and tree-specific approaches. Our results indicate that the effectiveness of these methods varies based on the calibration metric used, revealing no single best approach.

Surprisingly, none of the calibration techniques consistently improve performance in instance-wise or probability-wise calibration. In fact, a hyperparameter-optimized RF (RF\_opt) often matches or exceeds the performance of the best calibration methods. 

Another effective strategy for enhancing RF calibration is increasing the ensemble size, referred to as RF\_large. Both RF\_opt and RF\_large exhibit similar calibration performance across real datasets, with no significant differences. However, RF\_large may underperform in low-dimensional, high-overlap scenarios. The choice between these models should be based on data dimensionality, classification difficulty, and user needs: RF\_opt requires longer training but offers faster predictions, while RF\_large has the opposite characteristics.

Additionally, techniques like Laplace correction and out-of-bag data can improve calibration performance depending on the evaluation metric. Contrary to previous studies suggesting Laplace correction is ineffective, our findings demonstrate its statistical significance, particularly regarding logistic loss and ECE.

% \subsubsection*{References}
\bibliographystyle{apalike} % Change to the desired style
\bibliography{refs}

\appendix

\onecolumn
\aistatstitle{Random Forest Calibration: \\
Supplementary Materials}

% The supplement is organized as follows:
% Section \ref{sec:rf} gives an overview of the RF algorithm, and Section \ref{sec:rw} explores related work on calibrating RF from different perspectives.
% Section \ref{sec:cm} gives an overview of the post-calibration techniques applicable to RF. 
% Section \ref{sec:exp} is continuation and expansion of the experimental section on the main paper, in which we compare calibration methods on both real and synthetic data.

\section{Calibration Methods}
\label{sec:cm}

Calibration is commonly understood as a post-processing method in which a mapping is sought from predicted scores to well-calibrated probabilities. It is motivated by the observation that many machine learning models yield biased predictions in the first place, for example, predictions that are overly confident or systematically over- or underestimate probabilities. 

As a starting point, calibration methods proceed from a trained predictor $f$ and a set of calibration data $\mathcal{D}_{cal} \subset \mathcal{X} \times \mathcal{Y}$. In the following, we consider the case of binary classification with $\mathcal{Y} = \{ 0, 1 \}$, which is assumed by most calibration methods (although extensions to the multinomial case are normally possible). In this case, $f$ is a scoring classifier $\mathcal{X} \fromto \mathbb{R}$, where the value $f(\vec{x})$ is an indicator of the positive class. In many cases, including our case of RF, the scores are already normalized to $[0,1]$ and can be interpreted as (pseudo-)probabilities, but this is not required. What is assumed, however, is that higher scores $f(\vec{x})$ indicate a higher propensity for the positive class. Therefore, score-to-probability mappings induced by calibration methods are guaranteed to be \emph{monotonic}. Subsequently, without loss of generality, we assume the calibration data $(\vec{x}_1, y_1), (\vec{x}_2, y_2), \ldots, (\vec{x}_M, y_M)$ to be ordered such that $f(\vec{x}_1) \leq  f(\vec{x}_2) \leq \ldots \leq f(\vec{x}_M)$. For brevity, we denote the score $f(\vec{x}_m)$ by $s_m$.

\subsection{Platt Scaling and Beta Calibration}

Platt scaling \citep{platt1999probabilistic} was among the first calibration methods used in machine learning, It has originally been devised for calibrating support vector machines (where scores are signed distances from the decision boundary), but can be applied for binary classification in general. It proceeds from the assumption that class-conditional distributions of scores are normal with equal variance. This assumption justifies the following logistic calibration map:
\begin{equation}\label{eq:platt}
p_{\text{Platt}}: \, \mathbb{R} \fromto [0,1] , \, s \mapsto  \frac{1}{1 + e^{\gamma \cdot s + \delta}} \, .
\end{equation}
The parameters $\gamma \geq 0$ and $\delta \in \mathbb{R}$, which specify the sigmoidal shape of the function, are fit to the calibration data by  
minimizing log-loss  
\begin{equation}
\sum_{m=1}^M  - y_m \log \left(p_{\text{Platt}}(s_m)\right) - \left(1 - y_m\right) \log \left(1 - p_{\text{Platt}}(s_m)\right)  
\end{equation}
of the predicted probabilities $p_m = p_{\text{Platt}}(s_m)$ on the calibration data.
While this can be done quite efficiently using gradient-based optimization, 
the effectiveness of Platt scaling strongly hinges on the underlying assumption on the distribution of scores. In particular, (\ref{eq:platt}) does not appear suitable if scores $s_m$ are bounded by 0 and 1. Also note that (\ref{eq:platt}) cannot reproduce the identity $s \mapsto s$, which would be needed if the scores are already well-calibrated.

More recently, beta calibration has therefore been introduced as an alternative \citep{kull2017beta}. As the name suggests, this method assumes class-wise scores to follow a beta (instead of a normal) distribution.

It comes down to fitting a calibration function that has three parameters $a,b \geq 0$, $c \in \mathbb{R}$ and, therefore, is slightly more flexible than the logistic calibration map\footnote{Logistic calibration (\ref{eq:platt}) is obtained as a special case of beta calibration for $a = b$.}: 
\begin{equation}
p_{\text{beta}}: [0,1] \fromto [0,1], \,  s \mapsto \frac{1}{1+1 /\left(\exp(c) \frac{s^a}{(1-s)^b}\right)} \, .
\end{equation}

Again, this model is fitted by minimizing log-loss on the calibration data, which can be done using any appropriate optimization method. Using a suitable parameterization, \citet{kull2017beta} show that the problem can also be reduced to fitting a bivariate logistic regression model.

\subsection{Isotonic Regression and Venn-Abers Calibration}

Platt scaling and beta calibration are parametric methods, both coming with (more or less restrictive) assumptions about the  distribution of scores. A non-parametric alternative is provided by isotonic regression \citep{robertson1988order}, which has first been used for calibration by \citet{zadrozny2002transforming}. It fits a piece-wise constant function $p_{iso}: \mathbb{R} \fromto [0,1]$ with steps around the scores $s_m$ in the calibration data. The corresponding step sizes $p_i$ are determined by minimizing the squared error loss
$$
\sum_{m=1}^M (p_m - y_m)^2 \quad \text{s.t.} \quad p_1 \leq p_2 \leq \ldots \leq p_M \, .
$$
This is a constrained (convex) optimization problem that can be solved quite efficiently in linear time, e.g., using the pool-adjacent violators (PAV) algorithm \citep{ayer1955empirical}.

Isotonic regression comes with an automatic binning of scores that are mapped to the same probability, i.e., segments $s_i, s_{i+1}, \ldots , s_j$ in the calibration data such that $p_i = p_{i+1} = \ldots = p_j$. Apart from compliance with the monotonicity constraint, it allows for fitting the calibration data in a very flexible manner. As for non-parametric methods in general, this can be an advantage and disadvantage at the same time: it avoids any bias due to incorrect model assumptions but increases the risk of over-fitting the (calibration) data.

The Venn-Abers predictor \citep{vovk2012venn} uses isotonic regression, too, albeit in a slightly different way. Venn-Abers is a specific type of Venn predictor \citep{vovk2003self}, which in turn is rooted in conformal prediction, a statistical framework for set-valued prediction \citep{bala_cp}. Instead of producing a point prediction in the form of a single probability degree $p$, Venn-Abers constructs an interval $[p_0,p_1]$ that comes with a certain guarantee of validity. Broadly speaking, under certain technical assumptions, the interval (which is a random object as it depends on the data) is guaranteed to contain the true probability \emph{in expectation}; in other words, the true probability is contained in $[\mathbb{E}(P_0), \mathbb{E}(P_1)]$, where $P_0$ and $P_1$ denote the (random) lower and upper bounds of the interval, and the expectation is taken over the data-generating process. 

\begin{algorithm}
\caption{\text{Venn-Abers predictor}} \label{venn_abers_algo}
\hspace*{\algorithmicindent} \textbf{Inputs:} \\
\hspace*{\algorithmicindent} \hspace*{\algorithmicindent}
calibration data $(\vec{x}_1, y_1),\ldots,(\vec{x}_M, y_M)$, query $\vec{x}$ \\
\hspace*{\algorithmicindent} \textbf{Outputs:} \\
\hspace*{\algorithmicindent} \hspace*{\algorithmicindent}
multiprobabilistic prediction $(p_0,p_1)$
\begin{algorithmic}[1]
\For{$y \in \{ 0, 1 \}$}
\State set $s_y$ to the scoring function for $(\vec{x}_1, y_1),\ldots,(\vec{x}_M, y_M), (\vec{x}, 0)$
\State set $g_y$ to the isotonic calibrator for
$(s_y(\vec{x}_1), y_1),\ldots,(s_y(\vec{x}_M), y_M), (s_y(\vec{x}), 0)$
%$(s_y (x_1), y_1), \dots, (s_y (x_l), y_l), (s_y (x), y)$
\State set $p_y$ to $g_y(s_y (\vec{x}))$
\EndFor
\end{algorithmic} 
\end{algorithm}

Given a query point $\vec{x}$, Venn-Abers produces a prediction interval by doing isotonic regression twice, first on the calibration data augmented by $\vec{x}$ hypothetically labeled negative, and then on the calibration data augmented by $\vec{x}$ hypothetically labeled positive. The lower bound $p_0$ is taken from the first isotonic function and the upper bound $p_1$ from the second one. If a single point-prediction is needed, the following probability can be motivated by an argument based on the minimax principle:
\begin{equation*}
    p_{\text{VA}} = \frac{p_0 + p_1}{2} +\left(p_1-p_0\right)\left(\frac{1}{2}-\frac{p_0 + p_1}{2} \right),
\end{equation*}
Algorithm \ref{venn_abers_algo} shows Venn-Abers in pseudo-code. Obviously, this algorithm is computationally more demanding, as it requires repeated execution of isotonic regression.

\subsection{PPA Calibration}

\citet{bostrom2008calibrating} found that RF using classification trees as base learners tends to outperform RF using probability estimation trees in terms of the Brier score, whereas the latter exhibits superior performance in terms of classification accuracy and AUC. Upon closer inspection, averaging the probabilities predicted by PETs seems to bias the estimates toward the uniform distribution.

To counter this effect, \citet{bostrom2008calibrating} introduced Parameterized Probability Adjustment (PPA), 
which increases the estimated probability for the most probable class and decreases the others:
$$
\vec{p}_{\text{PPA}}= r \, \vec{p}_0 + (1-r ) \, \vec{p} \, , 
$$
where $\vec{p}$ is the probability originally predicted and $\vec{p}_0$ the distribution in which the mass of 1 is uniformly distributed among the labels with highest probability in $\vec{p}$ (setting the probability to 0 for all others). 
The parameter \(r\) is optimized on the calibration data with the objective of minimizing the Brier score.

Obviously, this approach leaves the class prediction, and hence the classification accuracy, unchanged. By retaining the high accuracy obtained with PETs while simultaneously achieving a low Brier score with CTs (indicating better calibration performance), it combines the strengths of both PETs and CTs as base learners in RF.

\subsection{Curtailment}

\citet{zadrozny2001obtaining} emphasize two main issues with tree probability estimates: high bias due to purification of leaf nodes, and high variance due to fragmentation and estimation based on low sample size.
Obviously, both problems can be countered by pruning, although standard pruning techniques may yield suboptimal results in the case of class imbalance.

\citet{zadrozny2001obtaining} propose to set a threshold parameter $v$ and to ensure that any leaf node must contain at least $v$ samples to make probabilistic predictions. They call this approach \emph{curtailment},  
The complexity of this approach depends on how the threshold \(v\) is determined, by means of a simple heuristic or by fitting it to the calibration data (e.g., using cross-validation).

Applying curtailment to individual trees within RF can enhance calibration. Nonetheless, one drawback of this method, as highlighted by \citet{wu2021should}, is that aggregating calibrated probabilities from individual trees through averaging could potentially undo the calibration achieved. This notion will become clearer in the next section, wherein the calibration methods are compared based on different calibration metrics.

\subsection{RF Rank Calibrator}

An interesting observation is that isotonic regression (like other non-parametric methods) is invariant against strictly monotonic transformation of the (calibration) scores $s_m$. Consequently, the underlying learner producing these scores can be very uncalibrated, as long as it makes sure that the scores are well \emph{ordered}: if $\vec{x}_i$ is assigned score $s_i$ and $\vec{x}_j$ is assigned scores $s_j$, then $s_i > s_j$ implies that the probability for $\vec{x}_i$ is indeed higher than the probability for $\vec{x}_j$. Indeed, due to the monotonicity constraint, an incorrect ranking $s_i < s_j$ cannot be repaired by isotonic regression. 

From this, one may conclude that, in the first place, the underlying learner should be a strong ranker. 
With this idea in mind, \citet{menon2012predicting} propose a calibration method that optimizes a ranking loss first and applies isotonic regression thereafter, and indeed find improved calibration performance. Therefore, we are interested in applying the same approach with RF.

The initial step in this approach is to rank instances using individual trees. This problem was studied by \citet{hullermeier2009fuzzy}, who show that diversifying scores and resolving ties improves ranking (though not necessarily classification) performance. They recommend using unpruned trees with Laplace correction for probability estimation and ranking instances according to these estimates. We adopt the same approach in our experimental study.

Extending this approach from trees to forests requires the aggregation of the rankings coming from the individual trees. To this end, any rank aggregation procedure can be used. Here, we apply a score-based approach, which is in line with the so-called Borda aggregation \citep{borda1781m} and comes down to scoring an instance $\vec{x}$ as follows:
\begin{equation}\label{eq:borda}
s(\vec{x}) = \sum_{j=1}^T \sum_{i=1}^M \mathbf{1} \{ p_j(\vec{x}) > p_j(\vec{x}_i) \} + \frac{1}{2} \mathbf{1} \{ p_j(\vec{x}) = p_j(\vec{x}_i) \} \, ,
\end{equation}
where $p_j(\vec{x})$ is the probability (of the positive class) predicted for $\vec{x}$ by the $j^{th}$ tree, and $\vec{x}_1, \ldots , \vec{x}_M$ is the calibration data. Eventually, a calibrated probability estimate for a query instance $\vec{x}$ is hence obtained as follows: Scores $s_1, \ldots, s_M$ with $s_i = s(\vec{x}_i)$ are obtained from (\ref{eq:borda}) for the calibration data, a calibration map $p_{rank}: \mathbb{R} \fromto [0,1]$ is constructed by applying isotonic regression to these scores, and this map is used to produce the estimate $p_{rank}(s(\vec{x}))$.  

\section{Random Forest}
\label{sec:rf}

% RF short intro \citet{ho1995random}
% Random Forest, introduced by \citep{breiman2001random}, is a tree-based ensemble that works by aggregating the results of different decision trees grown on random sub-spaces selected from the feature space parallel. 

Random Forest as introduced by \citet{breiman2001random} is an ensemble method based on decision trees. 
%To extract different trees from the same training data, it combines different types of randomization, some of which had already been proposed in earlier work. 
%The methodology developed by  forms the basis for modern Random Forest techniques. 
It improves upon earlier work by \citet{ho1995random} by combining two types of randomization, namely, bootstrapping \citep{breiman1996bagging} and random feature selection by \citep{AmitG97}: 
\begin{itemize}
 \item    
% bootstrapping
Each member of the ensemble is trained on a bootstrap sample of the original training data $\mathcal{D}$, which is obtained by sampling $N =  |\mathcal{D}|$ data points from $\mathcal{D}$ with replacement \citep{breiman1996bagging}. The use of bagging appears to improve accuracy particularly when random features are introduced. 
\item
At each node of a decision tree, the best split if found, not among all features, but only among a randomly chosen subset of features \citep{AmitG97}. 
\end{itemize}
Using these randomization techniques, a diverse set of trees is produced, which, for a given data point, can produce different predictions.
A simplified version of the algorithm for constructing a Random Forest is provided in Pseudocode~\ref{algo:RF}.

\begin{algorithm}
\caption{Random Forest Algorithm}
\label{algo:RF}
\begin{algorithmic}[1]
\Procedure{RandomForest}{$\text{TrainingData}, \text{NumTrees}, \text{NumFeatures}$}
    \State $\text{Forest} \gets \emptyset$
    \For{$i \gets 1$ to $\text{NumTrees}$}
        \State $\text{BootstrapSample} \gets \text{CreateBootstrapSample}(\text{TrainingData})$
        \State $\text{Tree} \gets \text{BuildDecisionTree}(\text{BootstrapSample}, \text{NumFeatures})$
        
        \State $\text{Forest} \gets \text{Forest} \cup \text{Tree}$
    \EndFor
    \State \textbf{return} $\text{Forest}$
\EndProcedure

\Procedure{CreateBootstrapSample}{$\text{Data}$}
    \State $\text{SampleSize} \gets \text{length}(\text{Data})$
    \State $\text{BootstrapSample} \gets \emptyset$
    \For{$i \gets 1$ to $\text{SampleSize}$}
        \State $\text{RandomIndex} \gets \text{RandomInteger}(1, \text{SampleSize})$
        \State $\text{BootstrapSample} \gets \text{BootstrapSample} \cup \text{Data}[\text{RandomIndex}]$
    \EndFor
    \State \textbf{return} $\text{BootstrapSample}$
\EndProcedure

\Procedure{BuildDecisionTree}{$\text{Data}, \text{NumFeatures}$}
    \If{$\text{StoppingCriteriaMet}(\text{Data})$}
        \State \textbf{return} $\text{CreateLeafNode}(\text{Data})$
    \EndIf
    \State $\text{FeatureSubset} \gets \text{RandomSubset}(\text{NumFeatures})$
    \State $\text{SplitFeature}, \text{SplitValue} \gets \text{FindBestSplit}(\text{Data}, \text{FeatureSubset})$
    \State $\text{DataLeft}, \text{DataRight} \gets \text{SplitData}(\text{Data}, \text{SplitFeature}, \text{SplitValue})$
    \State $\text{LeftChild} \gets \text{BuildDecisionTree}(\text{DataLeft}, \text{NumFeatures})$
    \State $\text{RightChild} \gets \text{BuildDecisionTree}(\text{DataRight}, \text{NumFeatures})$
    \State \textbf{return} $\text{CreateDecisionNode}(\text{SplitFeature}, \text{SplitValue}, \text{LeftChild}, \text{RightChild})$
\EndProcedure

\end{algorithmic}
\end{algorithm}

%Upon the independent generalization of trees, it becomes imperative to aggregates the results in a manner that retains their respective accuracies. Leveraging the posterior probability obtained from each individual tree, the function to compute the posterior probability distribution for the Random Forest is expressed as follows:
A single decision tree can be used for both deterministic and probabilistic prediction\,---\,we speak of a classifier tree (CT) in the first and of a probability estimation trees (PET) in the second case. 
In the case of PET, the probability distribution $\vec{p}(\vec{x})$ predicted for a query instance $\vec{x}$ is normally given by the relative frequency distribution of class labels in the leaf node that $\vec{x}$ is assigned to. In the case of CT, the prediction is a degenerate distribution that assigned probability 1 to the most frequent class label and 0 to all others. In either case, the prediction produced by the entire ensemble is obtained by averaging over the predictions of all trees: 
\begin{equation}
g( \vec{x} )=\frac{1}{T} \sum_{j=1}^T \vec{p}_j(\vec{x}) \, ,
\end{equation}
where $T$ denotes the number of trees and $\vec{p}_j(\vec{x})$ the distribution predicted by the $j^{th}$ tree.

%$P\left(c \mid v_j(x)\right)$ signifies the posterior probability of sample $x$ belonging to class $y~(y \in Y)$ in the $j$-th tree ($j=1,2,\dots, T$), with $v_j(x)$ representing the leaf node reached by $x$ during traversal of the $j$-th tree. The posterior probability of a tree for class $y$ is determined as the ratio of the number of samples associated with class $y$ in leaf $v_j(x)$ to the total number of samples in that leaf. To predict the class label for a given sample $x$, $g_y(x)$ is computed for all class labels $(y \in Y)$, and the prediction involves selecting the class label associated with the maximum $g_y(x)$ value.

RF has different (hyper-)parameters that may strongly influence the learning process: 
%Here we explain some of the most important parameters:
\begin{itemize}
  \item \textbf{Number of estimators:} The number of decision trees (ensemble size), which has an influence on the m Increasing it will make the model more powerful and complex, and decreasing it will make the model simpler.
  \item \textbf{Split criterion:} A measure of information gain used to find the optimal split in a decision node.
 % The characteristic that measures the degree of purity for splitting a decision node of a tree. It can be log loss, Gini impurity or information gain that we explained earlier.
  \item \textbf{Max depth:} The maximum depth of each decision tree, after which no further splitting is allowed.
  %is a point which no further splitting happens in the tree nodes after the set max depth. 
  Increasing it can make the model more expressive but also comes with a risk of over-fitting.
  \item \textbf{Max features:} The number of features to consider when choosing the best feature to split on. Increasing it will increase the chance of finding a better split but at the same time decrease diversity between trees in the RF.
  \item \textbf{Min samples leaf:} The minimum number of samples required in a leaf node.
  \item \textbf{Minimum samples split:} The minimum number of samples required to split an internal node.
  \item \textbf{Laplace correction:} If activated, the probability estimates based on relative frequencies are regularized by adding a pseudo-count of 1 for each class label.
  %each tree's probability output in the Random Forest will be adjusted by incorporating a sample from all classes into the determining leaf node to prevent probabilities of 0 or 1 as the tree's output.
  
\end{itemize}
Thanks to the use of bootstrapping, RF can benefit from out-of-bag (OOB) data. OOB data refers to the data that is not included in a single bootstrap sample (recall that sampling is done without replacement). Thus, for the tree trained on that sample, the OOD data is out-of-sample and can be used for purposes that require ``clean'' data, such as validation. 
%a specific concept related to the training process and evaluation of individual base learners within an ensemble. As mentioned, RF combines multiple decision trees to make predictions. Each tree of the RF is trained on a bootstrap sample from the original training data set, which involves randomly selecting data points with replacement to create a new subset for each tree's training. The key point is that not all data points are included in the training set of each individual decision tree. 
In the following sections, we will elucidate the advantageous of OOB data in the calibration performance of an RF.

% When constructing each tree, about one-third of the original data points are left out due to the random selection with replacement. These left-out data points constitute the out-of-bag data for that particular tree. Another way to look at this is to say that for each data point in the original training data, there exists a subset of the trees in the Random Forest that are not trained on that specific data point. 

\section{Experiments}
\label{apx:exp}

This section includes a continuation of the experiments in section \ref{sec:exp}. The experiments are separated into four parts. The first part focuses on investigating the RF model for calibration under synthetically generated data for which we know the true probability distribution of each sample. The second part is a continuation on the effect of RF hyper-parameters on calibration performance. The third part is related to synthetic data manipulation, where we show the effect of calibration set size, data overlap and dimensionality on calibration performance. Finally, the fourth part compares calibration methods for RF and evaluates the calibration performance of RF against other ML algorithms on real datasets.

We begin with an explanation of the experimental setup used to evaluate different post-calibration methods on an RF model.
Given that certain calibration methods introduced in appendix \ref{sec:cm} are specifically designed for the binary classification context, our experiments in this paper are confined to a binary classification setting. All experiments of the paper where conducted on a M1 Pro MacBook Pro  with 16GB of RAM.

\subsection{Simple Synthetic Data} 
\label{apx:exp_sd}

To compare the probability distributions predicted by RF with those of other machine learning models, we generate a synthetic dataset which allows us to compute the true probability distribution for each instance. 

The synthetic data utilized in this study was generated to produce a binary classification dataset with known true probabilities for each instance. For this purpose, two multi-variate Gaussian distributions were sampled. The dimensionality of the distribution corresponds to the number of features (which in this case is two). Sampling from the first Gaussian distribution yields samples for the positive class, while sampling from the second Gaussian distribution yields samples for the negative class. With known mean and covariance parameters of the distribution, we can calculate the true probability for each sample using the Bayes role. To maintain a balanced dataset, both distributions were sampled equally; with a dataset size of 1000, each distribution was sampled 500 times.

\begin{figure*}[ht]
\begin{center}

  \includegraphics[scale=0.70]{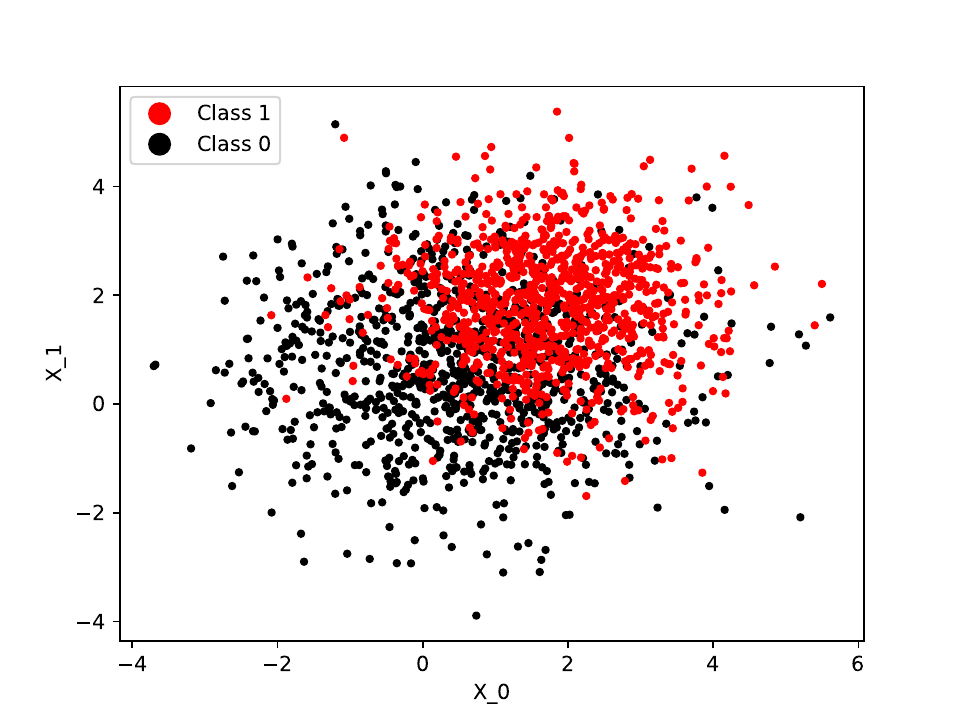}

\caption{Synthetic two dimensional binary dataset generated from two overlapping Gaussian's with the same covariance matrix but different means. The axis on the plot shows the two features of this dataset.}
\label{fig:synthetic_data_1}
\end{center}
\end{figure*}

The means of the Gaussian distributions are determined by two arrays, each sized to correspond with the number of features. The values in these arrays are sampled from a uniform distribution between 0 and 1 for the first Gaussian distribution, and between 1 and 3 for the second Gaussian distribution. The covariance of the Gaussian distributions is represented by a diagonal matrix. For both Gaussian distributions, the values in this matrix are uniformly drawn between 1 and 2. This data is visualized in Figure \ref{fig:synthetic_data_1}.

The default Random Forest parameters utilized in the study are presented in Table \ref{tab:rfd_hyper-parameters}. These values are the default settings from the scikit-learn package\footnote{\url{https://scikit-learn.org/stable/modules/generated/sklearn.ensemble.RandomForestClassifier.html}}.

\begin{table}[h]
    \centering
    \caption{The hyper-parameters used for the default Random Forest}
    \begin{tabular}{|l|l|}
        \hline
        \textbf{hyper-parameter} & \textbf{Value} \\
        \hline
        Number of trees & 100 \\
        \hline
        Criterion & gini \\
        \hline
        Maximum depth & None \\
        \hline
        Minimum samples split & 2 \\
        \hline
        Minimum samples leaf & 1 \\
        \hline
        Maximum features & sqrt \\
        \hline
        Class Weight & None \\
        \hline
        Bootstrap & True \\
        \hline
        Laplace & False \\
        \hline
    \end{tabular}
    \label{tab:rfd_hyper-parameters}
\end{table}

\begin{figure*}[ht]
\begin{center}

  \subfloat{\includegraphics[scale=0.3]{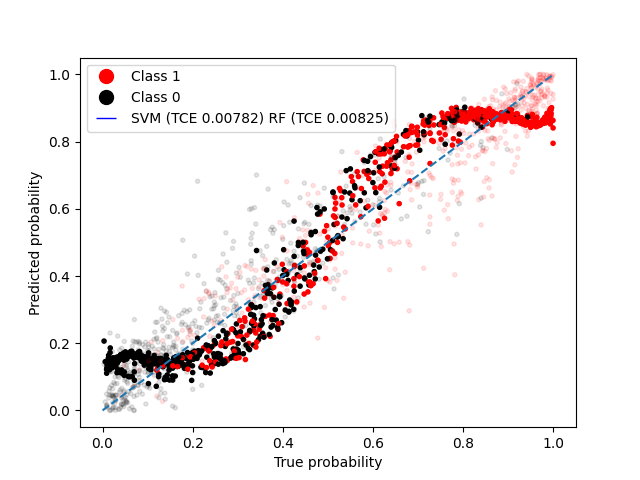}}
  \subfloat{\includegraphics[scale=0.3]{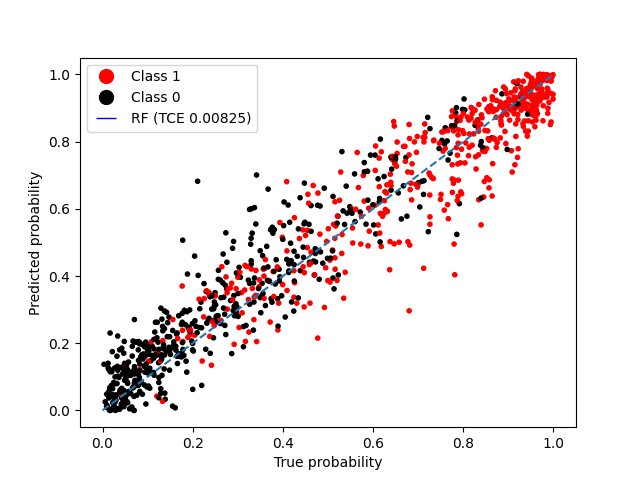}}
  \subfloat{\includegraphics[scale=0.3]{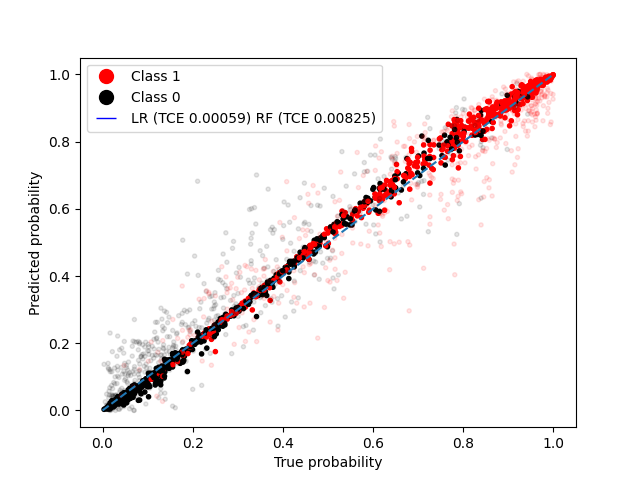}}
  
  \subfloat{\includegraphics[scale=0.3]{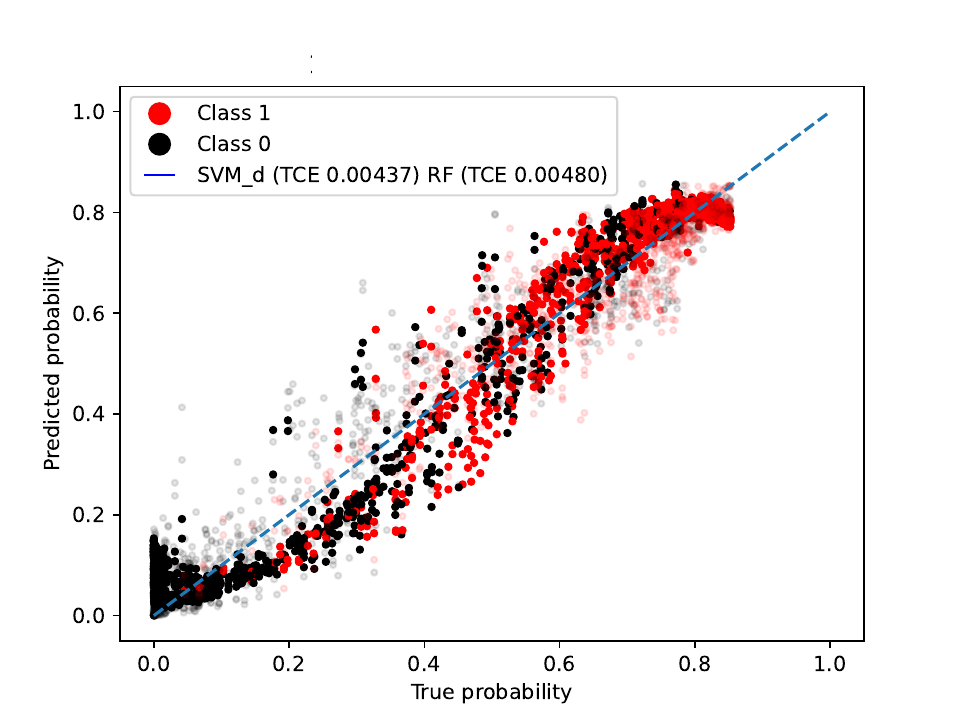}}
  \subfloat{\includegraphics[scale=0.3]{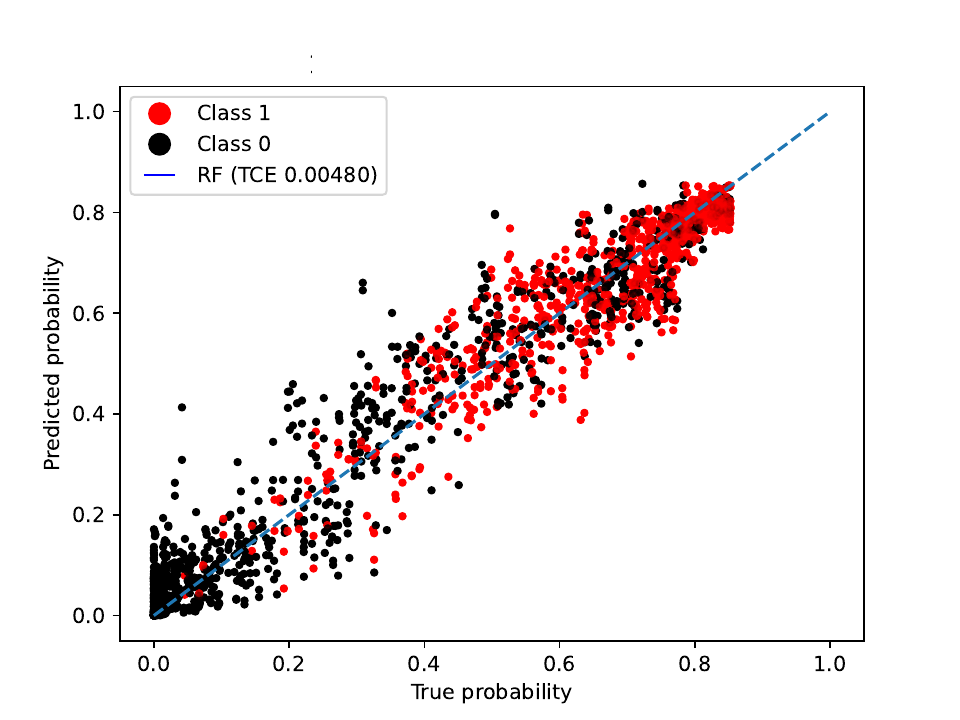}}
  \subfloat{\includegraphics[scale=0.3]{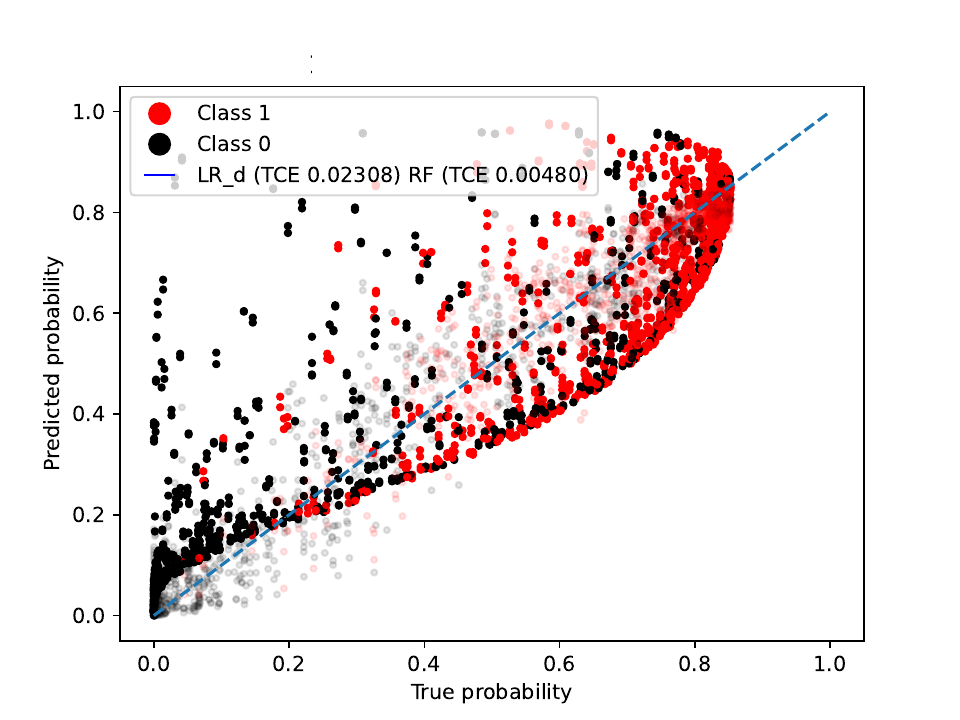}}

\caption{The reliability diagram of comparison between SVM classifier on the left, RF (in the middle) and logistic regression (on the right). All the hyper-parameters of the three models are set to default values of the scikit-learn library. To facilitate the comparison, the light points serve as an overlay of the RF outputs.}
\label{fig:models}
\end{center}
\end{figure*}

In Figure \ref{fig:models}, we contrast a default RF (of probability estimation trees) with Support Vector Machine (SVM) using probabilities from Platt scaling, as defined in the scikit-learn implementations, along with logistic regression (LR) classifiers.

The plot displays the true probability on the $x$-axis and the predicted probability on the $y$-axis. In this reliability diagram, the dashed diagonal line represents the ideal scenario, indicating perfect calibration where the predicted probability equals the true probability. Among all models, LR exhibits the closest alignment with the diagonal. 

One should note, however, that the model assumptions underlying LR are indeed exactly fulfilled by the data-generating process. As can be seen in the second row of Figure \ref{fig:models}, when the Gaussians generating the data have different covariance matrices (the first distribution's covariance diagonal matrix consists of values uniformly drawn between 4 and 5), LR yields biased probability predictions. The reliability diagram of the SVM model has a curvature, suggesting that the model tends to be under-confident on instances with high predictive probabilities and over-confident in cases with low predictive probabilities. 

As shown by these examples, methods relying on specific model assumptions are likely to yield biased probability estimates as soon as these assumptions are violated. RF, for which such a bias is not visible, appears to be more robust in this regard. Yet, the distribution produced by RF looks more scattered and noisy. This observation motivates a closer examination of how various hyper-parameters on RF influence the probability predictions.

% \subsection{Effect of RF Parameters on Calibration}

\subsection{Effect of RF Number of Trees on Calibration }

\begin{figure*}[ht]
\begin{center}

  \subfloat{\includegraphics[scale=0.3]{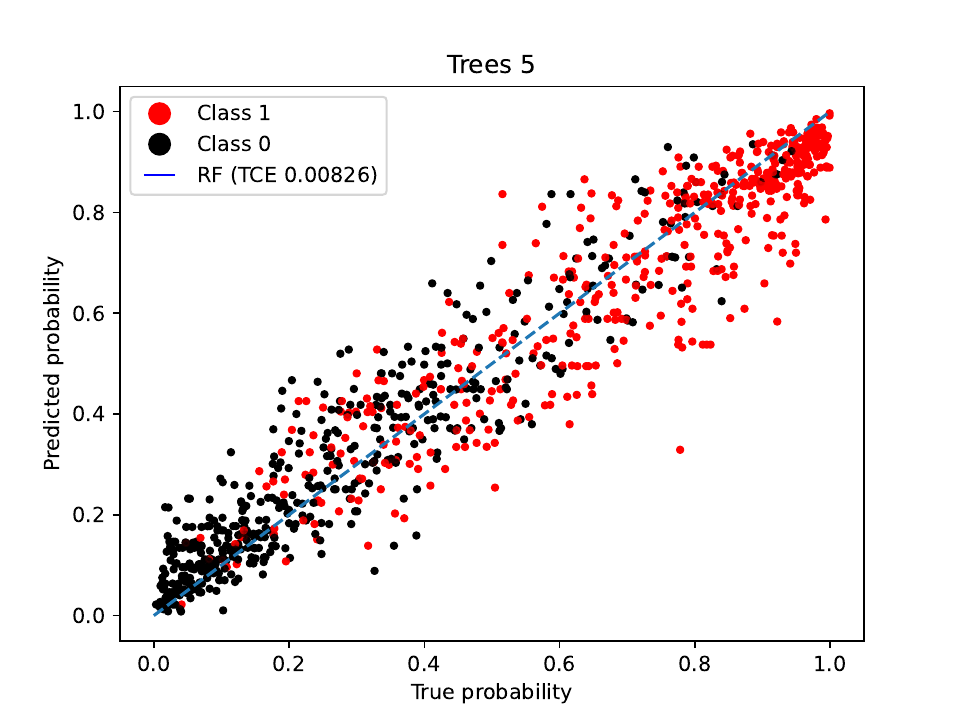}}
  \subfloat{\includegraphics[scale=0.3]{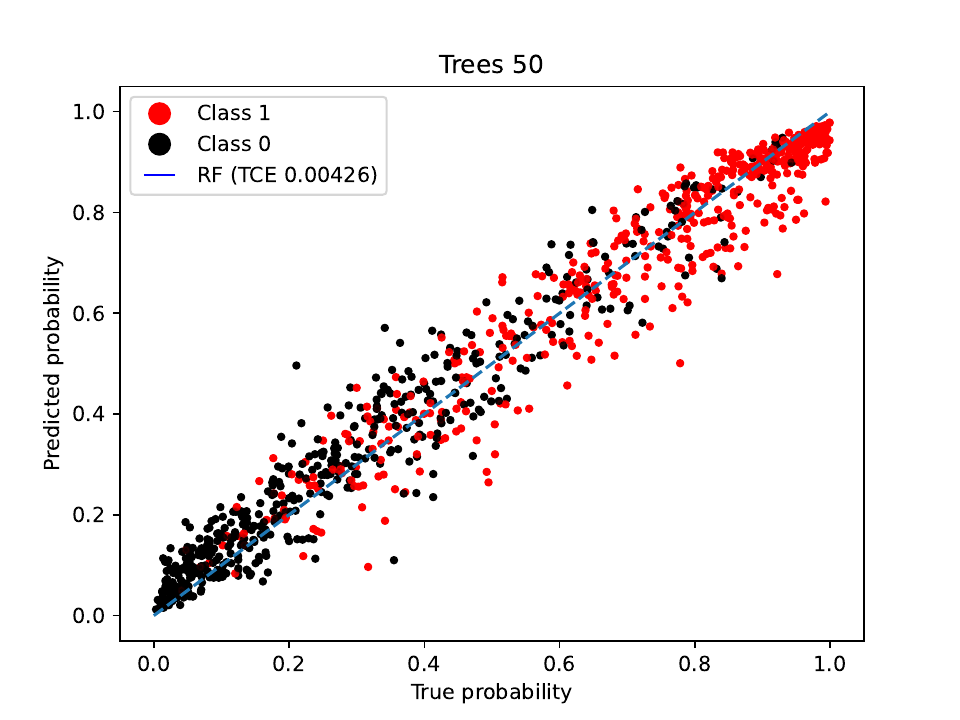}}
  \subfloat{\includegraphics[scale=0.3]{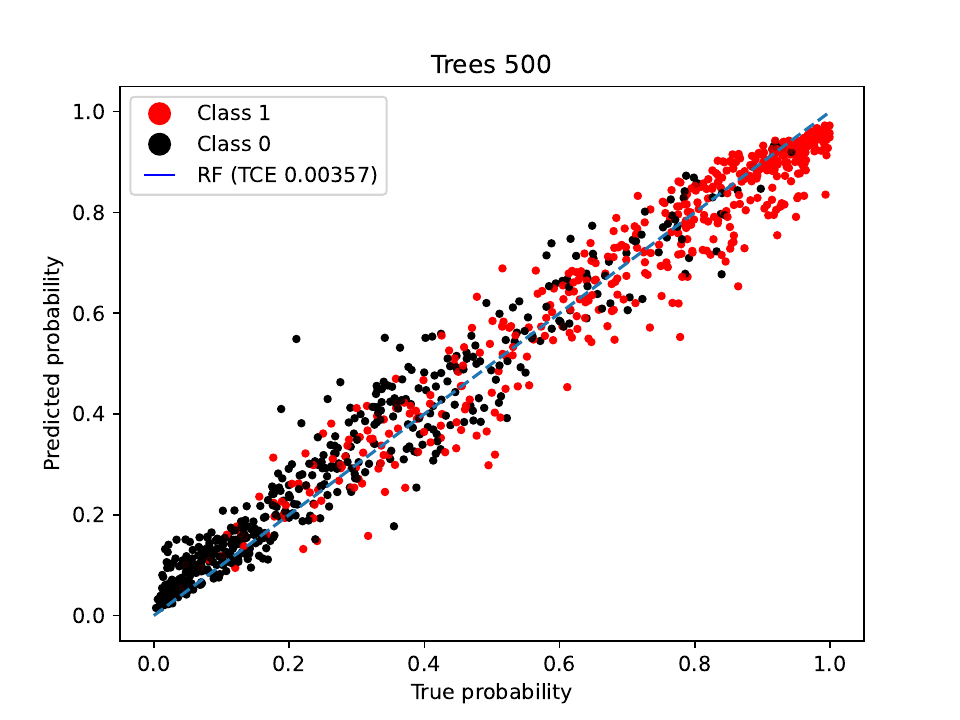}}

\caption{The effect of the number of trees in an RF on calibration represented as a reliability diagram along with the TCE of each forest. From left to right, the number of trees are set to 5, 50, and 500.}
\label{fig:tree}
\end{center}
\end{figure*}

We demonstrated that variations in the hyper-parameter Maximum Depth significantly influence the calibration performance. Thus, conducting hyper-parameter optimization seems to be a promising approach for enhancing RF calibration. A forest consists of multiple decision trees, and the number of trees is one of the most important hyper-parameters. In RF, each tree is trained on a bootstrapped version of the original training data, along with a subset of the features selected at random. Performing the bootstrap ensures diversity between the trees in a forest, and more trees lead to more diversity, which should improve the overall performance.

This is confirmed by Figure \ref{fig:tree}, which shows three RFs side by side trained on the same synthetic data shown in Figure \ref{fig:synthetic_data_1} with 5, 50, and 500 trees, respectively. We can see the reduction in noise around the diagonal calibration line as we increase the number of trees. Both visually and regarding the true calibration error value shown in the plots, it is clear that more trees yield a better calibrated forest.

\subsection{Data Manipulation with Synthetic Data}
\label{apx:exp_SDM}

In this section, we would like to get a deeper understanding of how different post-calibration methods perform under changes to the underlying synthetic data. We provide additional information on the already introduced Overlapping Distributions experiment in section \ref{sec:expGO} as well as a second experiment on calibration set size.

As introduced in the main paper, RF\_d, RF\_opt and RF\_large are also added as baselines. For RF\_opt we have chosen the hyper-parameters that result in the most significant changes in the output probability distributions of RF. The details of these hyper-parameters are provided in Table \ref{tab:rf_hyperopt_params}. We demonstrated that variations in the hyper-parameters significantly influence the calibration performance. Thus, conducting hyper-parameter optimization seems to be a promising approach for enhancing RF calibration.

\begin{table}[!ht]
    \centering
    \caption{Random Forest Search Space}
    \begin{tabular}{|l|l|}
        \hline
        \textbf{hyper-parameter} & \textbf{Values} \\
        \hline
        Number of trees & 100 \\
        \hline
        Criterion & [gini, entropy] \\
        \hline
        Maximum depth & [2, 3, ..., 100] \\
        \hline
        Minimum samples split & [2, 2, ..., 10] \\
        \hline
        Minimum samples leaf & [1, 2, ..., 10] \\
        \hline
        Maximum features & [sqrt, log2, None] \\
        \hline
        Class Weight & [None, balanced, balanced\_subsample] \\
        \hline
        Bootstrap & True \\
        \hline
        Laplace & [False, True] \\
        \hline
    \end{tabular}
    \label{tab:rf_hyperopt_params}
\end{table}

\subsubsection{Overlapping Distributions}
\label{apx:exp_Over}

% exp run names 100 tree
% 2D  1726176638_Overlap_fix100tree_2D
% 5D  1726193690_Overlap_fix100tree_5D
% 10D 1726214273_Overlap_fix100tree_10D
% 20D 1726244802_Overlap_fix100tree_20D

To generate the synthetic dataset used in this experiment, we modified the described synthetic dataset as follows: a complete overlap between the classes is initially achieved by setting the mean vector for both distributions to a single vector sampled uniformly from values between 0 and 1. Over 20 steps, a constant value is incrementally added to all elements of the mean vector for class 1, gradually separating this distribution from class 0 until there is no overlap between the two distributions. Since adding a constant value to the mean vector of a higher-dimensional multivariate Gaussian distribution reduces overlap more significantly than in lower-dimensional distributions, we adjust the constant value for each of the 2, 5, 10, and 20-dimensional datasets. This adjustment ensures that when the two distributions no longer overlap, the Bhattacharyya distance \citep{bhattacharyya1943measure} reaches a value of 5.72.

\begin{figure*}[ht]
\begin{center}

  \subfloat{\includegraphics[scale=0.27]{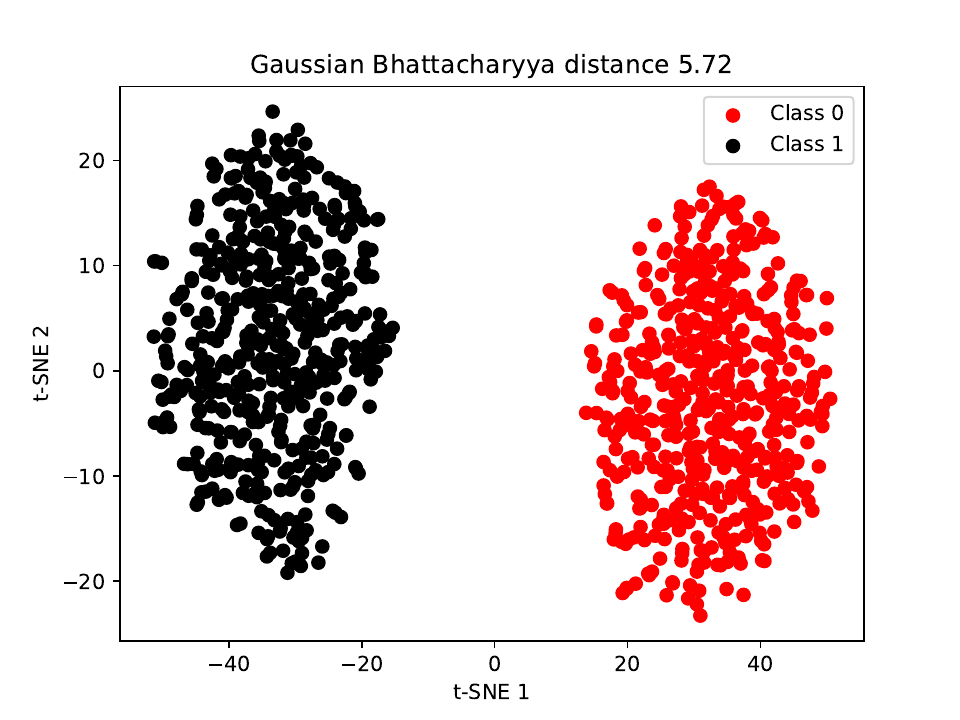}}
  \subfloat{\includegraphics[scale=0.27]{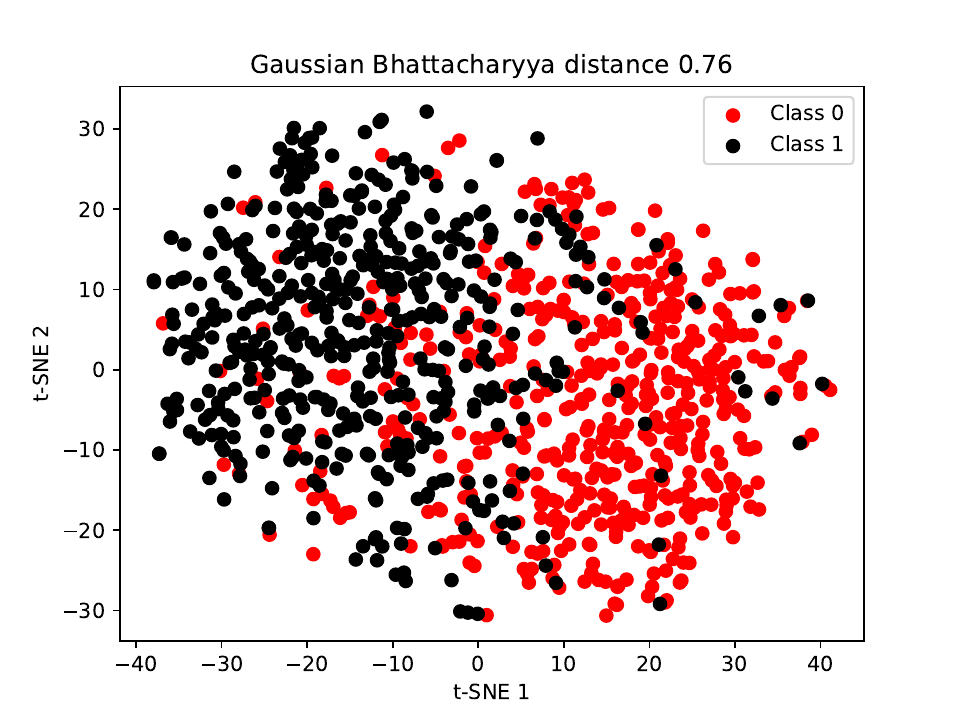}}
  \subfloat{\includegraphics[scale=0.27]{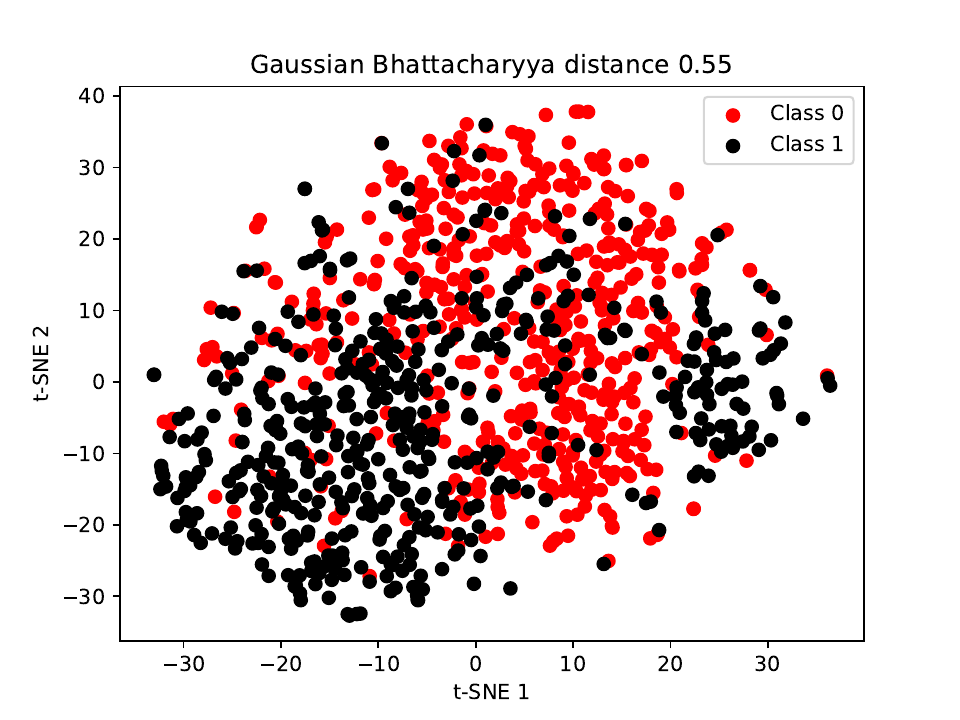}}
  \subfloat{\includegraphics[scale=0.27]{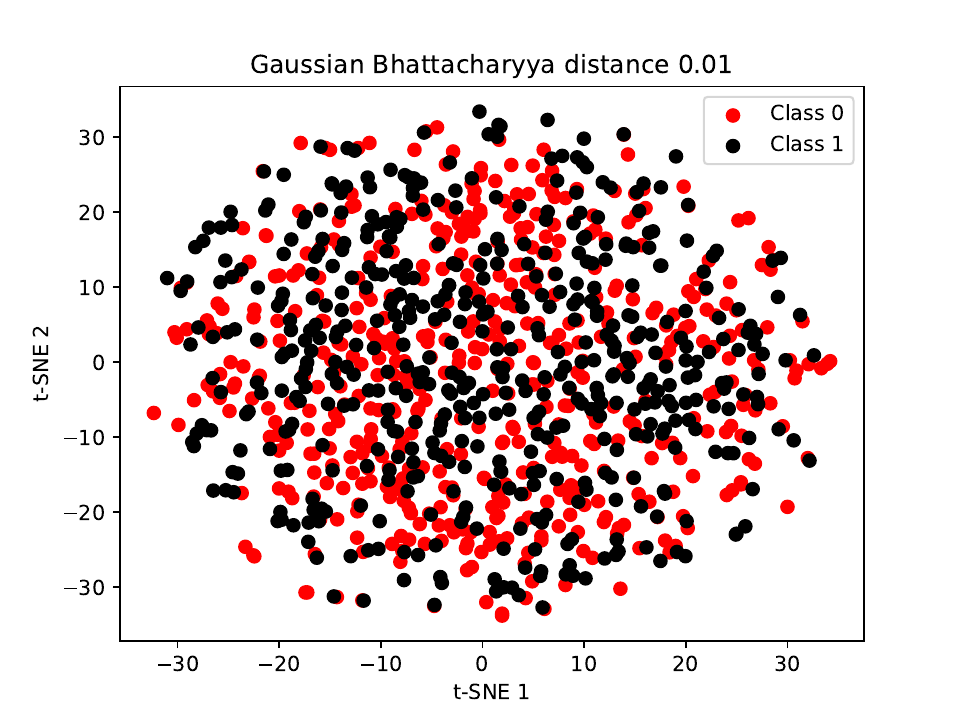}}

\caption{A 2D t-SNE visualization of the 10-dimensional synthetic data, depicting the variation in overlap between the two classes.}
\label{fig:o_data}
\end{center}
\end{figure*}

Figure \ref{fig:o_data} shows a 2D t-SNE visualization of the 10-dimensional synthetic data, depicting four scenarios ranging from no overlap (left) to complete overlap (right) between the two classes.

\begin{figure*}[ht]
\begin{center}

  \subfloat{\includegraphics[scale=0.27]{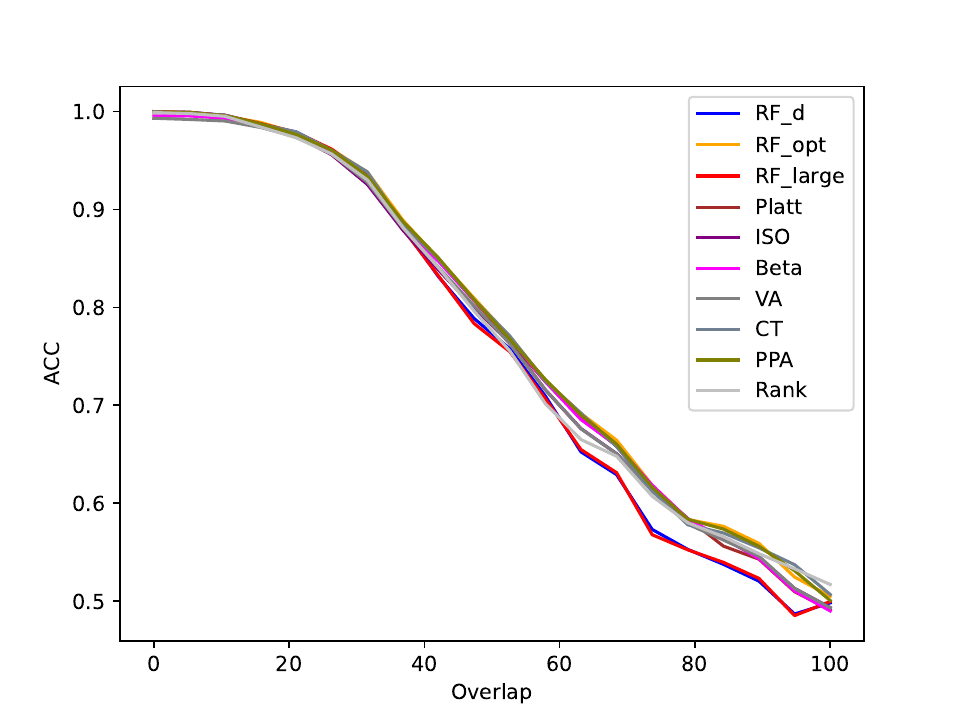}}
  \subfloat{\includegraphics[scale=0.27]{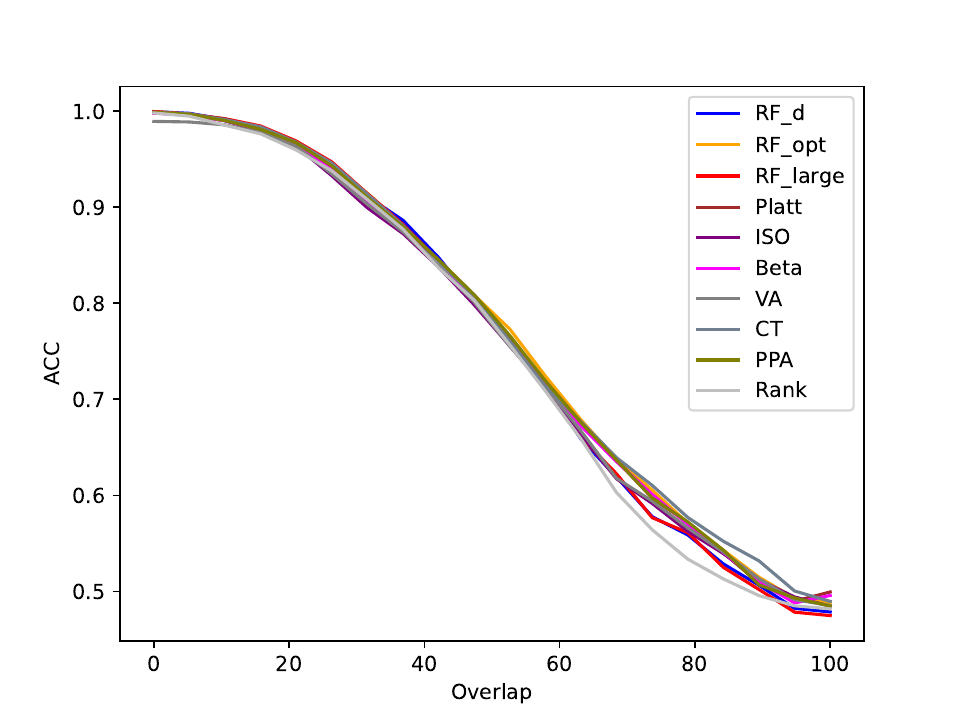}}
  \subfloat{\includegraphics[scale=0.27]{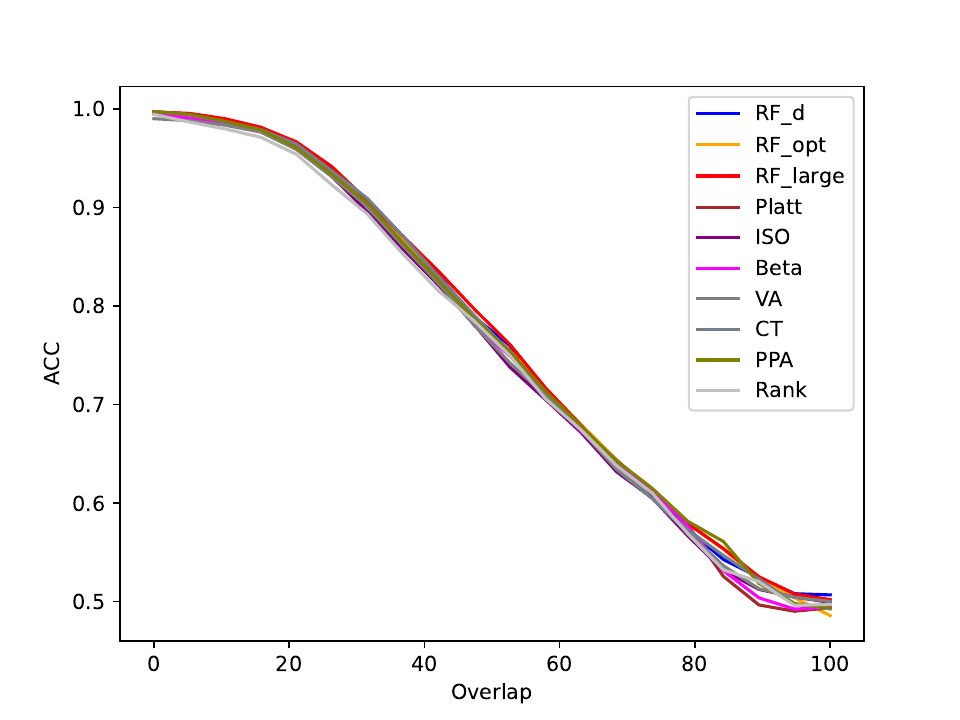}}
  \subfloat{\includegraphics[scale=0.27]{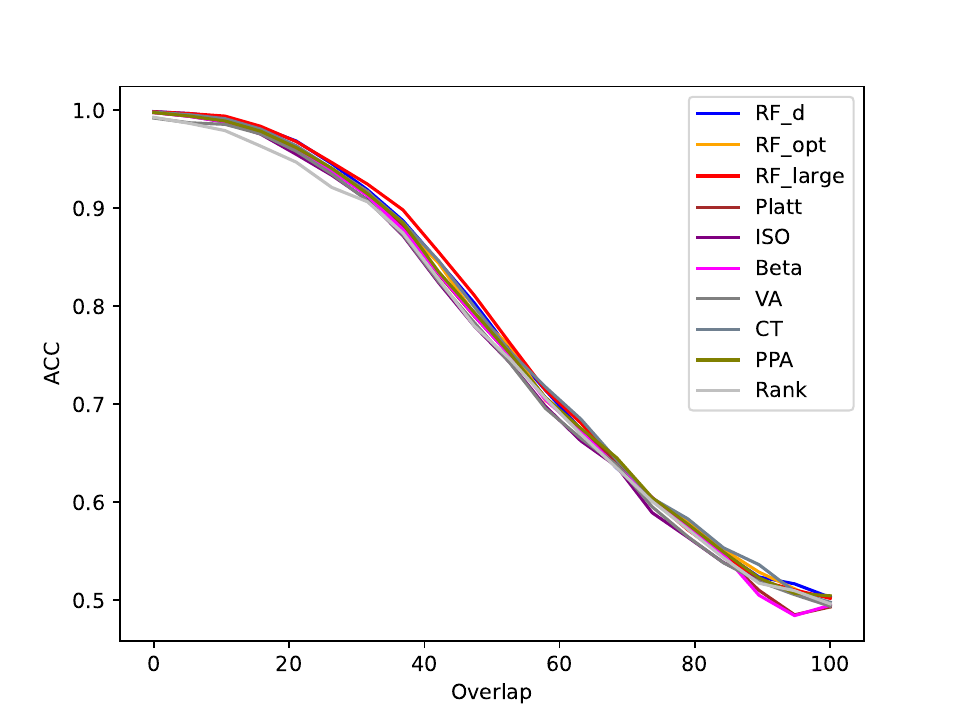}}

  \subfloat{\includegraphics[scale=0.27]{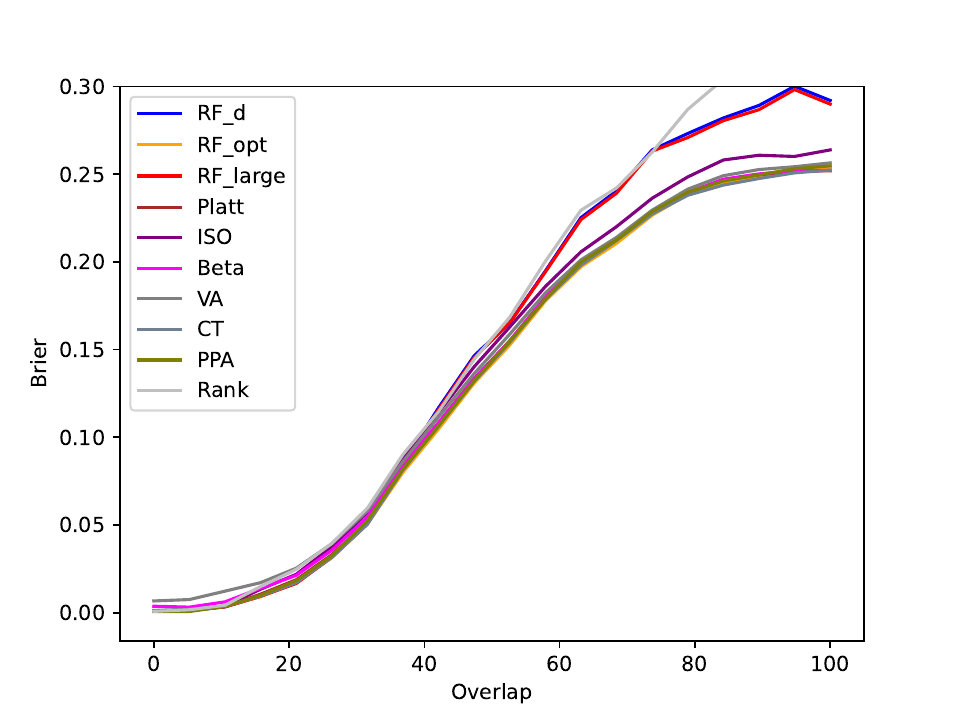}}
  \subfloat{\includegraphics[scale=0.27]{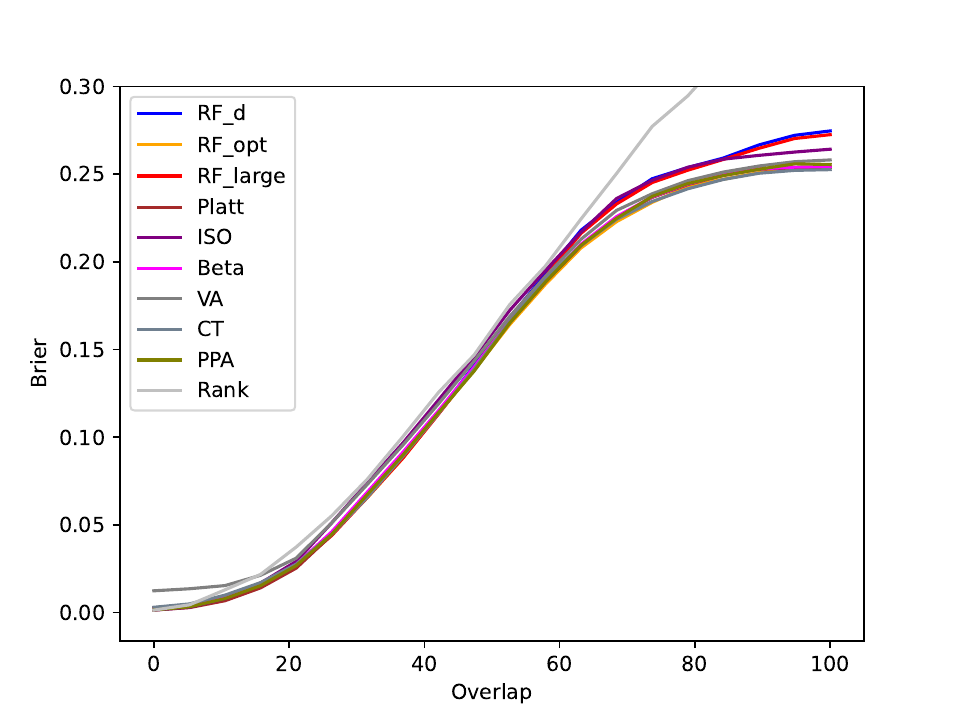}}
  \subfloat{\includegraphics[scale=0.27]{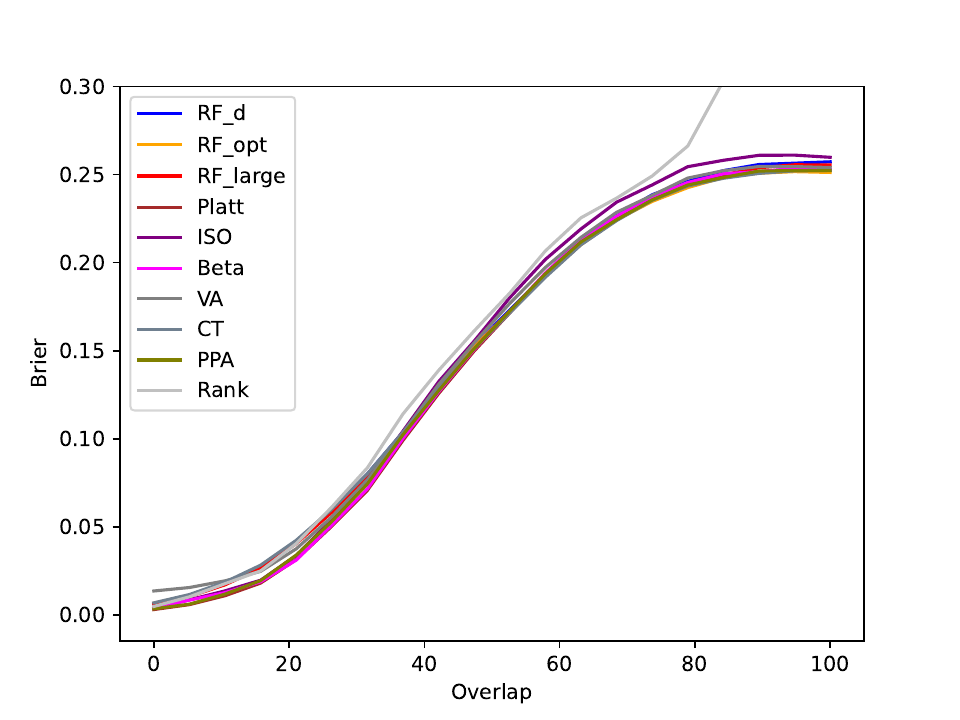}}
  \subfloat{\includegraphics[scale=0.27]{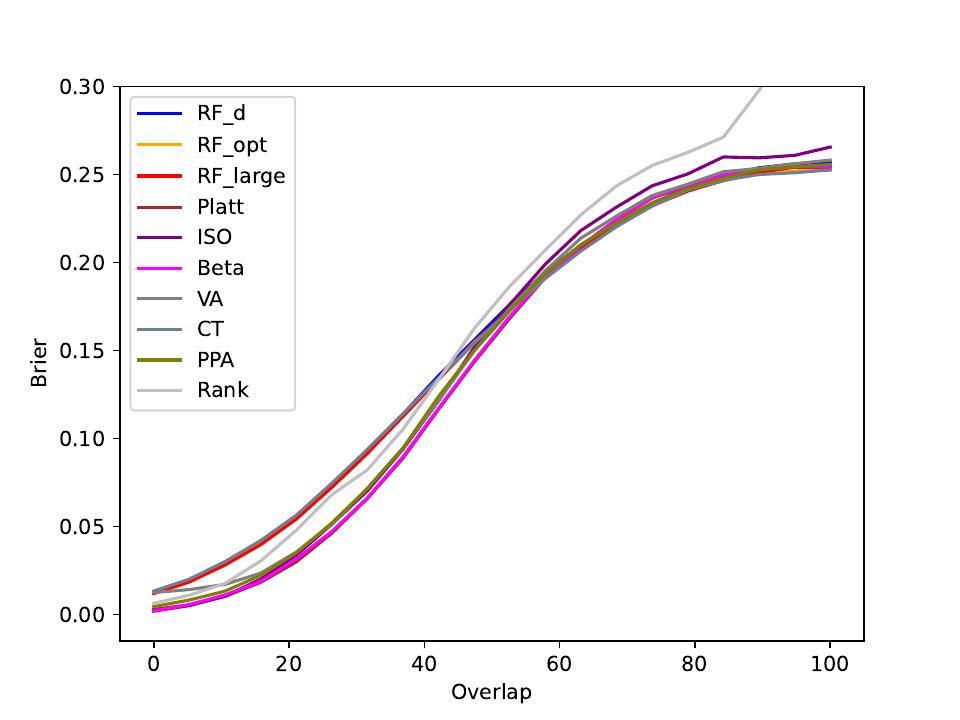}}

  \subfloat{\includegraphics[scale=0.27]{images/o2D_tce_mse.pdf}}
  \subfloat{\includegraphics[scale=0.27]{images/o5D_tce_mse.pdf}}
  \subfloat{\includegraphics[scale=0.27]{images/o10D_tce_mse.pdf}}
  \subfloat{\includegraphics[scale=0.27]{images/o20D_tce_mse.pdf}}

  \subfloat{\includegraphics[scale=0.27]{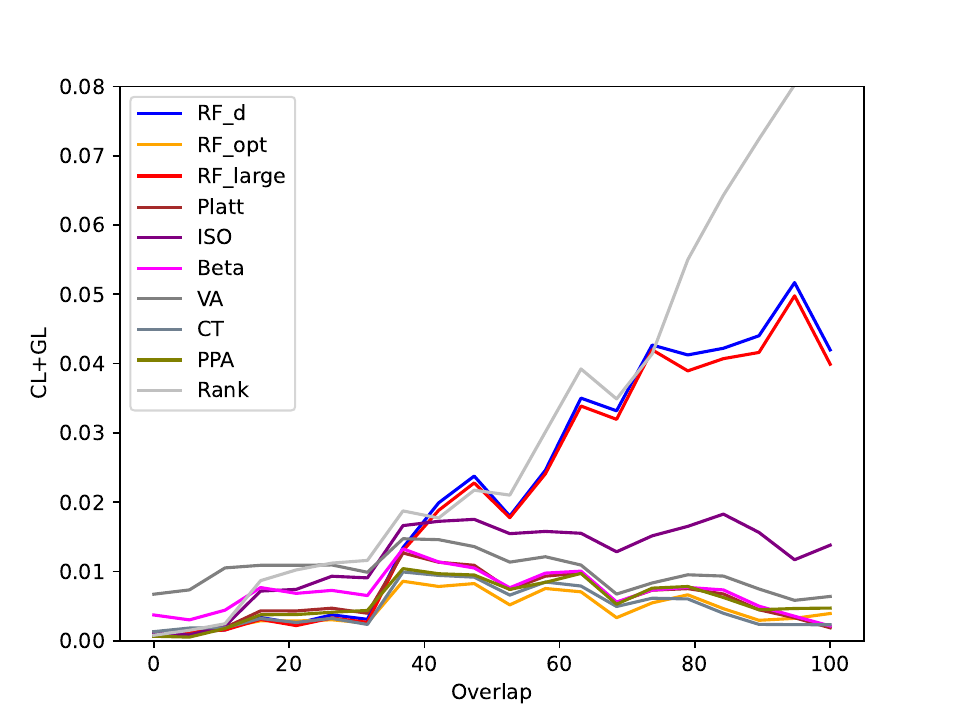}}
  \subfloat{\includegraphics[scale=0.27]{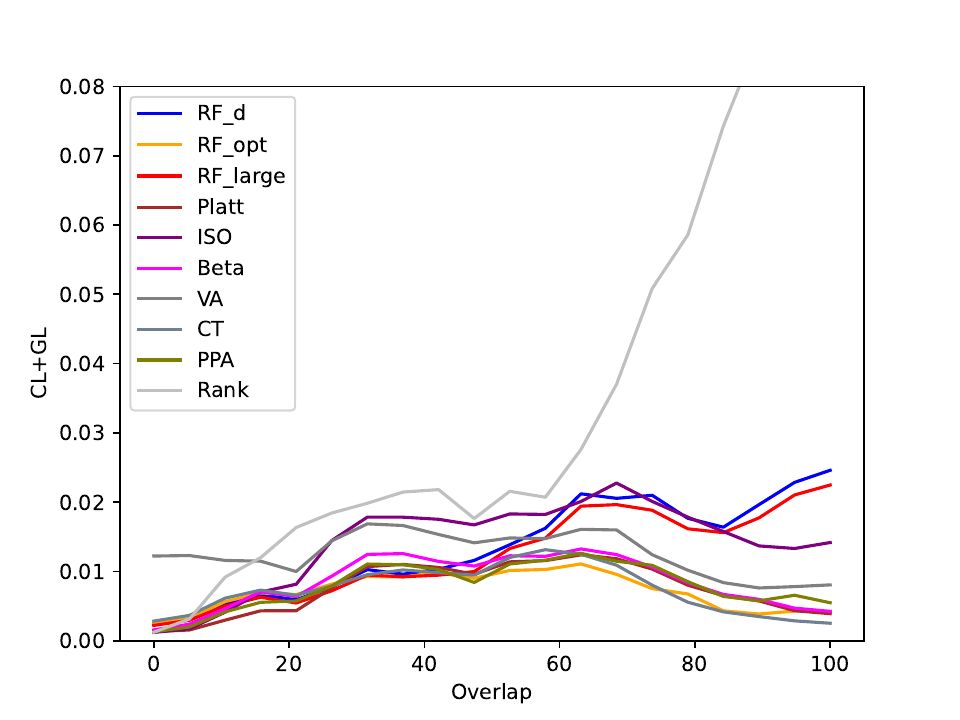}}
  \subfloat{\includegraphics[scale=0.27]{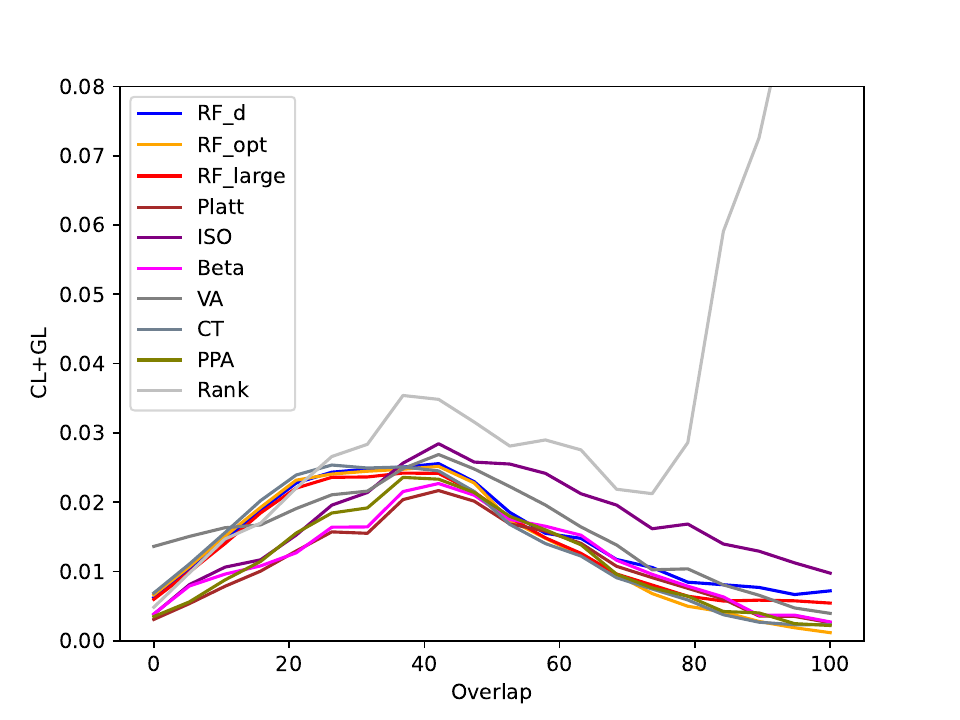}}
  \subfloat{\includegraphics[scale=0.27]{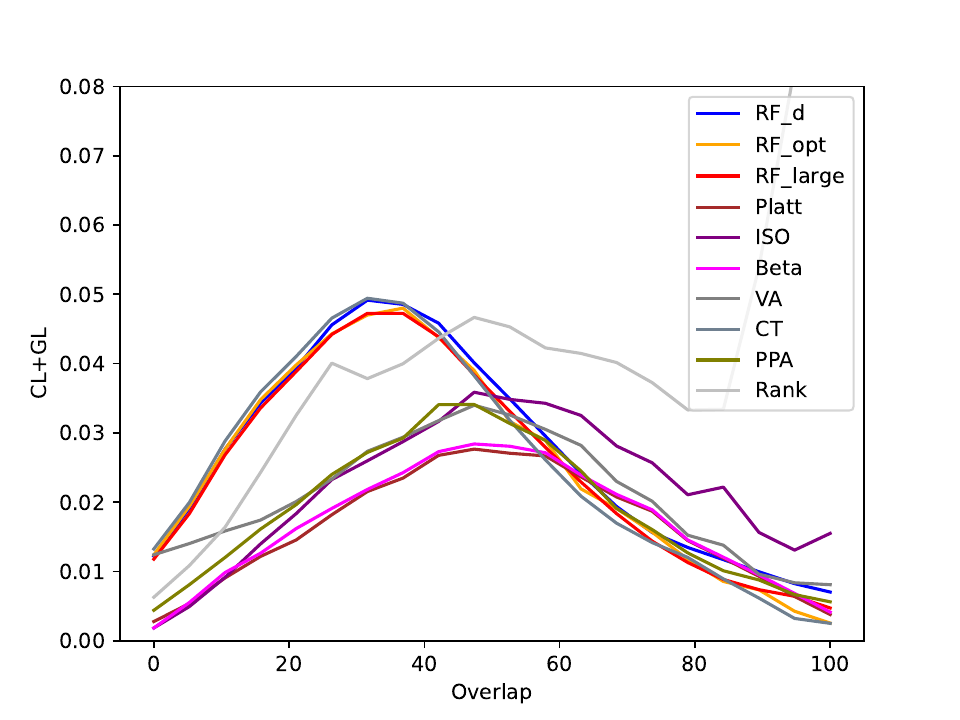}}

  \subfloat{\includegraphics[scale=0.27]{images/o2D_ece.pdf}}
  \subfloat{\includegraphics[scale=0.27]{images/o5D_ece.pdf}}
  \subfloat{\includegraphics[scale=0.27]{images/o10D_ece.pdf}}
  \subfloat{\includegraphics[scale=0.27]{images/o20D_ece.pdf}}

  \subfloat{\includegraphics[scale=0.27]{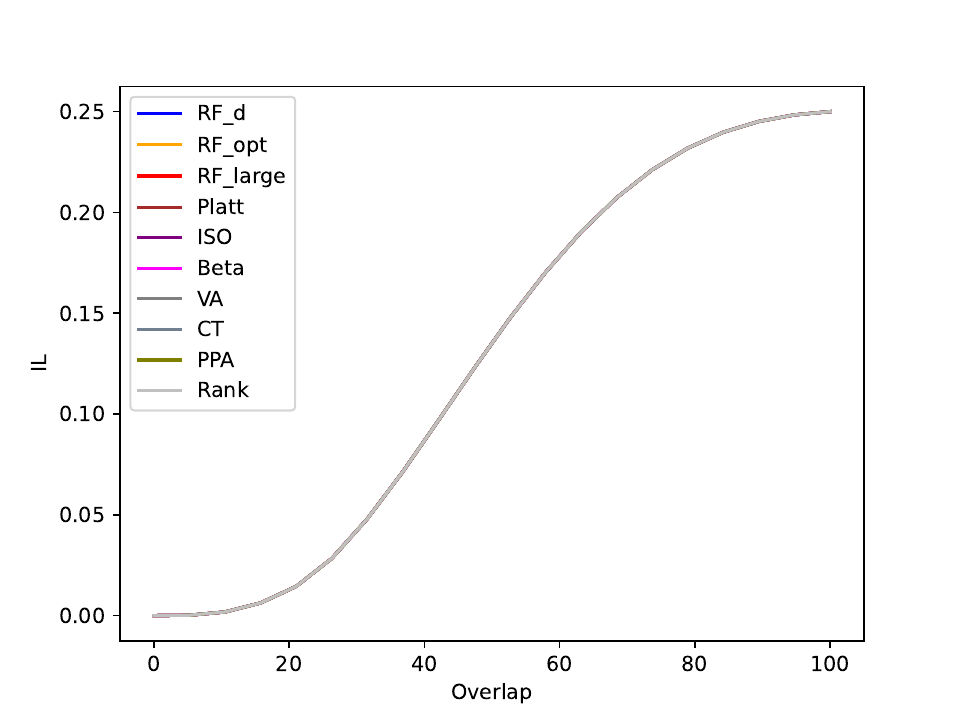}}
  \subfloat{\includegraphics[scale=0.27]{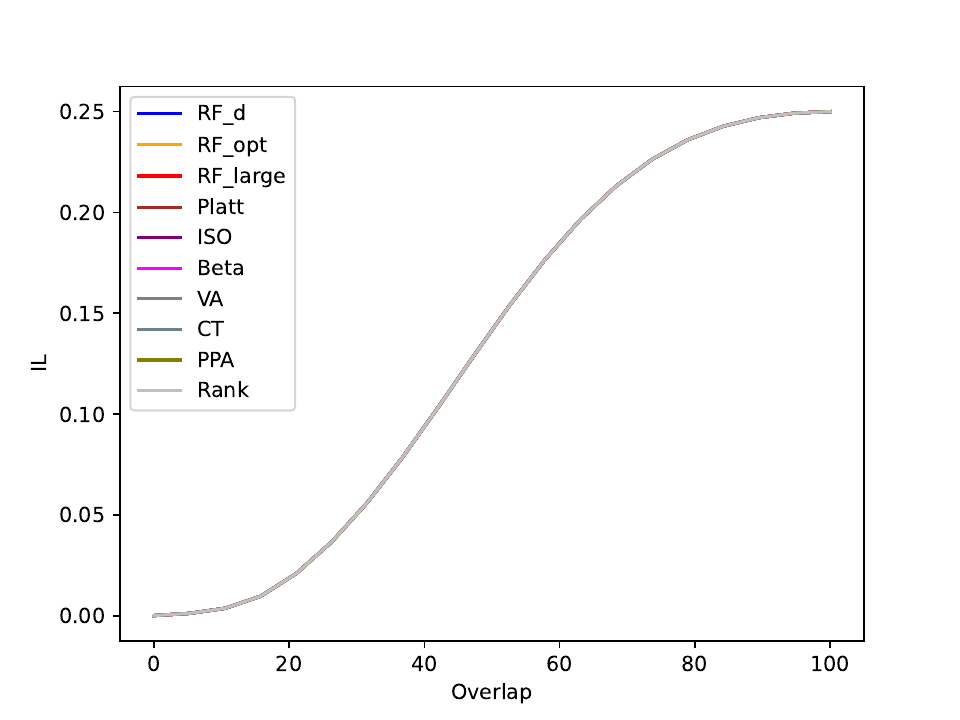}}
  \subfloat{\includegraphics[scale=0.27]{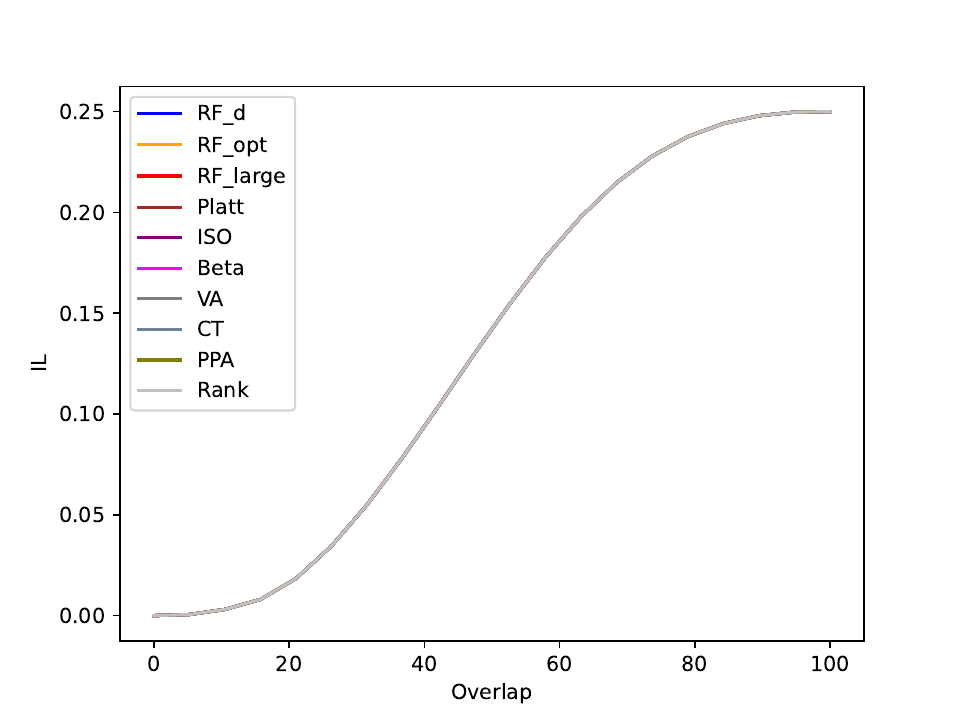}}
  \subfloat{\includegraphics[scale=0.27]{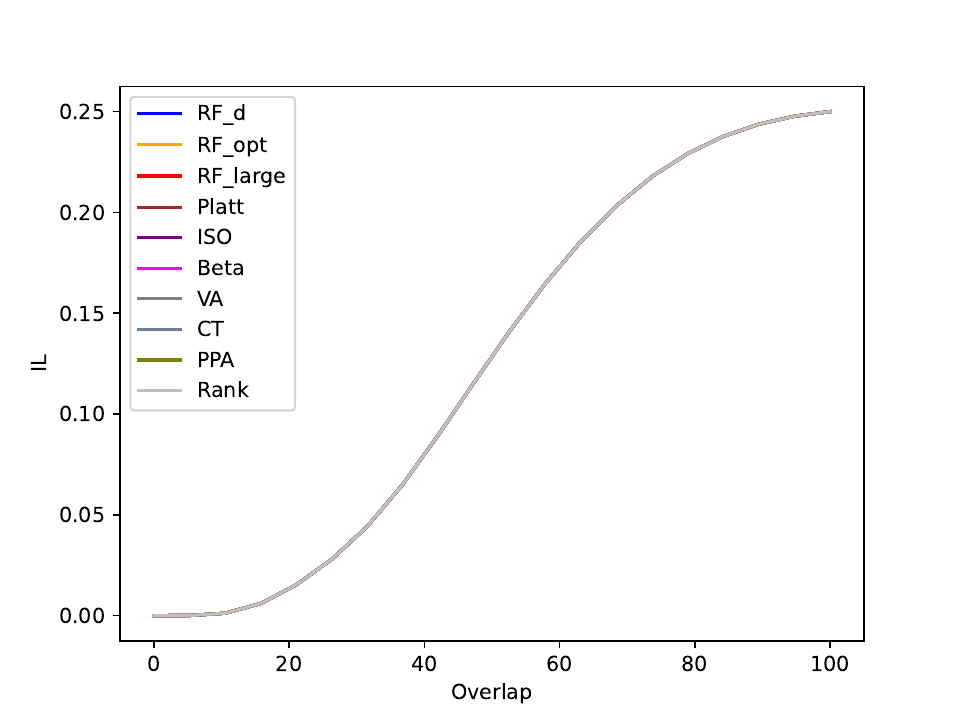}}

\caption{The impact of varying the overlap between two Gaussian distributions on the performance of calibration methods, analyzed using synthetic data of increasing dimensions. The results are presented in columns corresponding to each dimensionality -2, 5, 10, and 20- from left to right, while the performance metrics -Accuracy, Brier score, TCE, CL+GL, ECE, and IL- are displayed in rows from top to bottom.}
\label{fig:overlap_s}
\end{center}
\end{figure*}

The extended average performance of the calibration methods for the overlapping distributions experiment, including accuracy, Brier score, CL$+$GL, and IL is shown in Figure \ref{fig:overlap_s}. As mentioned, calibration is easier if ground-truth probabilities are extreme and more difficult for close-to-uniform distributions, which is why we see a small deterioration of TCE in the middle. This is also confirmed by the CL$+$GL curves, which first increase and then decrease again. Note that Brier score is monotonically increasing, due to the increase in irreducible uncertainty. For example, in the case of full overlap, even the best prediction (the uniform distribution) has an expected Brier score of $1/4$. 

% plot reults

\clearpage

\subsubsection{Calibration Set Size}
\label{sec:expCS}
% exp run name 1726071621_synthetic_mg_100tree_100runs 
% old exp run name 1721940149_synthetic_mg - data is mixture gaussian

One of the important factors in each post-calibration method is the amount of data required to train the calibrator to output properly calibrated probability distribution. Since we are working with synthetic datasets for this experiment, we can fix the training and test set size and, by manipulating the size of the calibration set, observe the differences between each post-calibration method's performance in terms of TCE. 

\begin{figure*}[ht]
\begin{center}

  \includegraphics[scale=0.70]{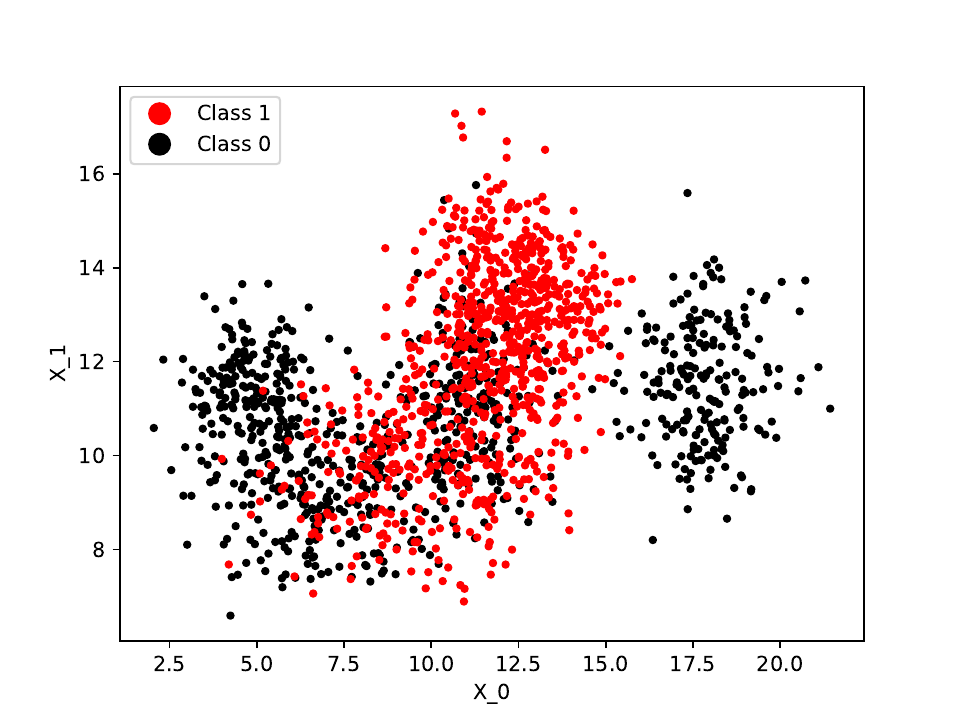}

\caption{Synthetic binary dataset generated from a mixture of 4 Gaussian distributions per class.  The axes of the plot represent the two features of this dataset.}
\label{fig:synthetic_data_mg}
\end{center}
\end{figure*}

To make the synthetic dataset more challenging, we used a mixture of Gaussian distributions with 4 clusters per class and doubled the sample size, which allows us to allocate a calibration dataset matching the size of the training set  used for training the RF. Figure \ref{fig:synthetic_data_mg} depicts the synthetic dataset generated from the mixture of Gaussian distributions. 

The mixture of Gaussian distributions is defined as follows: For the specified number of clusters per class, we generate a multivariate Gaussian distribution. The mean values are sampled from a uniform distribution between 0 and 20, while the diagonal values of the covariance matrix are uniformly drawn from a range between 1 and 5. 

\begin{figure*}[ht]
\begin{center}

  \subfloat{\includegraphics[scale=0.4]{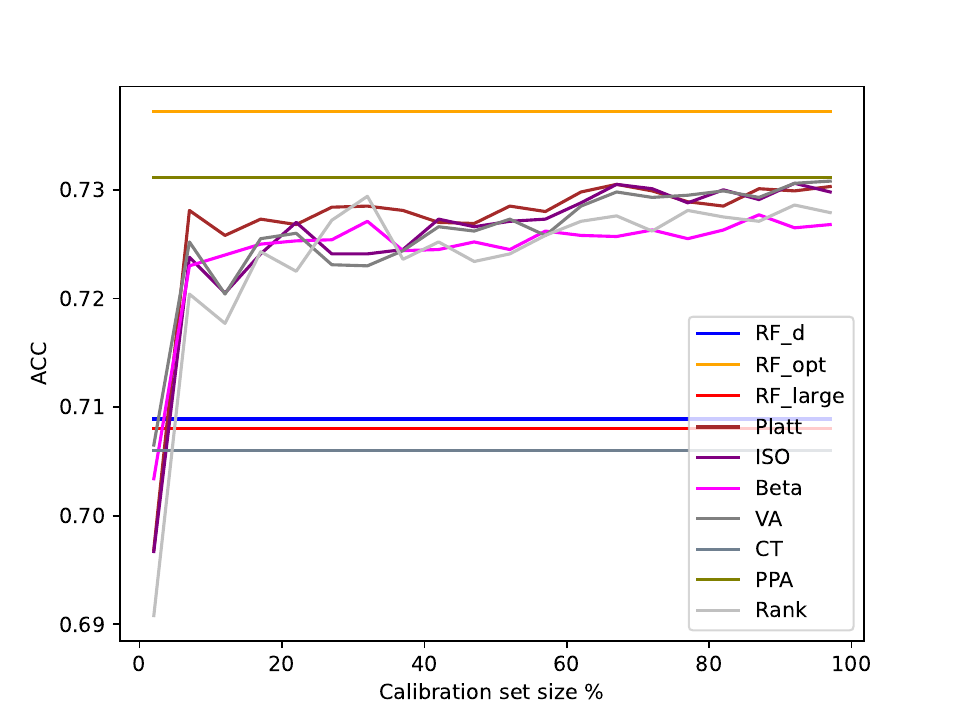}}
  \subfloat{\includegraphics[scale=0.4]{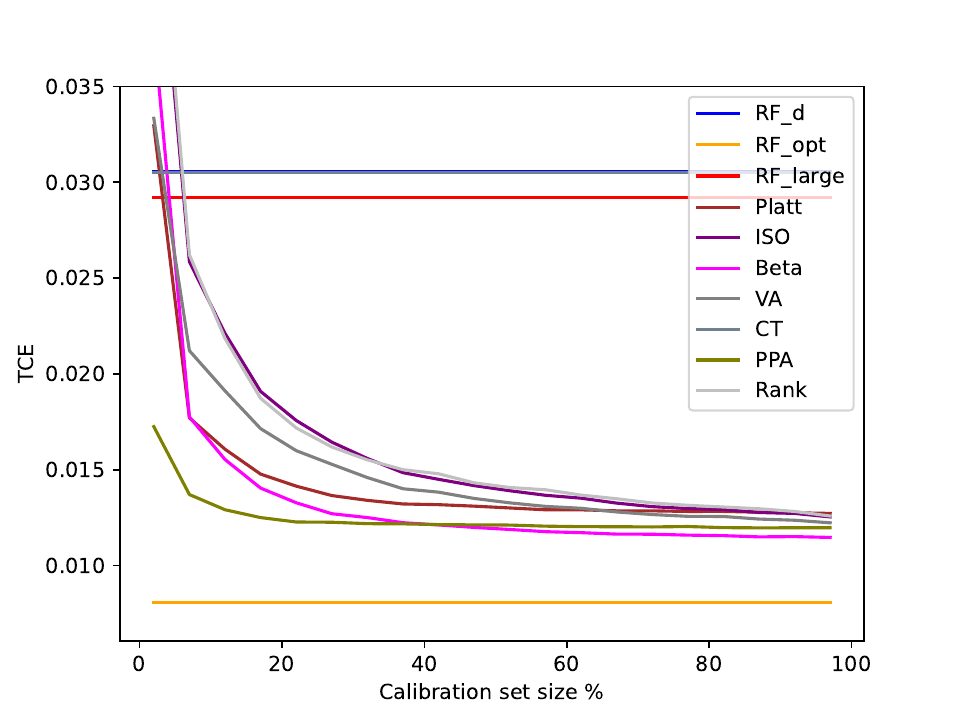}}

  \subfloat{\includegraphics[scale=0.4]{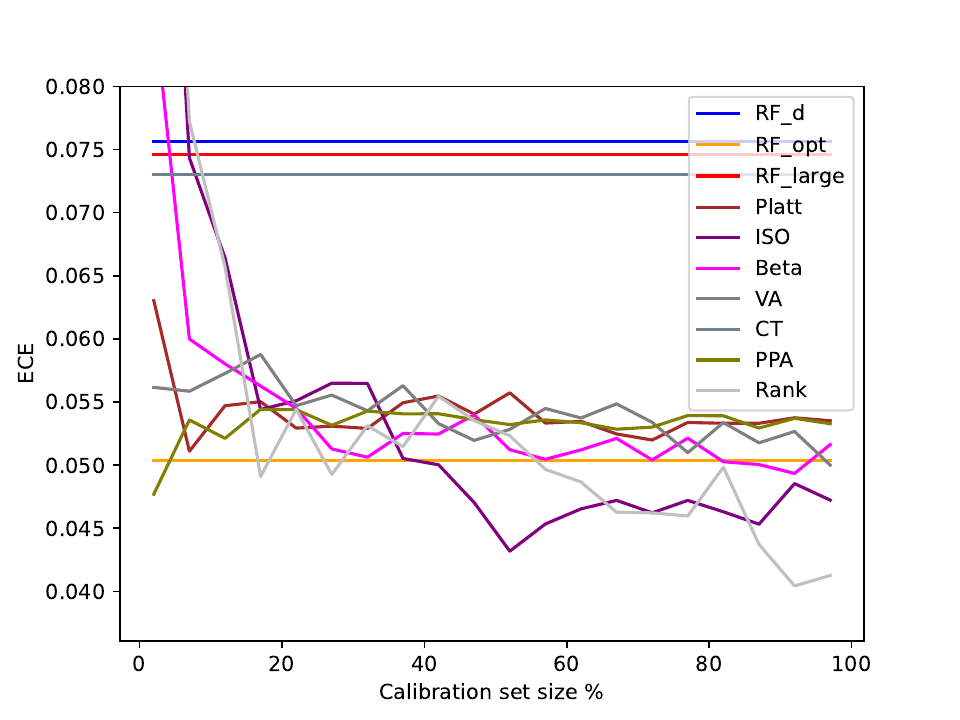}}
  \subfloat{\includegraphics[scale=0.4]{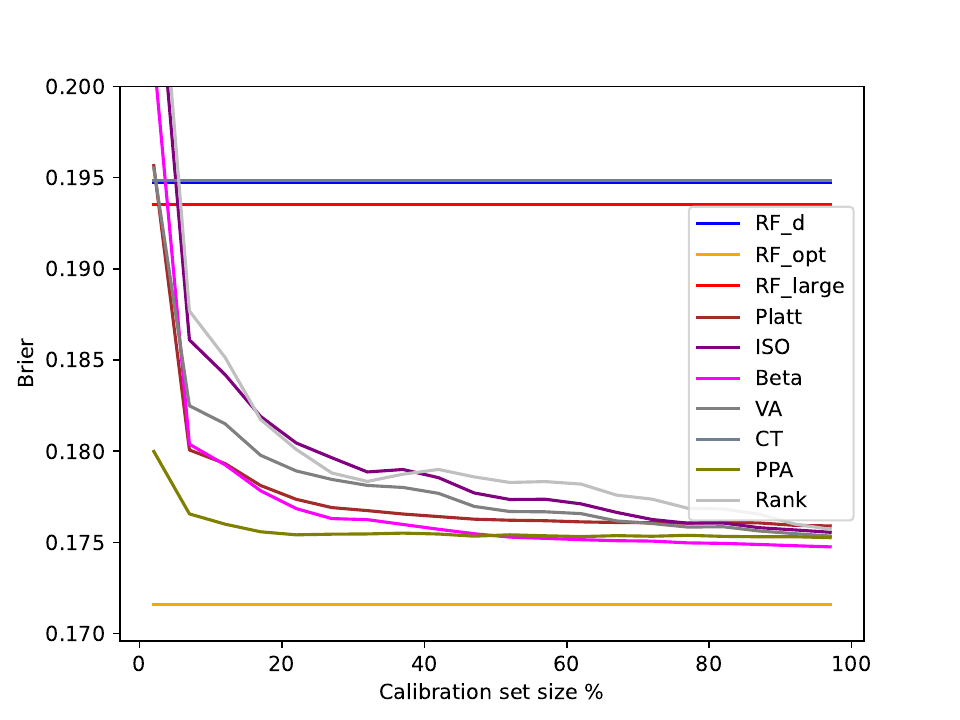}}

\caption{The impact of adjusting the calibration set size on the performance of post-calibration methods in terms of accuracy, TCE, ECE, and Brier score. 
}
\label{fig:calib_size}
\end{center}
\end{figure*}

Figure \ref{fig:calib_size} illustrates how adjusting the calibration set size affects the performance of post-calibration methods for accuracy, TCE, Brier score, and ECE with 20 bins of equal width. We also included the additional three RF variants as baselines. The calibration set size is represented as a percentage of the training data, varying from 2 to 100 percent. The experiment is run 100 times and the average performance is reported. As expected, the calibration error decreases when increasing the data available for calibration. 

% calib size exp 100 trees

Interestingly, even with a calibration set matching the training data size, no calibration method achieves a TCE as low as RF\_opt (top right plot in Figure \ref{fig:calib_size}). This reaffirms the importance of optimizing hyper-parameters, such as max-depth.

Furthermore, we can rank the calibration methods based on the amount of data needed to achieve a stable low calibration error. The ranking from most to least data-efficient based on the TCE metric is as follows: PPA, Beta, Platt, Venn-Abers, ISO, and Rank method. This ranking reflects the parametric vs.\ non-parametric nature of these methods: Simple parametric methods such as PPA can be trained effectively with only a small portion of the calibration set, whereas non-parametric methods such as isotonic regression demand a much larger dataset. On the other side, as already said, parametric methods may produce biased results if their model assumptions are not satisfied. Indeed, the results also reflect this bias-variance tradeoff: Compared to its competitors, ISO under-performs in the beginning (with little calibration data) but performs strongly in the end (with more calibration data). 

The results for ECE may look surprising at first sight. One has to be a bit careful with the interpretation of this metric, however, because the (expected) ECE can be strongly influenced by the distribution of data points into bins: The more imbalanced this distribution, the higher becomes the average number of data points on which a prediction is based, and hence the better the predictions tend to be. In the extreme case, a method may always predict more or less the same probability and hence put all probabilities into the same bin\,---\,the true expectation for this bin can then be estimated in a statistically stable way, and much more precisely than the expectation of bins with only a few data points. To give a rough idea, Table \ref{tab:entropy} shows the entropy of the distribution over bins per method (the lower, the more imbalanced), averaged over the different experiments. These numbers are quite coherent with the performance in terms of ECE. 

\begin{table}[ht]
\centering
\caption{Average entropy of the predicted probability distributions from all calibration methods in the Calibration Set Size experiment.}
\label{tab:entropy}
\begin{tabular}{lrrrrrrrrrr}
\toprule
{} &     RF\_d &   RF\_opt &  RF\_large &    Platt &      ISO &     Beta &       VA &        CT &      PPA &     Rank \\
   &          &          &           &          &          &          &          &           &          &          \\
\midrule
Entropy & 5.60 & 6.50 & 6.29 & 6.54 & 3.30 & 6.54 & 4.22 & 5.61 & 6.54 & 3.27 \\

\bottomrule
\end{tabular}
\end{table}

\subsection{Assessing Calibration Performance on Real datasets}
\label{apx:exp_real}
% exp run name 1725451866_paper10CV100tree
% old run name 1721403432_Real_opt_seed_fix2

This section provides more information on the experiments of the section \ref{sec:exp_real}. The 30 openly available real datasets from UCI and PROMISE repositories are presented in Table \ref{tab:data}. 

Each of the tables \ref{tab:acc}, \ref{tab:brier}, \ref{tab:log-loss}, and \ref{tab:ece} display the value of the evaluation measures for all 30 datasets across the mentioned calibration methods and the three RF variants RF\_d, RF\_opt, and RF\_large. Additionally, the mean\footnote{We include the mean performance of all the datasets for comparison purposes, as it is also reported in related works. Theoretically, averaging performance over datasets is clearly debatable.} performance and ranking for each method are presented in the last two rows of each table.

In terms of accuracy, Brier score, and log-loss as seen in Tables \ref{tab:acc}, \ref{tab:brier} and \ref{tab:log-loss}, RF\_large and RF\_opt lead in performance based on average rankings. In terms of ECE, as shown in Table \ref{tab:ece}, VA achieves the best performance, followed by RF\_opt and Curtailment. However, based on our experiments with synthetic datasets, ECE should be interpreted with caution.

\begin{figure*}[ht]
\begin{center}

  \includegraphics[scale=0.55]{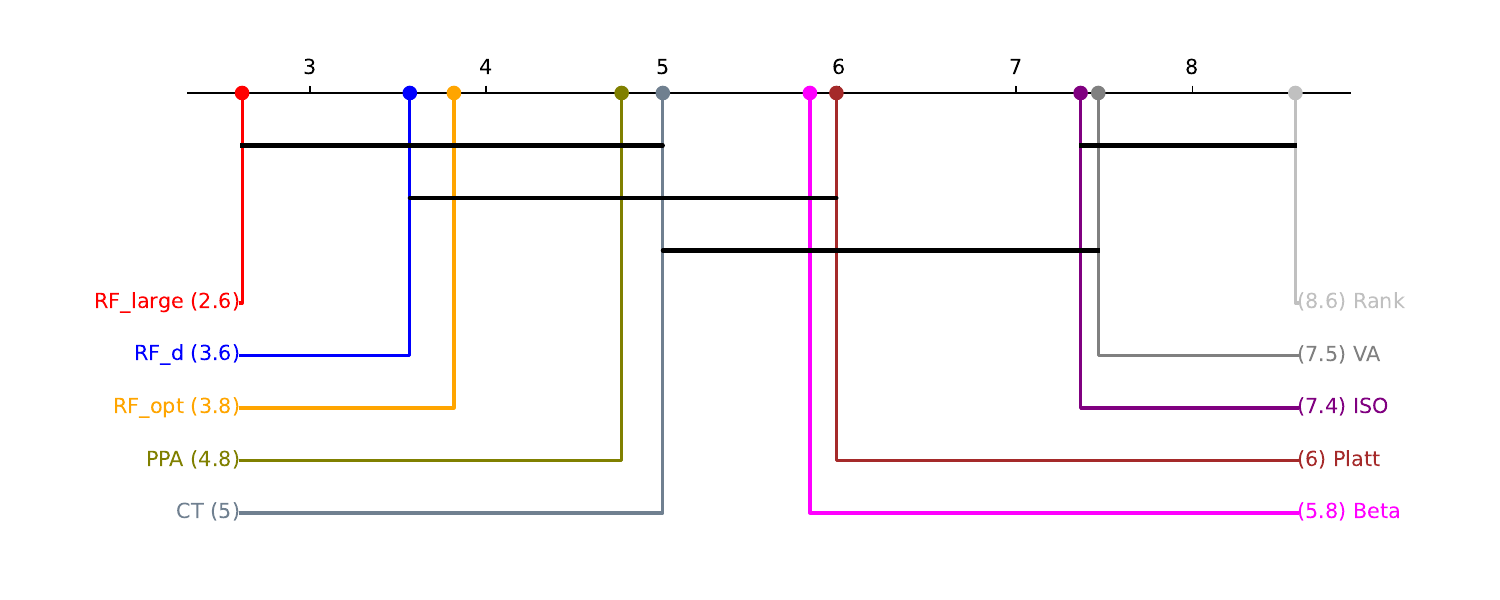}

\caption{Critical difference diagram of 30 real datasets on accuracy.}
\label{fig:cd_acc}
\end{center}
\end{figure*}

\begin{figure*}[ht]
\begin{center}

  \includegraphics[scale=0.55]{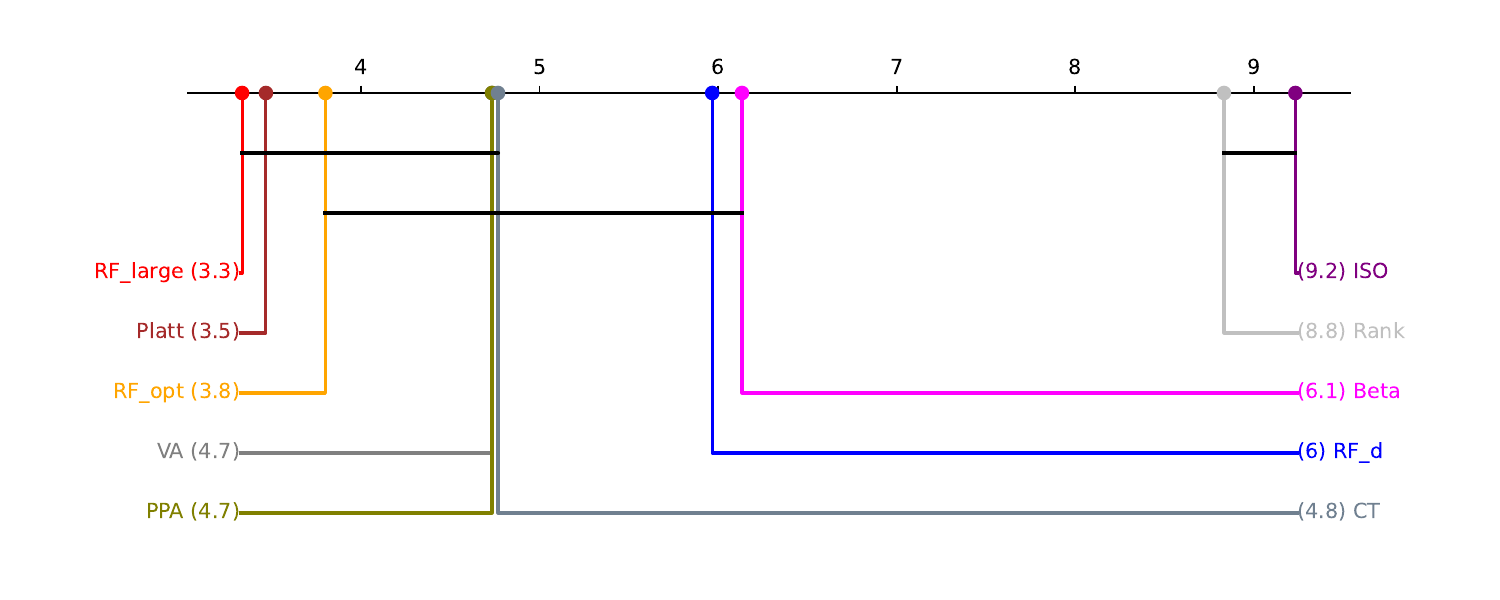}

\caption{Critical difference diagram of 30 real datasets on log-loss.}
\label{fig:cd_log}
\end{center}
\end{figure*}

Additionally, Figures \ref{fig:cd_acc} and \ref{fig:cd_log} show the critical difference diagram of the Nemenyi-Friedman statistical test using a significance level of 0.05. The null hypothesis posits that there is no statistical difference between any two calibration methods.

% To further examine the differences in performance between calibration methods, we applied a Nemenyi-Friedman test for statistical significance \citep{demvsar2006statistical}, using a significance level of 0.05. The null hypothesis posits that there is no statistical difference between any two calibration methods. The results of these tests can be visualized using a critical difference diagram. In this diagram, methods for which the null hypothesis cannot be rejected are grouped together.

% The top-performing group varies depending on the metric being considered. For example, as shown in Figure \ref{fig:cd_brier}, the top group for the Brier score includes RF\_large, RF\_opt, RF\_d, CT, and PPA. When looking at log-loss (Figure \ref{fig:cd_log}), VA is also included in the top group. Lastly, the top grouping in the critical difference diagram for ECE, illustrated in Figure \ref{fig:cd_ece} are VA, RF\_opt, and CT. It also shows that the calibration methods are more closely clustered, indicating only minimal significant statistical differences among most of the methods in terms of ECE. This finding further underscores that ECE is not the most accurate approximation of the probability-wise calibration error.

% dataset info
\begin{table}[ht]
\centering
\caption{30 Real datasets from UCI and PROMISE Repositories}
\label{tab:data}

\begin{tabular}{llrrr}
\toprule
\# &                  Name &  instances &  features & \% of majority class \\
\midrule
1  &                      datatrieve &        130 &         8 &         91.5 \\
2  &  kc1\_class\_level\_defectiveornot &        145 &        94 &         58.6 \\
3  &                      parkinsons &        195 &        22 &         75.4 \\
4  &            Sonar\_Mine\_Rock\_Data &        208 &        60 &         53.4 \\
5  &                           spect &        267 &        22 &         79.4 \\
6  &                          spectf &        267 &        44 &         79.4 \\
7  &              HRCompetencyScores &        300 &         9 &         53.0 \\
8  &                           heart &        303 &        13 &         54.5 \\
9  &                       vertebral &        310 &         6 &         67.7 \\
10 &                      ionosphere &        351 &        34 &         64.1 \\
11 &                             kc3 &        458 &        39 &         90.6 \\
12 &                             cm1 &        498 &        21 &         90.2 \\
13 &                             kc2 &        522 &        21 &         79.5 \\
14 &                            wdbc &        569 &        30 &         62.7 \\
15 &                          breast &        569 &        30 &         62.7 \\
16 &                        diabetes &        768 &         8 &         65.1 \\
17 &                            QSAR &       1055 &        41 &         66.3 \\
18 &                             pc1 &       1109 &        21 &         93.1 \\
19 &                      hillvalley &       1212 &       100 &         50.0 \\
20 &                            bank &       1372 &         4 &         55.5 \\
21 &                             pc4 &       1458 &        37 &         87.8 \\
22 &                             SPF &       1941 &        33 &         65.3 \\
23 &                             kc1 &       2109 &        21 &         84.5 \\
24 &                           scene &       2407 &       299 &         82.1 \\
25 &                  Customer\_Churn &       3150 &        13 &         84.3 \\
26 &                        spambase &       4601 &        57 &         60.6 \\
27 &                            wilt &       4839 &         5 &         94.6 \\
28 &                         phoneme &       5404 &         5 &         70.7 \\
29 &                             jm1 &      10880 &        21 &         80.7 \\
30 &                             eeg &      14980 &        14 &         55.1 \\
\bottomrule
\end{tabular}
\end{table}

% 100 tree run name 1725451866_paper10CV100tree

% Real30 ACC
\begin{table}[ht]
\centering
\caption{Accuracy of calibration methods applied on RF trained on 30 real datasets.}

\label{tab:acc}
\resizebox{0.8\columnwidth}{!}{

\rotatebox{90}{
\begin{tabular}{lrrrrrrrrrr}
\toprule
{} &     RF\_d &   RF\_opt &  RF\_large &    Platt &      ISO &     Beta &       VA &       CT &      PPA &     Rank \\
Data                           &          &          &           &          &          &          &          &          &          &          \\
\midrule
cm1                            &  0.89400 &  0.90123 &   0.89401 &  0.89842 &  0.89475 &  0.89721 &  0.88352 &  0.90083 &  0.90044 &  0.88512 \\
datatrieve                     &  0.90154 &  0.89846 &   0.90462 &  0.90154 &  0.90615 &  0.87692 &  0.86615 &  0.91077 &  0.90462 &  0.91077 \\
kc1\_class\_level\_defectiveornot &  0.74486 &  0.73610 &   0.74210 &  0.71733 &  0.72543 &  0.73390 &  0.72533 &  0.73200 &  0.73771 &  0.72162 \\
kc1                            &  0.86003 &  0.85633 &   0.86325 &  0.85567 &  0.85596 &  0.85548 &  0.85615 &  0.84960 &  0.85510 &  0.84988 \\
kc2                            &  0.83401 &  0.83668 &   0.83513 &  0.83485 &  0.82988 &  0.83059 &  0.82870 &  0.83939 &  0.83480 &  0.82750 \\
kc3                            &  0.89782 &  0.89824 &   0.89651 &  0.89912 &  0.89389 &  0.89738 &  0.88822 &  0.89345 &  0.89913 &  0.90527 \\
pc1                            &  0.93815 &  0.93545 &   0.93851 &  0.93671 &  0.93310 &  0.93436 &  0.93707 &  0.93527 &  0.93671 &  0.93292 \\
spect                          &  0.82339 &  0.83761 &   0.82493 &  0.81946 &  0.80900 &  0.82108 &  0.81578 &  0.82556 &  0.83162 &  0.79860 \\
spectf                         &  0.80980 &  0.80900 &   0.81504 &  0.80524 &  0.80536 &  0.80912 &  0.80917 &  0.81197 &  0.80094 &  0.80909 \\
vertebral                      &  0.83677 &  0.83484 &   0.83677 &  0.82194 &  0.81226 &  0.82581 &  0.81806 &  0.83484 &  0.83355 &  0.82387 \\
wilt                           &  0.98281 &  0.98545 &   0.98305 &  0.98367 &  0.98297 &  0.98392 &  0.98343 &  0.98260 &  0.98450 &  0.97644 \\
parkinsons                     &  0.90784 &  0.86084 &   0.91195 &  0.86695 &  0.86274 &  0.87416 &  0.86184 &  0.84837 &  0.86821 &  0.85047 \\
heart                          &  0.81865 &  0.81744 &   0.82062 &  0.81065 &  0.79295 &  0.80733 &  0.79753 &  0.83127 &  0.81665 &  0.79170 \\
wdbc                           &  0.96135 &  0.96064 &   0.96310 &  0.95712 &  0.95504 &  0.95818 &  0.95327 &  0.96170 &  0.95678 &  0.94658 \\
bank                           &  0.99388 &  0.99184 &   0.99359 &  0.99199 &  0.99111 &  0.99184 &  0.98704 &  0.99257 &  0.99228 &  0.98557 \\
ionosphere                     &  0.93281 &  0.93792 &   0.93224 &  0.92883 &  0.92083 &  0.92425 &  0.91968 &  0.93508 &  0.93170 &  0.92537 \\
HRCompetencyScores             &  0.93467 &  0.92533 &   0.93200 &  0.92000 &  0.91400 &  0.91600 &  0.91267 &  0.92267 &  0.92067 &  0.91333 \\
spambase                       &  0.95444 &  0.95331 &   0.95518 &  0.95136 &  0.94940 &  0.95088 &  0.95005 &  0.95249 &  0.95201 &  0.94001 \\
QSAR                           &  0.86864 &  0.86940 &   0.87320 &  0.86466 &  0.86198 &  0.86351 &  0.86426 &  0.86599 &  0.86751 &  0.85271 \\
diabetes                       &  0.76219 &  0.76686 &   0.77001 &  0.76062 &  0.75672 &  0.76040 &  0.75699 &  0.76478 &  0.76166 &  0.75621 \\
breast                         &  0.96135 &  0.96064 &   0.96310 &  0.95712 &  0.95504 &  0.95818 &  0.95327 &  0.96170 &  0.95678 &  0.94658 \\
SPF                            &  0.99433 &  1.00000 &   0.99464 &  1.00000 &  1.00000 &  1.00000 &  1.00000 &  0.99206 &  1.00000 &  1.00000 \\
hillvalley                     &  0.57606 &  0.55002 &   0.57969 &  0.53033 &  0.54668 &  0.53859 &  0.54767 &  0.50827 &  0.54833 &  0.49653 \\
pc4                            &  0.90837 &  0.90906 &   0.90974 &  0.90604 &  0.90069 &  0.90480 &  0.90303 &  0.90645 &  0.91084 &  0.89671 \\
scene                          &  0.91325 &  0.98288 &   0.91591 &  0.98363 &  0.98380 &  0.98397 &  0.98247 &  0.90918 &  0.98255 &  0.97133 \\
Sonar\_Mine\_Rock\_Data           &  0.83257 &  0.81057 &   0.84219 &  0.79910 &  0.78510 &  0.80490 &  0.79648 &  0.77581 &  0.80286 &  0.78467 \\
Customer\_Churn                 &  0.95810 &  0.95689 &   0.95879 &  0.95543 &  0.95517 &  0.95733 &  0.95594 &  0.95721 &  0.95625 &  0.93733 \\
jm1                            &  0.81954 &  0.81426 &   0.82044 &  0.81259 &  0.81268 &  0.81318 &  0.81204 &  0.81362 &  0.81410 &  0.80768 \\
eeg                            &  0.93402 &  0.72413 &   0.93653 &  0.72987 &  0.73202 &  0.73019 &  0.73167 &  0.81760 &  0.72327 &  0.60364 \\
phoneme                        &  0.91225 &  0.90899 &   0.91336 &  0.90537 &  0.90259 &  0.90466 &  0.90307 &  0.90714 &  0.90489 &  0.88239 \\
\bottomrule
Mean                           &  0.88225 &  0.87435 &   0.88401 &  0.87019 &  0.86758 &  0.87027 &  0.86668 &  0.87134 &  0.87288 &  0.85766 \\
Rank                           &  3.56667 &  3.81667 &   2.61667 &  5.98333 &  7.36667 &  5.83333 &  7.46667 &  5.00000 &  4.76667 &  8.58333 \\
\bottomrule
\end{tabular}
}
}

\end{table}

% Real30 Brier
\begin{table}[ht]
\centering
\caption{Brier score performance of calibration methods applied on RF trained on 30 real datasets.}
\label{tab:brier}

\resizebox{0.8\columnwidth}{!}{
\rotatebox{90}{
\begin{tabular}{lrrrrrrrrrr}
\toprule
{} &     RF\_d &   RF\_opt &  RF\_large &    Platt &      ISO &     Beta &       VA &       CT &      PPA &     Rank \\
Data                           &          &          &           &          &          &          &          &          &          &          \\
\midrule
cm1                            &  0.08772 &  0.08415 &   0.08697 &  0.08692 &  0.09169 &  0.08646 &  0.09234 &  0.08583 &  0.08512 &  0.10037 \\
datatrieve                     &  0.08241 &  0.08213 &   0.08199 &  0.08942 &  0.09485 &  0.11338 &  0.10440 &  0.07744 &  0.08447 &  0.08642 \\
kc1\_class\_level\_defectiveornot &  0.17060 &  0.16960 &   0.16966 &  0.17295 &  0.21164 &  0.19406 &  0.18457 &  0.17263 &  0.17405 &  0.21012 \\
kc1                            &  0.10302 &  0.10629 &   0.10221 &  0.10843 &  0.10906 &  0.10741 &  0.10808 &  0.10951 &  0.10681 &  0.11414 \\
kc2                            &  0.11961 &  0.11421 &   0.11950 &  0.11796 &  0.12495 &  0.11895 &  0.12063 &  0.11824 &  0.11766 &  0.12925 \\
kc3                            &  0.07452 &  0.07338 &   0.07315 &  0.07791 &  0.08709 &  0.07895 &  0.08048 &  0.07406 &  0.07661 &  0.08256 \\
pc1                            &  0.04985 &  0.05030 &   0.04961 &  0.05299 &  0.05519 &  0.05345 &  0.05474 &  0.05095 &  0.05146 &  0.06098 \\
spect                          &  0.13076 &  0.12109 &   0.13013 &  0.12743 &  0.13828 &  0.12942 &  0.13313 &  0.12123 &  0.12135 &  0.14518 \\
spectf                         &  0.12580 &  0.12408 &   0.12364 &  0.12917 &  0.14178 &  0.13209 &  0.12980 &  0.12357 &  0.12885 &  0.14079 \\
vertebral                      &  0.10675 &  0.10802 &   0.10499 &  0.11611 &  0.12214 &  0.11690 &  0.12260 &  0.10590 &  0.11095 &  0.14082 \\
wilt                           &  0.01277 &  0.01139 &   0.01261 &  0.01274 &  0.01313 &  0.01251 &  0.01330 &  0.01297 &  0.01193 &  0.02079 \\
parkinsons                     &  0.06961 &  0.09297 &   0.06952 &  0.09419 &  0.10153 &  0.10410 &  0.11116 &  0.09413 &  0.09316 &  0.11486 \\
heart                          &  0.12904 &  0.12965 &   0.12644 &  0.13360 &  0.15038 &  0.13942 &  0.14492 &  0.12226 &  0.13241 &  0.14953 \\
wdbc                           &  0.03121 &  0.03055 &   0.03018 &  0.03165 &  0.03592 &  0.03509 &  0.04009 &  0.03121 &  0.03216 &  0.04198 \\
bank                           &  0.00563 &  0.00587 &   0.00555 &  0.00597 &  0.00704 &  0.00751 &  0.01282 &  0.00669 &  0.00638 &  0.01352 \\
ionosphere                     &  0.05140 &  0.05338 &   0.05037 &  0.05539 &  0.06533 &  0.06246 &  0.07207 &  0.05249 &  0.05443 &  0.06144 \\
HRCompetencyScores             &  0.06117 &  0.06341 &   0.06077 &  0.07139 &  0.07991 &  0.07597 &  0.08721 &  0.06305 &  0.06727 &  0.07687 \\
spambase                       &  0.03835 &  0.03956 &   0.03778 &  0.03691 &  0.03803 &  0.03728 &  0.03798 &  0.03975 &  0.03863 &  0.04779 \\
QSAR                           &  0.09440 &  0.09413 &   0.09356 &  0.09563 &  0.10029 &  0.09627 &  0.09835 &  0.09525 &  0.09519 &  0.11328 \\
diabetes                       &  0.16144 &  0.15750 &   0.16060 &  0.16203 &  0.16957 &  0.16282 &  0.16599 &  0.15868 &  0.16294 &  0.17020 \\
breast                         &  0.03121 &  0.03055 &   0.03018 &  0.03165 &  0.03592 &  0.03509 &  0.04009 &  0.03121 &  0.03216 &  0.04198 \\
SPF                            &  0.01779 &  0.00000 &   0.01735 &  0.00011 &  0.00000 &  0.00000 &  0.00022 &  0.01986 &  0.00000 &  0.00000 \\
hillvalley                     &  0.25145 &  0.24703 &   0.24952 &  0.25059 &  0.25383 &  0.25036 &  0.24974 &  0.25095 &  0.24859 &  0.25234 \\
pc4                            &  0.06180 &  0.06131 &   0.06136 &  0.06474 &  0.06551 &  0.06401 &  0.06426 &  0.06314 &  0.06184 &  0.07942 \\
scene                          &  0.07103 &  0.01612 &   0.07025 &  0.01501 &  0.01523 &  0.01470 &  0.01819 &  0.06953 &  0.01604 &  0.02653 \\
Sonar\_Mine\_Rock\_Data           &  0.12791 &  0.14001 &   0.12677 &  0.13495 &  0.15123 &  0.14146 &  0.14792 &  0.15311 &  0.14133 &  0.15156 \\
Customer\_Churn                 &  0.03185 &  0.03243 &   0.03156 &  0.03284 &  0.03310 &  0.03259 &  0.03314 &  0.03260 &  0.03292 &  0.04949 \\
jm1                            &  0.13393 &  0.13624 &   0.13302 &  0.13742 &  0.13801 &  0.13674 &  0.13768 &  0.13764 &  0.13657 &  0.14552 \\
eeg                            &  0.06812 &  0.19969 &   0.06710 &  0.17866 &  0.17760 &  0.17858 &  0.17756 &  0.13505 &  0.18721 &  0.22880 \\
phoneme                        &  0.06717 &  0.06919 &   0.06646 &  0.07029 &  0.07186 &  0.07027 &  0.07177 &  0.06967 &  0.07081 &  0.08677 \\
\bottomrule
Mean                           &  0.08561 &  0.08814 &   0.08476 &  0.08984 &  0.09600 &  0.09294 &  0.09517 &  0.08929 &  0.08931 &  0.10278 \\
Rank                           &  3.96667 &  3.00000 &   2.46667 &  5.33333 &  8.06667 &  5.86667 &  7.73333 &  4.70000 &  4.76667 &  9.10000 \\
\bottomrule
\end{tabular}
}
}\end{table}

% Real30 log-loss
\begin{table}[ht]
\centering
\caption{Log Loss performance of calibration methods applied on RF trained on 30 real datasets.}
\label{tab:log-loss}

\resizebox{0.8\columnwidth}{!}{
\rotatebox{90}{
\begin{tabular}{lrrrrrrrrrr}
\toprule
{} &     RF\_d &   RF\_opt &  RF\_large &    Platt &      ISO &     Beta &       VA &       CT &      PPA &     Rank \\
Data                           &          &          &           &          &          &          &          &          &          &          \\
\midrule
cm1                            &  0.34404 &  0.28584 &   0.30514 &  0.30657 &  0.84358 &  0.31160 &  0.31824 &  0.35050 &  0.30761 &  1.75981 \\
datatrieve                     &  0.74000 &  0.39064 &   0.31912 &  0.33701 &  1.87272 &  1.23192 &  0.35884 &  0.58065 &  0.36249 &  1.94934 \\
kc1\_class\_level\_defectiveornot &  0.50759 &  0.50239 &   0.49964 &  0.51320 &  3.94946 &  1.07385 &  0.54777 &  0.50684 &  0.51582 &  3.84761 \\
kc1                            &  0.51141 &  0.34364 &   0.44639 &  0.35419 &  0.53144 &  0.34710 &  0.35020 &  0.35548 &  0.34586 &  0.42856 \\
kc2                            &  0.75841 &  0.38836 &   0.69166 &  0.38183 &  1.24538 &  0.42007 &  0.38902 &  0.49684 &  0.42097 &  1.91637 \\
kc3                            &  0.35456 &  0.36265 &   0.30813 &  0.27202 &  1.28291 &  0.34250 &  0.27891 &  0.32928 &  0.41404 &  0.75595 \\
pc1                            &  0.25835 &  0.19982 &   0.19813 &  0.20005 &  0.53042 &  0.21317 &  0.20315 &  0.21231 &  0.22193 &  0.28201 \\
spect                          &  0.44787 &  0.38377 &   0.42046 &  0.40835 &  1.67646 &  0.43211 &  0.42312 &  0.38315 &  0.38867 &  1.68885 \\
spectf                         &  0.38524 &  0.37898 &   0.37992 &  0.39931 &  1.89808 &  0.44280 &  0.40735 &  0.37466 &  0.38905 &  1.87045 \\
vertebral                      &  0.33123 &  0.33112 &   0.32686 &  0.36733 &  1.22963 &  0.39304 &  0.39299 &  0.32641 &  0.40035 &  3.08194 \\
wilt                           &  0.06635 &  0.05200 &   0.05575 &  0.05310 &  0.16454 &  0.05887 &  0.05086 &  0.06129 &  0.05949 &  0.25226 \\
parkinsons                     &  0.23549 &  0.30034 &   0.23476 &  0.31524 &  1.54191 &  0.96759 &  0.37179 &  0.29582 &  0.29293 &  1.66942 \\
heart                          &  0.40464 &  0.40565 &   0.39867 &  0.42222 &  1.92134 &  0.47250 &  0.45449 &  0.38696 &  0.41214 &  1.57580 \\
wdbc                           &  0.18249 &  0.14528 &   0.14683 &  0.12824 &  0.66693 &  0.34707 &  0.16769 &  0.15778 &  0.14830 &  0.49223 \\
bank                           &  0.02711 &  0.02814 &   0.02710 &  0.03282 &  0.13224 &  0.08354 &  0.06757 &  0.03104 &  0.03646 &  0.41225 \\
ionosphere                     &  0.22357 &  0.21389 &   0.18485 &  0.21042 &  1.16134 &  0.45430 &  0.26662 &  0.22933 &  0.19415 &  1.20464 \\
HRCompetencyScores             &  0.34997 &  0.33550 &   0.28702 &  0.26170 &  1.40324 &  0.58260 &  0.31156 &  0.25400 &  0.39266 &  1.04220 \\
spambase                       &  0.18099 &  0.15996 &   0.15204 &  0.13653 &  0.28678 &  0.15111 &  0.14171 &  0.17686 &  0.15101 &  0.23887 \\
QSAR                           &  0.34198 &  0.34713 &   0.31701 &  0.31627 &  0.91980 &  0.33994 &  0.32699 &  0.33731 &  0.33151 &  0.62808 \\
diabetes                       &  0.49559 &  0.47766 &   0.48513 &  0.49378 &  1.08301 &  0.49996 &  0.50304 &  0.48143 &  0.49279 &  0.88515 \\
breast                         &  0.18249 &  0.14528 &   0.14683 &  0.12824 &  0.66693 &  0.34707 &  0.16769 &  0.15778 &  0.14830 &  0.49223 \\
SPF                            &  0.10368 &  0.00001 &   0.10370 &  0.01008 &  0.00000 &  0.00000 &  0.01047 &  0.11040 &  0.00000 &  0.00000 \\
hillvalley                     &  0.70298 &  0.68760 &   0.69778 &  0.69477 &  0.93638 &  0.69434 &  0.69355 &  0.69519 &  0.69095 &  0.73128 \\
pc4                            &  0.19637 &  0.23117 &   0.19538 &  0.21326 &  0.46890 &  0.23660 &  0.20763 &  0.19902 &  0.23972 &  0.41385 \\
scene                          &  0.25442 &  0.09200 &   0.25024 &  0.07206 &  0.18049 &  0.09835 &  0.08688 &  0.24034 &  0.09196 &  0.17579 \\
Sonar\_Mine\_Rock\_Data           &  0.41125 &  0.44403 &   0.40972 &  0.41653 &  2.37837 &  0.82419 &  0.46170 &  0.47003 &  0.43811 &  1.40972 \\
Customer\_Churn                 &  0.13440 &  0.14098 &   0.11849 &  0.11923 &  0.30259 &  0.13393 &  0.12005 &  0.13278 &  0.13264 &  0.27842 \\
jm1                            &  0.54955 &  0.43127 &   0.49348 &  0.43591 &  0.48262 &  0.43296 &  0.43538 &  0.43826 &  0.43241 &  0.47278 \\
eeg                            &  0.25426 &  0.58559 &   0.25307 &  0.52957 &  0.55855 &  0.52912 &  0.52620 &  0.42866 &  0.55458 &  0.65237 \\
phoneme                        &  0.25496 &  0.26020 &   0.23672 &  0.23749 &  0.35514 &  0.25779 &  0.24009 &  0.26491 &  0.26671 &  0.32920 \\
\bottomrule
Mean                           &  0.33971 &  0.30170 &   0.30300 &  0.29224 &  1.02237 &  0.42400 &  0.30939 &  0.31551 &  0.30912 &  1.03125 \\
Rank                           &  5.96667 &  3.80000 &   3.33333 &  3.46667 &  9.23333 &  6.13333 &  4.73333 &  4.76667 &  4.73333 &  8.83333 \\
\bottomrule
\end{tabular}
}
}
\end{table}

% Real30 ECE
\begin{table}[ht]
\centering
\caption{ECE performance of calibration methods applied on RF trained on 30 real datasets.}
\label{tab:ece}

\resizebox{0.8\columnwidth}{!}{
\rotatebox{90}{
\begin{tabular}{lrrrrrrrrrr}
\toprule
{} &     RF\_d &   RF\_opt &  RF\_large &    Platt &      ISO &     Beta &       VA &       CT &      PPA &     Rank \\
Data                           &          &          &           &          &          &          &          &          &          &          \\
\midrule
cm1                            &  0.08585 &  0.06327 &   0.09443 &  0.06886 &  0.06949 &  0.05918 &  0.06823 &  0.07002 &  0.08052 &  0.15296 \\
datatrieve                     &  0.09778 &  0.09100 &   0.10087 &  0.06250 &  0.06711 &  0.09234 &  0.08902 &  0.07833 &  0.09238 &  0.05357 \\
kc1\_class\_level\_defectiveornot &  0.15259 &  0.14115 &   0.15703 &  0.14451 &  0.12648 &  0.14699 &  0.11087 &  0.14771 &  0.13055 &  0.13007 \\
kc1                            &  0.06695 &  0.05572 &   0.06734 &  0.08428 &  0.07030 &  0.06423 &  0.05078 &  0.05734 &  0.06237 &  0.04853 \\
kc2                            &  0.11614 &  0.11684 &   0.12629 &  0.12415 &  0.11194 &  0.11977 &  0.07200 &  0.13700 &  0.11380 &  0.10309 \\
kc3                            &  0.11268 &  0.09968 &   0.10412 &  0.11730 &  0.09674 &  0.10420 &  0.06992 &  0.11748 &  0.11349 &  0.05434 \\
pc1                            &  0.10781 &  0.10413 &   0.10973 &  0.13300 &  0.08879 &  0.11414 &  0.06987 &  0.10612 &  0.10076 &  0.03035 \\
spect                          &  0.12102 &  0.09684 &   0.11683 &  0.12539 &  0.09290 &  0.12300 &  0.08344 &  0.10285 &  0.10323 &  0.08558 \\
spectf                         &  0.10431 &  0.11872 &   0.10414 &  0.13851 &  0.12114 &  0.15343 &  0.07969 &  0.11685 &  0.11656 &  0.13038 \\
vertebral                      &  0.11566 &  0.11758 &   0.11352 &  0.13280 &  0.10383 &  0.14002 &  0.08328 &  0.11219 &  0.12282 &  0.13262 \\
wilt                           &  0.08824 &  0.08965 &   0.08920 &  0.15046 &  0.07986 &  0.12104 &  0.06311 &  0.08926 &  0.09367 &  0.15664 \\
parkinsons                     &  0.08764 &  0.11169 &   0.09584 &  0.12012 &  0.11238 &  0.12357 &  0.08278 &  0.10889 &  0.10681 &  0.09765 \\
heart                          &  0.09520 &  0.09564 &   0.09333 &  0.11881 &  0.11324 &  0.13231 &  0.08548 &  0.08973 &  0.09692 &  0.08027 \\
wdbc                           &  0.08561 &  0.08189 &   0.07581 &  0.10339 &  0.09845 &  0.11989 &  0.06328 &  0.08321 &  0.09043 &  0.07964 \\
bank                           &  0.06301 &  0.05730 &   0.05993 &  0.08850 &  0.08251 &  0.04631 &  0.04812 &  0.06321 &  0.06612 &  0.06049 \\
ionosphere                     &  0.08202 &  0.07852 &   0.08319 &  0.09363 &  0.08559 &  0.10942 &  0.06379 &  0.07851 &  0.08339 &  0.09023 \\
HRCompetencyScores             &  0.07438 &  0.07719 &   0.07401 &  0.09237 &  0.06913 &  0.09582 &  0.07555 &  0.08684 &  0.08835 &  0.08160 \\
spambase                       &  0.03635 &  0.03435 &   0.03658 &  0.07979 &  0.05692 &  0.06818 &  0.05067 &  0.03498 &  0.03379 &  0.09354 \\
QSAR                           &  0.06413 &  0.06933 &   0.06210 &  0.09233 &  0.07090 &  0.08957 &  0.05746 &  0.06939 &  0.06009 &  0.08850 \\
diabetes                       &  0.07172 &  0.07209 &   0.06280 &  0.07207 &  0.06819 &  0.06827 &  0.06001 &  0.07222 &  0.07095 &  0.07677 \\
breast                         &  0.08561 &  0.08189 &   0.07581 &  0.10339 &  0.09845 &  0.11989 &  0.06328 &  0.08321 &  0.09043 &  0.07964 \\
SPF                            &  0.06359 &  0.00000 &   0.06234 &  0.00013 &  0.00000 &  0.00000 &  0.00294 &  0.06185 &  0.00000 &  0.00000 \\
hillvalley                     &  0.08172 &  0.06646 &   0.08302 &  0.05593 &  0.10724 &  0.05924 &  0.07681 &  0.04198 &  0.06745 &  0.07332 \\
pc4                            &  0.07553 &  0.08033 &   0.07833 &  0.12328 &  0.08939 &  0.10386 &  0.06528 &  0.07549 &  0.07927 &  0.11347 \\
scene                          &  0.06871 &  0.05690 &   0.07451 &  0.07637 &  0.06376 &  0.08142 &  0.05244 &  0.07042 &  0.06497 &  0.17361 \\
Sonar\_Mine\_Rock\_Data           &  0.08872 &  0.09252 &   0.08961 &  0.12026 &  0.08903 &  0.14568 &  0.08735 &  0.10250 &  0.08317 &  0.07194 \\
Customer\_Churn                 &  0.06237 &  0.06810 &   0.06468 &  0.12165 &  0.07005 &  0.09601 &  0.05371 &  0.06329 &  0.05981 &  0.15129 \\
jm1                            &  0.03254 &  0.03025 &   0.03164 &  0.03244 &  0.03798 &  0.03094 &  0.03879 &  0.01399 &  0.03297 &  0.03940 \\
eeg                            &  0.02636 &  0.03501 &   0.02744 &  0.00481 &  0.02276 &  0.00493 &  0.01717 &  0.01309 &  0.01617 &  0.02797 \\
phoneme                        &  0.01688 &  0.01754 &   0.01729 &  0.03089 &  0.04140 &  0.02466 &  0.03874 &  0.01664 &  0.01994 &  0.06617 \\
\bottomrule
Mean                           &  0.08104 &  0.07672 &   0.08106 &  0.09373 &  0.08020 &  0.09194 &  0.06413 &  0.07882 &  0.07804 &  0.08745 \\
Rank                           &  5.50000 &  4.86667 &   5.43333 &  7.76667 &  5.53333 &  7.10000 &  2.73333 &  5.10000 &  5.26667 &  5.70000 \\
\bottomrule
\end{tabular}
}
}
\end{table}

\clearpage

% 100 trees run name 1725451866_paper10CV100tree

% Real30 CD ACC

% % Real30 CD BS
% \begin{figure*}[ht]
% \begin{center}

%   \includegraphics[scale=0.55]{images/CD_brier100.pdf}

% \caption{Critical difference diagram of 30 real datasets on Brier score.}
% \label{fig:cd_brier}
% \end{center}
% \end{figure*}

% Real30 CD log-loss

% % Real30 CD ECE
% \begin{figure*}[ht]
% \begin{center}

%   \includegraphics[scale=0.55]{images/CD_ece100.pdf}

% \caption{Critical difference diagram of 30 real datasets on ECE.}
% \label{fig:cd_ece}
% \end{center}
% \end{figure*}

\subsubsection{Laplace Correction}
\label{apx:exp_lap}
% RF_d 100 trees run names ["1725451866_paper10CV100tree", "1725635197_paper10CV100treeL1"]

\begin{table}
\centering
\caption{Effectiveness of Laplace correction on calibration performance on 30 real datasets.}

\begin{tabular}{lllll}
\toprule
{} &        Accuracy &      Brier score &    log-loss &        ECE \\
Calibration methods &            &            &            &            \\
\midrule
RF\_d          &            &  \ding{55} &  \ding{51} &  \ding{51} \\
RF\_opt        &            &  \ding{55} &  \ding{51} &  \ding{51} \\
RF\_large      &  \ding{55} &  \ding{55} &            &  \ding{51} \\
Platt         &            &            &            &  \ding{51} \\
ISO           &            &            &            &            \\
Beta          &  \ding{55} &            &  \ding{51} &            \\
VA            &            &  \ding{55} &  \ding{55} &            \\
CT            &  \ding{55} &  \ding{55} &  \ding{51} &  \ding{51} \\
PPA           &            &  \ding{55} &  \ding{51} &  \ding{51} \\
Rank          &            &  \ding{51} &            &            \\
\bottomrule
\end{tabular}
\label{tab:lap}
\end{table}

In this section we evaluate the impact of Laplace correction on all calibration methods across various evaluation metrics. We conducted pairwise T-tests to compare calibration methods with and without Laplace correction, with a significance level (alpha) set to 0.05. Each calibration method was trained twice on the 30 real datasets with the same settings as the previous experiment, once with and once without Laplace correction applied to the final RF prediction. Results in Table \ref{tab:lap} indicate a positive impact of Laplace correction (as confirmed by the pairwise T-test) with a checkmark and a detrimental effect with a cross.

Upon examining the outcomes, we observe that the results are in agreement with the findings of \citet{bostrom2007estimating}, who reported a negative impact of Laplace correction on both accuracy and Brier score. When evaluating calibration performance using log-loss, the inclusion of the Laplace correction generally yields positive results. Consequently, the degree of performance enhancement is contingent upon the selected evaluation metric and calibration approach.

\subsubsection{The Benefit of Out of Bag Data}
\label{apx:exp_oob}

% % 50 trees. 5 run CV 10
% Out-of-bag (OOB) data refers to a specific concept related to the training process and evaluation of individual decision trees within the Random Forest ensemble. As we know, Random Forest combines multiple decision trees to make predictions. Each tree of the RF is trained on a bootstrap sample from the original training dataset, which involves randomly selecting data points with replacements to create a new subset for each tree's training. The key point is that not all data points are included in the training set of each individual decision tree. When constructing each tree, about one-third of the original data points are left out due to the random selection with replacement. These left-out data points constitute the out-of-bag data for that particular tree. Another way to look at this is to say that for each data point in the original training data, there exists a subset of the trees in the Random Forest that are not trained on that specific data point.

Each data point in the original training data is only used by a subset of the trees in the forest\,---\,for the remaining trees, it is ``out of bag'' (OOB), i.e., not contained in the bootstrap sample. Therefore, this latter subset can be used to produce unbiased predictions for the data point, and hence can be used for calibration. This effectively eliminates the need to set aside a portion of the training data as a calibration. The OOB predictions for any given sample from the training set represent the average probability distribution derived from the subset of trees that were not trained on that specific sample. 

The primary concern with OOB predictions is that only about one-third of the trees in the RF contribute to each individual OOB prediction. However, since a different subset of trees is used for each data point in the training dataset, the entire forest is indirectly utilized when considering all OOB predictions across the training dataset.

The results of our experiments on the 30 real datasets is similar to the findings by \citet{johansson2019efficient} with a more limited set of calibration methods.  An investigation of the comparison between using a calibration set and OOB data in terms of statistical difference in the pairwise T-test with a significance level of 0.05, shown in Table \ref{tab:oob}, also confirms the beneficial effect of OOB data for almost all calibration methods, except for Rank calibration. Keep in mind that no calibration is performed on RF\_d, RF\_opt, and RF\_large. Therefore, in training them, there is no need to set aside a calibration set, and therefore no need for OOB data.

% 100 tress run names ["1725451866_paper10CV100tree", "1725746609_paper10CV100treeOOB"]

\begin{table}
\centering
\caption{Effectiveness of OOB in comparison with separate calibration set on 30 real datasets}

\begin{tabular}{lllll}
\toprule
{} &        Accuracy &      Brier score &    log-loss &        ECE \\
Calibration methods &            &            &            &            \\
\midrule
Platt         &  \ding{51} &  \ding{51} &  \ding{51} &  \ding{55} \\
ISO           &  \ding{51} &  \ding{51} &  \ding{51} &  \ding{51} \\
Beta          &  \ding{51} &  \ding{51} &  \ding{51} &  \ding{51} \\
VA            &  \ding{51} &  \ding{51} &  \ding{51} &            \\
PPA           &  \ding{51} &  \ding{51} &  \ding{51} &            \\
Rank          &  \ding{51} &  \ding{55} &  \ding{55} &  \ding{55} \\
\bottomrule
\end{tabular}
\label{tab:oob}
\end{table}

\subsubsection{Random Forest Compared to Other ML algorithms}
\label{apx:exp_ml}
% run name 1726604185_Model_comp
% old exp run name 1698251557_Paper_v3_Real30_MLcomp

So far, we have examined the calibration performance of RF under various conditions using both synthetic and real datasets. We have also compared all the introduced post-calibration methods suitable for RF. In this final part of the experiments, we aim to compare the calibration performance of RF with other machine learning methods. This experiment seeks to determine whether RF can compete with or even outdo other machine learning models in terms of calibration performance.

In this experiment, we use the baseline RF\_d, RF\_opt and RF\_large to represent different calibrated versions of RF and we compare with two of the state of the art ensemble models that is Deep Neural Network ensembles (DNN\_ens) of size 10 and XGBoost forest (XGB) with 100 trees. Additionally, we trained Decision Trees (DT), Logistic Regression (LR), SVM, Deep Neural Network (DNN), and Gaussian Naive Bayes (GNB) models. Similar to the previous experiment, we used the 30 real datasets introduced in Table \ref{tab:data} and performed a 10-fold stratified cross-validation five times with different random seeds, to report the average results. To ensure a fair comparison, all machine learning models, except RF\_large and DNN\_ens (due to high runtime), underwent hyper-parameter optimization. We conducted a randomized grid search with 50 iterations for each learner to maintain consistency in the comparison. Tables \ref{tab:DT}, \ref{tab:SVM}, \ref{tab:LR}, \ref{tab:DNN}, \ref{tab:XGB} and \ref{tab:GNB} show the details of hyper-parameter search space for the DT, SVM, LR, DNN, XGB, and GNB, respectively.

\begin{table}[!ht]
    \centering
    \caption{Decision Tree Search Space}
    \begin{tabular}{|l|l|}
        \hline
        \textbf{hyper-parameter} & \textbf{Values} \\
        \hline
        Criterion & [gini, entropy, log\_loss] \\
        \hline
        Splitter & [best, random] \\
        \hline
        Max\_depth & [2, 3, ..., 100] \\
        \hline
        Min\_samples\_split & [2, 3, ..., 10] \\
        \hline
        Min\_samples\_leaf & [1, 2, ..., 10] \\
        \hline
        Max\_features & [sqrt, log2, None] \\
        \hline
    \end{tabular}
    \label{tab:DT}
\end{table}

\begin{table}[!ht]
    \centering
    \caption{SVM Search Space}
    \begin{tabular}{|l|l|}
        \hline
        \textbf{hyper-parameter} & \textbf{Values} \\
        \hline
        kernel & [linear, poly, rbf, sigmoid] \\
        \hline
        C & [0.1, 1, 10, 100] \\
        \hline
        degree & [2, 3, 4] \\
        \hline
        gamma & [scale, auto, 0.1, 1, 10] \\
        \hline
        coef0 & [0, 1, 2] \\
        \hline
        shrinking & [True, False] \\
        \hline
        class\_weight & [None, balanced] \\
        \hline
        max\_iter & [1000, 5000, 10000] \\
        \hline
        decision\_function\_shape & [ovo, ovr] \\
        \hline
        tol & [1e-4, 1e-3, 1e-2] \\
        \hline
        probability & True \\
        \hline
    \end{tabular}
    \label{tab:SVM}
\end{table}

\begin{table}[!ht]
    \centering
    \caption{Logistic Regression Search Space}
    \begin{tabular}{|l|l|}
        \hline
        \textbf{hyper-parameter} & \textbf{Values} \\
        \hline
        Penalty & [l2, None] \\
        \hline
        C & [0.001, 1, 10, 100] \\
        \hline
        Solver & [newton-cholesky, newton-cg, lbfgs, sag, saga] \\
        \hline
        Max\_iter & [100, 500, 1000] \\
        \hline
        Intercept\_scaling & [0.1, 1, 10] \\
        \hline
    \end{tabular}
    \label{tab:LR}
\end{table}

\begin{table}[!ht]
    \centering
    \caption{Deep Neural Network Search Space}
    \begin{tabular}{|l|l|}
        \hline
        \textbf{hyper-parameter} & \textbf{Values} \\
        \hline
        Hidden\_layer\_sizes & [(50, 25), (100, 50), (100, 50, 25), (100, 100, 50), (100, 100, 100, 50)] \\
        \hline
        Activation & [relu, tanh] \\
        \hline
        Solver & [adam, sgd] \\
        \hline
        Alpha & [0.0001, 0.001, 0.01] \\
        \hline
        Learning\_rate & [constant, invscaling, adaptive] \\
        \hline
        Max\_iter & [200, 300, 500] \\
        \hline
        early\_stopping & [False, True] \\
        \hline
    \end{tabular}
    \label{tab:DNN}
\end{table}

\begin{table}[!ht]
    \centering
    \caption{XGBoost Search Space}
    \begin{tabular}{|l|l|}
        \hline
        \textbf{hyper-parameter} & \textbf{Values} \\
        \hline
        n\_estimators & 100 \\
        \hline
        max\_depth & [2, 3, ..., 100] \\
        \hline
        learning\_rate & [0.01, 0.05, 0.1, 0.2] \\
        \hline
        subsample & [0.6, 0.7, 0.8, 0.9, 1.0] \\
        \hline
        colsample\_bytree & [0.6, 0.7, 0.8, 0.9, 1.0] \\
        \hline
        gamma & [0, 0.1, 0.2, 0.3, 0.4] \\
        \hline
        min\_child\_weight & [1, 2, 3, 4, 5] \\
        \hline
    \end{tabular}
    \label{tab:XGB}
\end{table}

\begin{table}[!ht]
    \centering
    \caption{Gaussian Naive Bayes Search Space}
    \begin{tabular}{|l|l|}
        \hline
        \textbf{hyper-parameter} & \textbf{Values} \\
        \hline
        Var\_smoothing & [1e-9, 1e-8, 1e-7, 1e-6, 1e-5, 1e-4, 1e-3, 1e-2, 1e-1, 1.0] \\
        \hline
    \end{tabular}
    \label{tab:GNB}
\end{table}

\begin{figure*}[ht]
\begin{center}

  \includegraphics[scale=0.55]{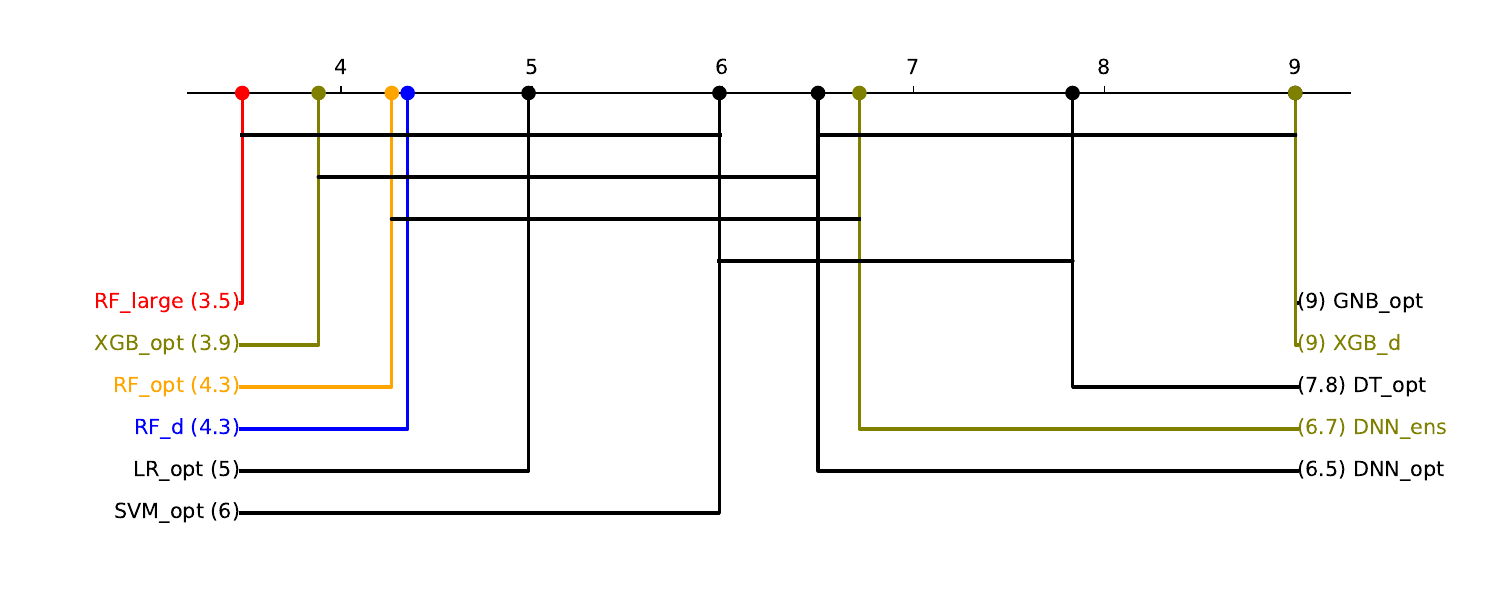}

\caption{Critical difference diagram of the accuracy of 30 real datasets on ML learners.}
\label{fig:cd_acc_ml}
\end{center}
\end{figure*}

\begin{figure*}[ht]
\begin{center}

  \includegraphics[scale=0.55]{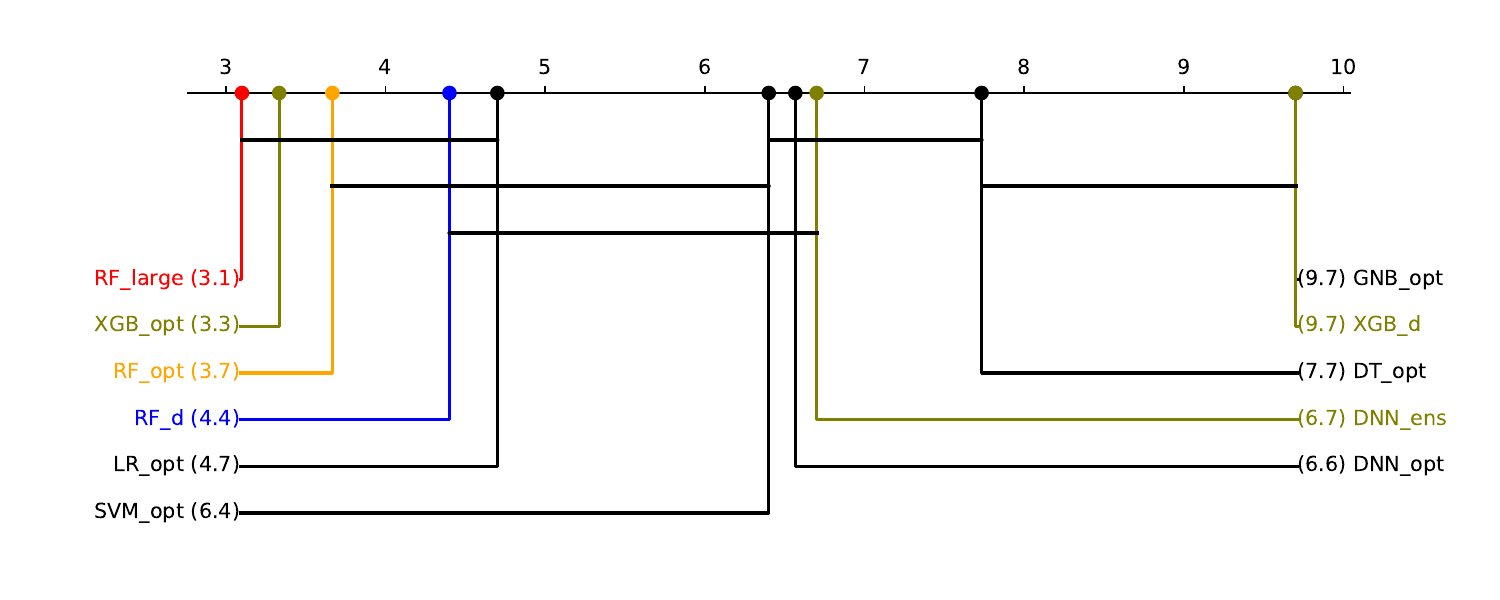}

\caption{Critical difference diagram of the Brier score of 30 real datasets on ML learners.}
\label{fig:cd_brier_ml}
\end{center}
\end{figure*}

\begin{figure*}[ht]
\begin{center}

  \includegraphics[scale=0.55]{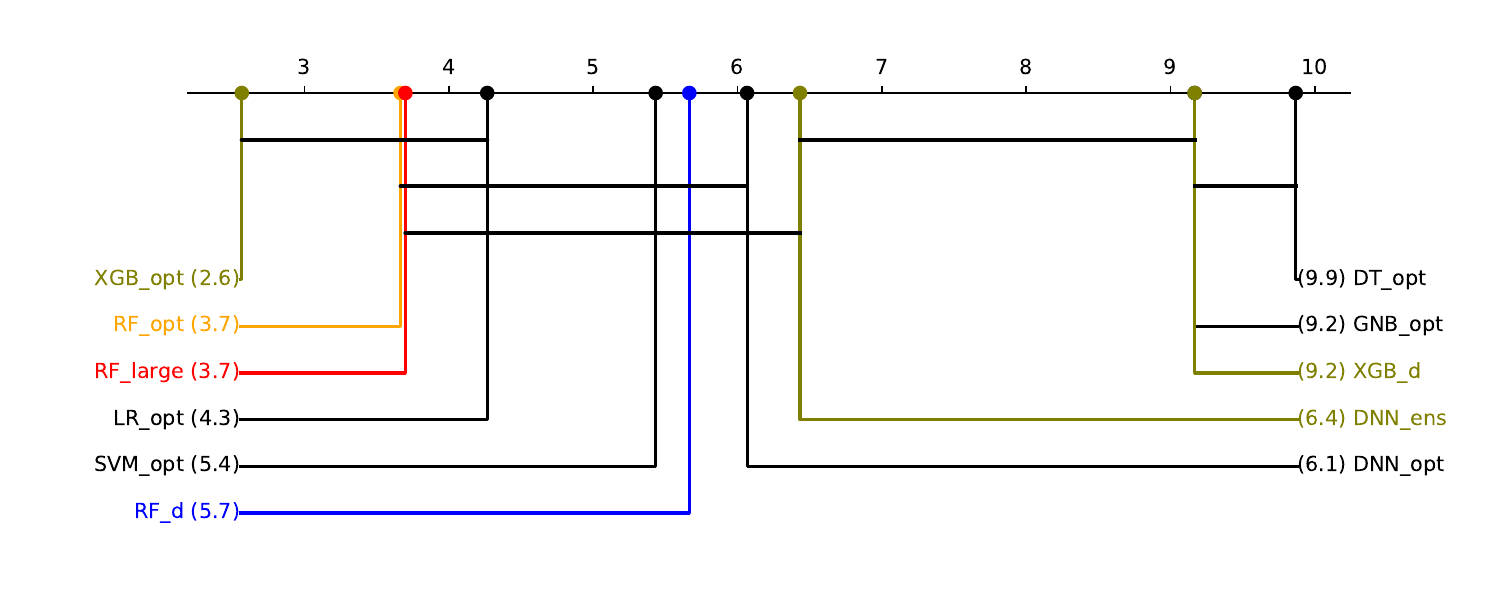}

\caption{Critical difference diagram of the log-loss of 30 real datasets on ML learners.}
\label{fig:cd_log_ml}
\end{center}
\end{figure*}

\begin{figure*}[ht]
\begin{center}

  \includegraphics[scale=0.55]{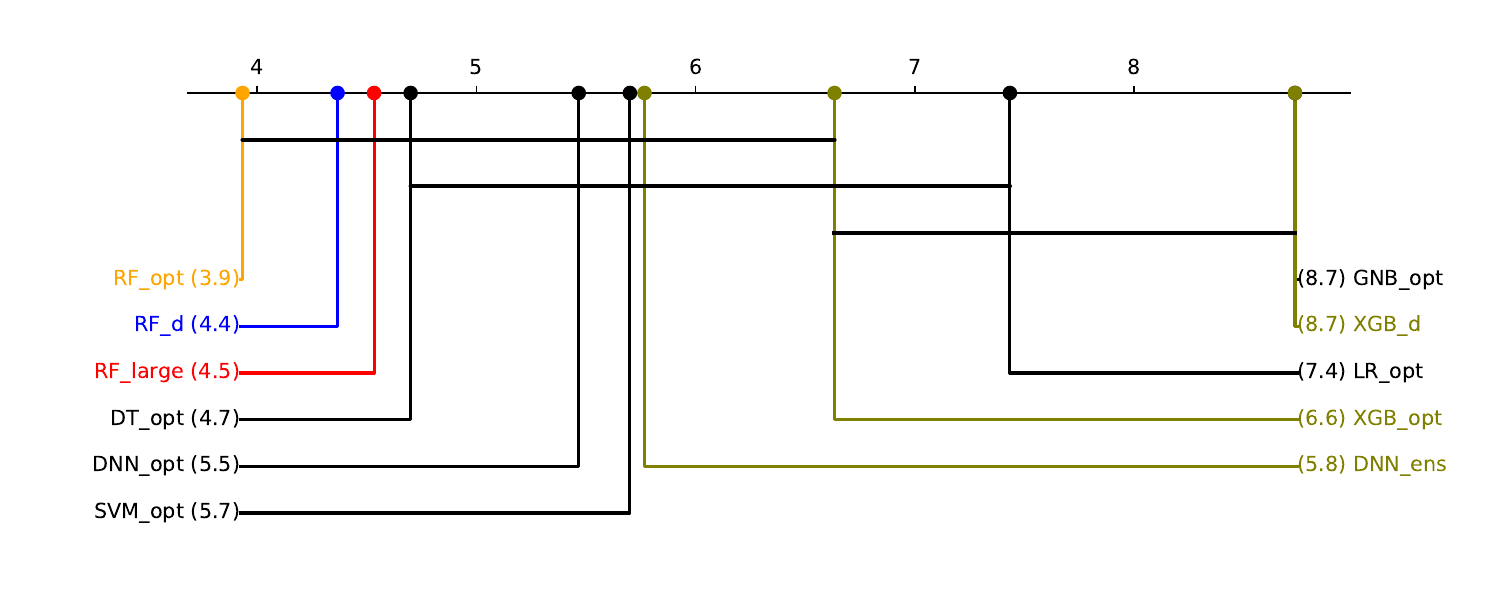}

\caption{Critical difference diagram of the ECE of 30 real datasets on ML learners.}
\label{fig:cd_ece_ml}
\end{center}
\end{figure*}

Tables \ref{tab:acc_m}, \ref{tab:brier_m}, \ref{tab:log-loss_m}, and \ref{tab:ece_m} present the results of the model comparisons based on accuracy, Brier score, log-loss, and ECE, respectively. Correspondingly, Figures \ref{fig:cd_acc_ml}, \ref{fig:cd_brier_ml}, \ref{fig:cd_log_ml}, and \ref{fig:cd_ece_ml} display critical difference diagrams from the Nemenyi-Friedman test for these metrics. Conducted at a 0.05 significance level, the test highlights statistically significant differences among the machine learning models across 30 real datasets, with the null hypothesis assuming no statistical difference between any two models.

Across all metrics, the RF model consistently ranks among the top and best-performing groups of learners. Interestingly, based on the ECE metric, the DNN, which is known for overestimating its predictive probability distributions, ranks in the first and best-calibrated group.

Lastly, we present the average run-time of each machine learning model across the 30 datasets in Table \ref{tab:time}. While RF may not be the fastest, it ranks as the second fastest when considering RF\_large. Moreover, it delivers top-tier calibration performance in significantly less time compared to logistic regression.

% Real30 ACC
\begin{table}[ht]
\centering
\caption{The accuracy of RF trained on 30 real datasets in comparison with other ensemble and base learners.}

\label{tab:acc_m}
\resizebox{0.8\columnwidth}{!}{

\rotatebox{90}{
\begin{tabular}{lrrrrrrrrrrr}
\toprule
{} &     RF\_d &   RF\_opt &  RF\_large &  DNN\_ens &    XGB\_d &  XGB\_opt &   DT\_opt &   LR\_opt &  SVM\_opt &  DNN\_opt &  GNB\_opt \\
Data                           &          &          &           &          &          &          &          &          &          &          &          \\
\midrule
cm1                            &  0.89400 &  0.90123 &   0.89401 &  0.90123 &  0.89842 &  0.89962 &  0.89963 &  0.89162 &  0.90001 &  0.89922 &  0.89842 \\
datatrieve                     &  0.90154 &  0.89846 &   0.90462 &  0.90923 &  0.90000 &  0.90308 &  0.89692 &  0.88462 &  0.91077 &  0.91231 &  0.90000 \\
kc1\_class\_level\_defectiveornot &  0.74486 &  0.73610 &   0.74210 &  0.72838 &  0.63410 &  0.68752 &  0.69705 &  0.70343 &  0.72867 &  0.69552 &  0.63410 \\
kc1                            &  0.86003 &  0.85633 &   0.86325 &  0.84230 &  0.85074 &  0.85207 &  0.84808 &  0.85871 &  0.83282 &  0.84637 &  0.85074 \\
kc2                            &  0.83401 &  0.83668 &   0.83513 &  0.78845 &  0.82303 &  0.84098 &  0.82298 &  0.83906 &  0.81914 &  0.79613 &  0.82303 \\
kc3                            &  0.89782 &  0.89824 &   0.89651 &  0.89176 &  0.90174 &  0.89959 &  0.90389 &  0.90173 &  0.90306 &  0.90614 &  0.90174 \\
pc1                            &  0.93815 &  0.93545 &   0.93851 &  0.93003 &  0.92986 &  0.93274 &  0.92895 &  0.93039 &  0.93184 &  0.93418 &  0.92986 \\
spect                          &  0.82339 &  0.83761 &   0.82493 &  0.82057 &  0.71826 &  0.83299 &  0.80245 &  0.83772 &  0.82795 &  0.82937 &  0.71826 \\
spectf                         &  0.80980 &  0.80900 &   0.81504 &  0.79182 &  0.68242 &  0.81513 &  0.76789 &  0.82479 &  0.81553 &  0.79205 &  0.68242 \\
vertebral                      &  0.83677 &  0.83484 &   0.83677 &  0.80581 &  0.75548 &  0.83226 &  0.80387 &  0.85032 &  0.85032 &  0.80452 &  0.75548 \\
wilt                           &  0.98281 &  0.98545 &   0.98305 &  0.97913 &  0.94606 &  0.98562 &  0.98186 &  0.96421 &  0.98392 &  0.97471 &  0.94606 \\
parkinsons                     &  0.90784 &  0.86084 &   0.91195 &  0.80205 &  0.77621 &  0.89568 &  0.83589 &  0.85732 &  0.80126 &  0.79095 &  0.77621 \\
heart                          &  0.81865 &  0.81744 &   0.82062 &  0.80662 &  0.80533 &  0.83325 &  0.77178 &  0.83112 &  0.79673 &  0.79551 &  0.80533 \\
wdbc                           &  0.96135 &  0.96064 &   0.96310 &  0.92478 &  0.93919 &  0.96766 &  0.93496 &  0.96416 &  0.93710 &  0.92549 &  0.93919 \\
bank                           &  0.99388 &  0.99184 &   0.99359 &  1.00000 &  0.84082 &  0.99679 &  0.98498 &  0.98834 &  1.00000 &  1.00000 &  0.84082 \\
ionosphere                     &  0.93281 &  0.93792 &   0.93224 &  0.94413 &  0.89684 &  0.93222 &  0.87867 &  0.87116 &  0.94070 &  0.93962 &  0.89684 \\
HRCompetencyScores             &  0.93467 &  0.92533 &   0.93200 &  0.92067 &  0.91333 &  0.92800 &  0.89867 &  0.92467 &  0.92400 &  0.93000 &  0.91333 \\
spambase                       &  0.95444 &  0.95331 &   0.95518 &  0.93367 &  0.85177 &  0.95305 &  0.91754 &  0.92797 &  0.84521 &  0.93627 &  0.85177 \\
QSAR                           &  0.86864 &  0.86940 &   0.87320 &  0.87679 &  0.69192 &  0.87037 &  0.83356 &  0.86923 &  0.86126 &  0.85802 &  0.69192 \\
diabetes                       &  0.76219 &  0.76686 &   0.77001 &  0.71349 &  0.75154 &  0.75283 &  0.73537 &  0.77235 &  0.75780 &  0.69737 &  0.75154 \\
breast                         &  0.96135 &  0.96064 &   0.96310 &  0.92478 &  0.93919 &  0.96766 &  0.93496 &  0.96416 &  0.93710 &  0.92549 &  0.93919 \\
SPF                            &  0.99433 &  1.00000 &   0.99464 &  0.63832 &  0.64874 &  1.00000 &  0.99866 &  0.99897 &  0.65048 &  0.65204 &  0.64874 \\
hillvalley                     &  0.57606 &  0.55002 &   0.57969 &  0.75938 &  0.50824 &  0.54852 &  0.51933 &  0.95711 &  0.71321 &  0.62281 &  0.50824 \\
pc4                            &  0.90837 &  0.90906 &   0.90974 &  0.87188 &  0.87257 &  0.90796 &  0.88655 &  0.91359 &  0.87765 &  0.87668 &  0.87257 \\
scene                          &  0.91325 &  0.98288 &   0.91591 &  0.98438 &  0.87229 &  0.98372 &  0.97915 &  0.98646 &  0.98912 &  0.98363 &  0.87229 \\
Sonar\_Mine\_Rock\_Data           &  0.83257 &  0.81057 &   0.84219 &  0.83705 &  0.69029 &  0.85000 &  0.70500 &  0.77624 &  0.63743 &  0.72695 &  0.69029 \\
Customer\_Churn                 &  0.95810 &  0.95689 &   0.95879 &  0.86190 &  0.84286 &  0.96089 &  0.93784 &  0.89156 &  0.85613 &  0.88933 &  0.84286 \\
jm1                            &  0.81954 &  0.81426 &   0.82044 &  0.80259 &  0.80814 &  0.81311 &  0.80886 &  0.81327 &  0.79767 &  0.80915 &  0.80814 \\
eeg                            &  0.93402 &  0.72413 &   0.93653 &  0.56154 &  0.44893 &  0.82077 &  0.55158 &  0.55120 &  0.55302 &  0.55113 &  0.44893 \\
phoneme                        &  0.91225 &  0.90899 &   0.91336 &  0.89082 &  0.75518 &  0.90652 &  0.85300 &  0.75044 &  0.86684 &  0.90163 &  0.75518 \\
\bottomrule
Mean                           &  0.88225 &  0.87435 &   0.88401 &  0.84812 &  0.79645 &  0.87902 &  0.84400 &  0.86986 &  0.84156 &  0.84009 &  0.79645 \\
Rank                           &  4.35000 &  4.26667 &   3.48333 &  6.71667 &  9.00000 &  3.88333 &  7.83333 &  4.98333 &  5.98333 &  6.50000 &  9.00000 \\
\bottomrule
\end{tabular}
}
}

\end{table}

% Real30 Brier
\begin{table}[ht]
\centering
\caption{The Brier score of RF trained on 30 real datasets in comparison with other ensemble and base learners}
\label{tab:brier_m}

\resizebox{0.8\columnwidth}{!}{
\rotatebox{90}{
\begin{tabular}{lrrrrrrrrrrr}
\toprule
{} &     RF\_d &   RF\_opt &  RF\_large &  DNN\_ens &    XGB\_d &  XGB\_opt &   DT\_opt &   LR\_opt &  SVM\_opt &  DNN\_opt &  GNB\_opt \\
Data                           &          &          &           &          &          &          &          &          &          &          &          \\
\midrule
cm1                            &  0.08772 &  0.08415 &   0.08697 &  0.09875 &  0.09190 &  0.08586 &  0.08791 &  0.08368 &  0.08956 &  0.08895 &  0.09190 \\
datatrieve                     &  0.08241 &  0.08213 &   0.08199 &  0.08268 &  0.08806 &  0.07986 &  0.08973 &  0.08998 &  0.08236 &  0.08058 &  0.08806 \\
kc1\_class\_level\_defectiveornot &  0.17060 &  0.16960 &   0.16966 &  0.18738 &  0.23110 &  0.18605 &  0.19939 &  0.20242 &  0.18454 &  0.19430 &  0.23110 \\
kc1                            &  0.10302 &  0.10629 &   0.10221 &  0.11524 &  0.12576 &  0.10857 &  0.11330 &  0.10649 &  0.12654 &  0.11453 &  0.12576 \\
kc2                            &  0.11961 &  0.11421 &   0.11950 &  0.14438 &  0.16696 &  0.11851 &  0.12901 &  0.11753 &  0.13337 &  0.13158 &  0.16696 \\
kc3                            &  0.07452 &  0.07338 &   0.07315 &  0.08876 &  0.08625 &  0.07460 &  0.08319 &  0.07797 &  0.08171 &  0.07945 &  0.08625 \\
pc1                            &  0.04985 &  0.05030 &   0.04961 &  0.07075 &  0.06555 &  0.05371 &  0.06230 &  0.05879 &  0.06260 &  0.06050 &  0.06555 \\
spect                          &  0.13076 &  0.12109 &   0.13013 &  0.13265 &  0.24668 &  0.12488 &  0.14288 &  0.11970 &  0.12204 &  0.12147 &  0.24668 \\
spectf                         &  0.12580 &  0.12408 &   0.12364 &  0.13224 &  0.29950 &  0.12612 &  0.16120 &  0.11910 &  0.12296 &  0.14320 &  0.29950 \\
vertebral                      &  0.10675 &  0.10802 &   0.10499 &  0.12076 &  0.16646 &  0.10919 &  0.13479 &  0.09946 &  0.10147 &  0.12532 &  0.16646 \\
wilt                           &  0.01277 &  0.01139 &   0.01261 &  0.01546 &  0.05012 &  0.01121 &  0.01482 &  0.02482 &  0.01143 &  0.01940 &  0.05012 \\
parkinsons                     &  0.06961 &  0.09297 &   0.06952 &  0.14952 &  0.17151 &  0.07539 &  0.12082 &  0.10892 &  0.14535 &  0.15381 &  0.17151 \\
heart                          &  0.12904 &  0.12965 &   0.12644 &  0.14171 &  0.14460 &  0.12269 &  0.16345 &  0.12483 &  0.14524 &  0.14515 &  0.14460 \\
wdbc                           &  0.03121 &  0.03055 &   0.03018 &  0.05259 &  0.05612 &  0.02360 &  0.05211 &  0.02770 &  0.04573 &  0.05651 &  0.05612 \\
bank                           &  0.00563 &  0.00587 &   0.00555 &  0.00003 &  0.10496 &  0.00264 &  0.01359 &  0.00703 &  0.00009 &  0.00001 &  0.10496 \\
ionosphere                     &  0.05140 &  0.05338 &   0.05037 &  0.04655 &  0.09328 &  0.05525 &  0.09978 &  0.09911 &  0.04409 &  0.04879 &  0.09328 \\
HRCompetencyScores             &  0.06117 &  0.06341 &   0.06077 &  0.05998 &  0.08260 &  0.06038 &  0.08265 &  0.05999 &  0.06143 &  0.05775 &  0.08260 \\
spambase                       &  0.03835 &  0.03956 &   0.03778 &  0.05148 &  0.11457 &  0.03569 &  0.06655 &  0.05898 &  0.10883 &  0.05058 &  0.11457 \\
QSAR                           &  0.09440 &  0.09413 &   0.09356 &  0.09389 &  0.20999 &  0.09204 &  0.12753 &  0.09863 &  0.10342 &  0.10558 &  0.20999 \\
diabetes                       &  0.16144 &  0.15750 &   0.16060 &  0.19393 &  0.17464 &  0.16062 &  0.18037 &  0.15760 &  0.16363 &  0.20136 &  0.17464 \\
breast                         &  0.03121 &  0.03055 &   0.03018 &  0.05259 &  0.05612 &  0.02360 &  0.05211 &  0.02770 &  0.04573 &  0.05651 &  0.05612 \\
SPF                            &  0.01779 &  0.00000 &   0.01735 &  0.23802 &  0.22118 &  0.02862 &  0.00096 &  0.00208 &  0.22564 &  0.22791 &  0.22118 \\
hillvalley                     &  0.25145 &  0.24703 &   0.24952 &  0.18358 &  0.39934 &  0.24856 &  0.25610 &  0.04293 &  0.18389 &  0.24316 &  0.39934 \\
pc4                            &  0.06180 &  0.06131 &   0.06136 &  0.11005 &  0.11454 &  0.06034 &  0.08019 &  0.06544 &  0.10640 &  0.10278 &  0.11454 \\
scene                          &  0.07103 &  0.01612 &   0.07025 &  0.01398 &  0.12587 &  0.01499 &  0.01986 &  0.01259 &  0.01065 &  0.01486 &  0.12587 \\
Sonar\_Mine\_Rock\_Data           &  0.12791 &  0.14001 &   0.12677 &  0.12427 &  0.20313 &  0.11447 &  0.21366 &  0.16265 &  0.21443 &  0.19161 &  0.20313 \\
Customer\_Churn                 &  0.03185 &  0.03243 &   0.03156 &  0.09912 &  0.12122 &  0.02851 &  0.04712 &  0.06977 &  0.11074 &  0.08141 &  0.12122 \\
jm1                            &  0.13393 &  0.13624 &   0.13302 &  0.15023 &  0.16375 &  0.13745 &  0.14270 &  0.14019 &  0.16392 &  0.14728 &  0.16375 \\
eeg                            &  0.06812 &  0.19969 &   0.06710 &  0.24489 &  0.31479 &  0.16554 &  0.24724 &  0.24684 &  0.25008 &  0.24743 &  0.31479 \\
phoneme                        &  0.06717 &  0.06919 &   0.06646 &  0.08024 &  0.15476 &  0.06805 &  0.11000 &  0.15806 &  0.09761 &  0.07628 &  0.15476 \\
\bottomrule
Mean                           &  0.08561 &  0.08814 &   0.08476 &  0.10919 &  0.15484 &  0.08657 &  0.10984 &  0.09237 &  0.11152 &  0.11227 &  0.15484 \\
Rank                           &  4.40000 &  3.66667 &   3.10000 &  6.70000 &  9.70000 &  3.33333 &  7.73333 &  4.70000 &  6.40000 &  6.56667 &  9.70000 \\
\bottomrule
\end{tabular}
}
}\end{table}

% Real30 log-loss
\begin{table}[ht]
\centering
\caption{The Log Loss of RF trained on 30 real datasets in comparison with other ensemble and base learners}
\label{tab:log-loss_m}

\resizebox{0.8\columnwidth}{!}{
\rotatebox{90}{
\begin{tabular}{lrrrrrrrrrrr}
\toprule
{} &     RF\_d &   RF\_opt &  RF\_large &  DNN\_ens &    XGB\_d &  XGB\_opt &   DT\_opt &   LR\_opt &  SVM\_opt &  DNN\_opt &  GNB\_opt \\
Data                           &          &          &           &          &          &          &          &          &          &          &          \\
\midrule
cm1                            &  0.34404 &  0.28584 &   0.30514 &  0.44681 &  0.41276 &  0.30143 &  0.69459 &  0.29882 &  0.34533 &  0.31967 &  0.41276 \\
datatrieve                     &  0.74000 &  0.39064 &   0.31912 &  0.37264 &  0.41449 &  0.27973 &  1.20116 &  0.31675 &  0.35517 &  0.38126 &  0.41449 \\
kc1\_class\_level\_defectiveornot &  0.50759 &  0.50239 &   0.49964 &  1.11941 &  0.65186 &  0.55237 &  1.58172 &  0.71746 &  0.56170 &  0.57589 &  0.65186 \\
kc1                            &  0.51141 &  0.34364 &   0.44639 &  0.37286 &  0.45253 &  0.35578 &  0.45643 &  0.36047 &  0.59598 &  0.36570 &  0.45253 \\
kc2                            &  0.75841 &  0.38836 &   0.69166 &  0.56043 &  1.73512 &  0.38679 &  0.96272 &  0.39190 &  0.44661 &  0.41471 &  1.73512 \\
kc3                            &  0.35456 &  0.36265 &   0.30813 &  0.41222 &  0.36397 &  0.26754 &  0.93353 &  0.29771 &  0.28759 &  0.27402 &  0.36397 \\
pc1                            &  0.25835 &  0.19982 &   0.19813 &  0.37817 &  0.29047 &  0.20390 &  0.54296 &  0.22336 &  0.25214 &  0.27985 &  0.29047 \\
spect                          &  0.44787 &  0.38377 &   0.42046 &  0.59101 &  1.62621 &  0.39752 &  1.55935 &  0.38292 &  0.39659 &  0.38968 &  1.62621 \\
spectf                         &  0.38524 &  0.37898 &   0.37992 &  0.39631 &  4.49957 &  0.38503 &  1.63603 &  0.35630 &  0.37198 &  0.43798 &  4.49957 \\
vertebral                      &  0.33123 &  0.33112 &   0.32686 &  0.35862 &  0.50048 &  0.33693 &  1.71634 &  0.30560 &  0.31436 &  0.38119 &  0.50048 \\
wilt                           &  0.06635 &  0.05200 &   0.05575 &  0.05559 &  0.20082 &  0.04331 &  0.28349 &  0.08680 &  0.04140 &  0.07010 &  0.20082 \\
parkinsons                     &  0.23549 &  0.30034 &   0.23476 &  0.48127 &  0.56079 &  0.24394 &  1.34141 &  0.34042 &  0.44961 &  0.49947 &  0.56079 \\
heart                          &  0.40464 &  0.40565 &   0.39867 &  0.44041 &  0.49625 &  0.38678 &  1.72870 &  0.40052 &  0.45131 &  0.45611 &  0.49625 \\
wdbc                           &  0.18249 &  0.14528 &   0.14683 &  0.17788 &  0.60946 &  0.08975 &  1.00428 &  0.11361 &  0.16345 &  0.20345 &  0.60946 \\
bank                           &  0.02711 &  0.02814 &   0.02710 &  0.00102 &  0.31807 &  0.01518 &  0.39723 &  0.02238 &  0.00208 &  0.00036 &  0.31807 \\
ionosphere                     &  0.22357 &  0.21389 &   0.18485 &  0.17769 &  1.00231 &  0.20413 &  2.21956 &  0.41472 &  0.15798 &  0.20799 &  1.00231 \\
HRCompetencyScores             &  0.34997 &  0.33550 &   0.28702 &  0.22406 &  0.76670 &  0.21816 &  1.50664 &  0.20809 &  0.21236 &  0.21079 &  0.76670 \\
spambase                       &  0.18099 &  0.15996 &   0.15204 &  0.18824 &  0.76028 &  0.13125 &  1.11660 &  0.23260 &  0.35195 &  0.18839 &  0.76028 \\
QSAR                           &  0.34198 &  0.34713 &   0.31701 &  0.39485 &  0.60997 &  0.30448 &  1.54867 &  0.33387 &  0.35113 &  0.40784 &  0.60997 \\
diabetes                       &  0.49559 &  0.47766 &   0.48513 &  0.58536 &  0.58203 &  0.48470 &  1.10947 &  0.48597 &  0.49785 &  0.60225 &  0.58203 \\
breast                         &  0.18249 &  0.14528 &   0.14683 &  0.17788 &  0.60946 &  0.08975 &  1.00428 &  0.11361 &  0.16345 &  0.20345 &  0.60946 \\
SPF                            &  0.10368 &  0.00001 &   0.10370 &  1.18113 &  0.64225 &  0.17288 &  0.01307 &  0.03734 &  0.65038 &  0.64867 &  0.64225 \\
hillvalley                     &  0.70298 &  0.68760 &   0.69778 &  0.56369 &  1.90916 &  0.69079 &  0.81839 &  0.90690 &  0.55075 &  1.74015 &  1.90916 \\
pc4                            &  0.19637 &  0.23117 &   0.19538 &  0.46422 &  0.45053 &  0.18996 &  0.65377 &  0.23340 &  0.37316 &  0.34091 &  0.45053 \\
scene                          &  0.25442 &  0.09200 &   0.25024 &  0.08180 &  3.56460 &  0.06919 &  0.50277 &  0.06107 &  0.05550 &  0.07549 &  3.56460 \\
Sonar\_Mine\_Rock\_Data           &  0.41125 &  0.44403 &   0.40972 &  0.46980 &  0.59555 &  0.36404 &  2.47486 &  0.49811 &  0.61975 &  0.56668 &  0.59555 \\
Customer\_Churn                 &  0.13440 &  0.14098 &   0.11849 &  0.31892 &  0.38788 &  0.09815 &  0.79182 &  0.22437 &  0.36859 &  0.27040 &  0.38788 \\
jm1                            &  0.54955 &  0.43127 &   0.49348 &  0.48032 &  0.53546 &  0.43583 &  0.47089 &  0.44750 &  0.66004 &  0.46364 &  0.53546 \\
eeg                            &  0.25426 &  0.58559 &   0.25307 &  0.68300 &  0.83746 &  0.51486 &  0.68798 &  0.68904 &  0.81550 &  0.68891 &  0.83746 \\
phoneme                        &  0.25496 &  0.26020 &   0.23672 &  0.26251 &  0.46562 &  0.22679 &  1.56563 &  0.47195 &  0.31852 &  0.29424 &  0.46562 \\
\bottomrule
Mean                           &  0.33971 &  0.30170 &   0.30300 &  0.41394 &  0.89507 &  0.28136 &  1.08414 &  0.33245 &  0.37274 &  0.39864 &  0.89507 \\
Rank                           &  5.66667 &  3.66667 &   3.70000 &  6.43333 &  9.16667 &  2.56667 &  9.86667 &  4.26667 &  5.43333 &  6.06667 &  9.16667 \\
\bottomrule
\end{tabular}
}
}
\end{table}

% Real30 ECE
\begin{table}[ht]
\centering
\caption{The ECE of RF trained on 30 real datasets in comparison with other ensemble and base learners}
\label{tab:ece_m}

\resizebox{0.8\columnwidth}{!}{
\rotatebox{90}{
\begin{tabular}{lrrrrrrrrrrr}
\toprule
{} &     RF\_d &   RF\_opt &  RF\_large &  DNN\_ens &    XGB\_d &  XGB\_opt &   DT\_opt &   LR\_opt &  SVM\_opt &  DNN\_opt &  GNB\_opt \\
Data                           &          &          &           &          &          &          &          &          &          &          &          \\
\midrule
cm1                            &  0.08585 &  0.06327 &   0.09443 &  0.03561 &  0.14952 &  0.05589 &  0.05602 &  0.12472 &  0.06122 &  0.02132 &  0.14952 \\
datatrieve                     &  0.09778 &  0.09100 &   0.10087 &  0.06755 &  0.09814 &  0.07750 &  0.07506 &  0.10899 &  0.07732 &  0.04623 &  0.09814 \\
kc1\_class\_level\_defectiveornot &  0.15259 &  0.14115 &   0.15703 &  0.15481 &  0.14999 &  0.14002 &  0.11228 &  0.17591 &  0.15774 &  0.12621 &  0.14999 \\
kc1                            &  0.06695 &  0.05572 &   0.06734 &  0.05686 &  0.20776 &  0.06051 &  0.04161 &  0.07765 &  0.04364 &  0.04261 &  0.20776 \\
kc2                            &  0.11614 &  0.11684 &   0.12629 &  0.11961 &  0.16718 &  0.12637 &  0.07093 &  0.12769 &  0.15945 &  0.05764 &  0.16718 \\
kc3                            &  0.11268 &  0.09968 &   0.10412 &  0.05774 &  0.18437 &  0.12575 &  0.08200 &  0.12314 &  0.09356 &  0.01455 &  0.18437 \\
pc1                            &  0.10781 &  0.10413 &   0.10973 &  0.05260 &  0.14411 &  0.11174 &  0.05634 &  0.11240 &  0.07910 &  0.06090 &  0.14411 \\
spect                          &  0.12102 &  0.09684 &   0.11683 &  0.16634 &  0.28492 &  0.11952 &  0.08172 &  0.10681 &  0.10854 &  0.11393 &  0.28492 \\
spectf                         &  0.10431 &  0.11872 &   0.10414 &  0.11530 &  0.26194 &  0.15097 &  0.11455 &  0.12395 &  0.12895 &  0.10032 &  0.26194 \\
vertebral                      &  0.11566 &  0.11758 &   0.11352 &  0.14897 &  0.06644 &  0.13213 &  0.12954 &  0.12550 &  0.11503 &  0.14533 &  0.06644 \\
wilt                           &  0.08824 &  0.08965 &   0.08920 &  0.08524 &  0.00145 &  0.11183 &  0.09048 &  0.09972 &  0.12135 &  0.12228 &  0.00145 \\
parkinsons                     &  0.08764 &  0.11169 &   0.09584 &  0.11771 &  0.17822 &  0.10877 &  0.08830 &  0.12274 &  0.11858 &  0.11183 &  0.17822 \\
heart                          &  0.09520 &  0.09564 &   0.09333 &  0.09760 &  0.13583 &  0.10923 &  0.09834 &  0.10773 &  0.11186 &  0.11499 &  0.13583 \\
wdbc                           &  0.08561 &  0.08189 &   0.07581 &  0.11176 &  0.12527 &  0.10244 &  0.08273 &  0.12388 &  0.10925 &  0.11250 &  0.12527 \\
bank                           &  0.06301 &  0.05730 &   0.05993 &  0.00069 &  0.08573 &  0.04668 &  0.04552 &  0.10348 &  0.00293 &  0.00011 &  0.08573 \\
ionosphere                     &  0.08202 &  0.07852 &   0.08319 &  0.10005 &  0.12614 &  0.09916 &  0.09282 &  0.10709 &  0.10542 &  0.10040 &  0.12614 \\
HRCompetencyScores             &  0.07438 &  0.07719 &   0.07401 &  0.09534 &  0.08505 &  0.08885 &  0.07830 &  0.09008 &  0.08646 &  0.08978 &  0.08505 \\
spambase                       &  0.03635 &  0.03435 &   0.03658 &  0.03703 &  0.06604 &  0.06540 &  0.06338 &  0.03531 &  0.02028 &  0.05997 &  0.06604 \\
QSAR                           &  0.06413 &  0.06933 &   0.06210 &  0.10513 &  0.05061 &  0.09415 &  0.07207 &  0.07428 &  0.06671 &  0.09421 &  0.05061 \\
diabetes                       &  0.07172 &  0.07209 &   0.06280 &  0.07272 &  0.10416 &  0.07954 &  0.07260 &  0.07302 &  0.08046 &  0.07447 &  0.10416 \\
breast                         &  0.08561 &  0.08189 &   0.07581 &  0.11176 &  0.12527 &  0.10244 &  0.08273 &  0.12388 &  0.10925 &  0.11250 &  0.12527 \\
SPF                            &  0.06359 &  0.00000 &   0.06234 &  0.15374 &  0.08339 &  0.03440 &  0.00181 &  0.00913 &  0.04048 &  0.03979 &  0.08339 \\
hillvalley                     &  0.08172 &  0.06646 &   0.08302 &  0.06166 &  0.21329 &  0.03591 &  0.11039 &  0.08205 &  0.08778 &  0.07857 &  0.21329 \\
pc4                            &  0.07553 &  0.08033 &   0.07833 &  0.05951 &  0.17384 &  0.08737 &  0.08041 &  0.09379 &  0.03485 &  0.06537 &  0.17384 \\
scene                          &  0.06871 &  0.05690 &   0.07451 &  0.07629 &  0.20868 &  0.07806 &  0.09517 &  0.06350 &  0.00250 &  0.09506 &  0.20868 \\
Sonar\_Mine\_Rock\_Data           &  0.08872 &  0.09252 &   0.08961 &  0.13661 &  0.15346 &  0.11306 &  0.11038 &  0.11418 &  0.11293 &  0.10341 &  0.15346 \\
Customer\_Churn                 &  0.06237 &  0.06810 &   0.06468 &  0.05132 &  0.01782 &  0.09421 &  0.09730 &  0.05343 &  0.08739 &  0.07550 &  0.01782 \\
jm1                            &  0.03254 &  0.03025 &   0.03164 &  0.05715 &  0.21604 &  0.03611 &  0.04463 &  0.04867 &  0.03345 &  0.03720 &  0.21604 \\
eeg                            &  0.02636 &  0.03501 &   0.02744 &  0.04774 &  0.07053 &  0.05804 &  0.00450 &  0.04215 &  0.05879 &  0.03979 &  0.07053 \\
phoneme                        &  0.01688 &  0.01754 &   0.01729 &  0.02034 &  0.02918 &  0.03000 &  0.04142 &  0.02199 &  0.01644 &  0.04385 &  0.02918 \\
\bottomrule
Mean                           &  0.08104 &  0.07672 &   0.08106 &  0.08583 &  0.13215 &  0.08920 &  0.07578 &  0.09323 &  0.08106 &  0.07669 &  0.13215 \\
Rank                           &  4.36667 &  3.93333 &   4.53333 &  5.76667 &  8.73333 &  6.63333 &  4.70000 &  7.43333 &  5.70000 &  5.46667 &  8.73333 \\
\bottomrule
\end{tabular}
}
}
\end{table}

\clearpage

% Real30 CD ACC

% Real30 CD BS

% Real30 CD log-loss

% Real30 CD ECE

% run time table
\begin{table}[ht]
\centering
\caption{Average run time of ML learners on 30 real datasets}
\label{tab:model_runtime}

\resizebox{\textwidth}{!}{%
\begin{tabular}{lrrrrrrrrrrr}
\toprule
{} &     RF\_d &      RF\_opt &  RF\_large &   DNN\_ens &    XGB\_d &    XGB\_opt &    DT\_opt &     LR\_opt &     SVM\_opt &    DNN\_opt &  GNB\_opt \\
                           &          &             &           &           &          &            &           &            &             &            &          \\
\midrule
Runtime                        &  0.33 &   200.26 &   1.70 &   4.92 &  0.08 &   45.25 &   1.56 &   31.94 &    93.18 &  119.83 &  0.05 \\
\bottomrule
\end{tabular}%
}
\label{tab:time}
\end{table}

% \vfill

% \end{document}

\end{document}